\documentclass{article}

\usepackage{microtype}
\usepackage{graphicx}
\usepackage{subcaption}
\usepackage{booktabs} 

\usepackage{hyperref}

\usepackage[preprint]{icml2026}

\usepackage{amsmath}
\usepackage{amssymb}
\usepackage{mathtools}
\usepackage{amsthm}
\usepackage{bbm}
\usepackage{colortbl}
\usepackage[capitalize,noabbrev]{cleveref}

\theoremstyle{plain}

\theoremstyle{definition}

\theoremstyle{remark}

\usepackage[textsize=tiny]{todonotes}

\icmltitlerunning{MOMA: Masked Orthogonal Matrix Alignment for Zero-Additional-Parameter Model Merging}

\begin{document}

\twocolumn[
  \icmltitle{MOMA: Masked Orthogonal Matrix Alignment for Zero-Additional-Parameter Model Merging}



  \icmlsetsymbol{equal}{*}

  \begin{icmlauthorlist}
    \icmlauthor{Fanshuang Kong}{111}
    \icmlauthor{Richong Zhang}{111}
    \icmlauthor{Zhijie Nie}{111}
    \icmlauthor{Hang Zhou}{111}
    \icmlauthor{Ziqiao Wang}{222}
    \icmlauthor{Qiang Sun}{333}
    \icmlauthor{Chunming Hu}{111}
  \end{icmlauthorlist}

  \icmlaffiliation{111}{Beihang University, Beijing, China}
  \icmlaffiliation{222}{Tongji University, Shanghai, China}
  \icmlaffiliation{333}{University of Toronto, Toronto, Canada}

  \icmlcorrespondingauthor{Richong Zhang}{zhangrc@act.buaa.edu.cn}

  \icmlkeywords{Machine Learning, ICML}

  \vskip 0.3in
]



\printAffiliationsAndNotice{}  

\begin{abstract}
  Model merging offers a scalable alternative to multi-task learning but often yields suboptimal performance on classification tasks. We attribute this degradation to a geometric misalignment between the merged encoder and static task-specific classifier heads. Existing methods typically rely on auxiliary parameters to enforce strict representation alignment. We challenge this approach by revealing that the misalignment is predominantly an orthogonal transformation, rendering such strict alignment unnecessary. Leveraging this insight, we propose MOMA (Masked Orthogonal Matrix Alignment), which rectifies the misalignment by jointly optimizing a global multi-task vector mask and task-specific orthogonal transformations. Crucially, MOMA absorbs corresponding new parameters directly into the existing model weights, achieving performance comparable to state-of-the-art baselines with zero additional parameters and zero added inference cost.
\end{abstract}

\section{Introduction}
Model merging aims to directly combine multiple homologous fine-tuned models in the parameter space, bypassing the need for raw training data or retraining, to synthesize a unified model capable of solving multiple tasks~\cite{izmailov2018averaging_weights,jin2022dataless_regmean,ilharco2022editing_task_arithmetic,yangadamerging,yadav2024ties}. While model merging has shown success in many areas such as continual learning~\cite{marczak2024magmax,tang2025merging} and information retrieval~\cite{li2024improving,braga2025investigating}, we observe that it often yields suboptimal results in classification tasks.

We attribute this performance degradation to a critical misalignment between the merged encoder and the static task-specific classifier heads. 
Existing justifications for parameter-space merging typically rely on the Neural Tangent Kernel (NTK)~\cite{jacot2018neural} linearization, also known as the \emph{lazy regime}~\cite{chizat2019lazy}, under which parameter updates translate approximately linearly into function-space changes~\cite{ortiz2024task,yoshida2025mastering}. This perspective provides a clear explanation for why simple parameter-space averaging can work. However, practical finite-width networks are inherently nonlinear, and behavior beyond the lazy regime (e.g., the rich/feature-learning regime~\cite{chizat2019lazy}) is not well captured by NTK-based theory~\cite{yehudai2019power,ghorbani2019limitations,karp2021local}. In particular, task vectors~\cite{ilharco2022editing_task_arithmetic} can move parameters far from initialization, so the merged model may depart substantially from the local tangent space and violate the linear approximation. As a result, parameter-space merging without further adjustment is often suboptimal in practice.
To mitigate the resulting encoder–classifier mismatch,
recent methods~\cite{pmlr-v235-yang24t,wei2025probsurgery} typically introduce auxiliary parameters to enforce strict alignment between the merged and fine-tuned representations. 
While effective, this architectural expansion can be an unnecessary over-constraint that conflicts with the ethos of model merging, namely combining existing weights with no extra storage or inference cost~\cite{wortsman2022model_model_soup}.
We argue that the appropriate target is functional consistency in logit space rather than exact representation matching~\cite{ortiz2024task}.
Consequently, it is sufficient to ensure that the merged representations preserve essential task-discriminative information and are properly aligned with the static classifier heads.

In this work, we conduct a series of geometric analyses to validate this argument, revealing that the discrepancy between the merged and fine-tuned representations is predominantly characterized by an orthogonal transformation. We substantiate this by first demonstrating, via visualization and KNN evaluation, that task-relevant information within the merged representations remains intact. Subsequently, by probing with learnable transformations under varying constraints, we confirm that mitigating the discrepancy via an orthogonal transformation is sufficient to restore the predictive capabilities of the fine-tuned models. This hypothesis is further corroborated by quantitative metric analysis, where high cosine similarity and coefficients of the orthogonal aligned and fine-tuned representation verify the orthogonality.

Based on this insight, we propose a zero-additional-parameter merging framework termed MOMA (Masked Orthogonal Matrix Alignment). Distinct from methods that enforce strict representation alignment, MOMA prioritizes output logit consistency by rectifying the inherent orthogonal misalignment between the merged encoder and the task-specific classifier heads. Specifically, we jointly optimize task-specific orthogonal transformations and a global multi-task vector mask. The orthogonal transformation rotates the merged representations to align with the classifier heads, while the mask selectively filters redundant parameters to relax optimization constraints and facilitate convergence. Crucially, we exploit the inner-product-preserving property of orthogonal transformations to absorb the matrices directly into the classifier head weights, and permanently apply the learned mask to the multi-task vector. This design ensures that MOMA achieves substantial performance gains with zero additional parameters and zero added inference cost.

Our contributions are summarized as follows:
\begin{itemize}
    \item We first identify the misalignment between merged encoders and static classifier heads as predominantly orthogonal, revealing that the widely used strict representation alignment is an unnecessary over-constraint.

    \item We propose MOMA, a zero-additional-parameter framework that rectifies orthogonal misalignment by jointly optimizing a multi-task vector mask and orthogonal transformations.
    
    \item Extensive experiments demonstrate that MOMA achieves state-of-the-art performance, bridging the gap to fine-tuned models with zero overhead.
\end{itemize}

\section{Background}

\subsection{Notation and Problem Setup}
Consider a $c$-class classification task. Let $\mathcal{X}\subseteq \mathbb{R}^d$ denote the instance space and $\mathcal{Y}\subseteq \mathbb{R}^c$ the label space. We consider a model $f: \mathcal{X}\to\mathcal{Y}$ (e.g., a neural network) that decomposes into a feature encoder $g: \mathcal{X}\to\mathbb{R}^k$, parameterized by $\theta$, and a classifier head $h:\mathbb{R}^k\to\mathcal{Y}$, parameterized by $\phi$. Thus,
$f=h\circ g$, or equivalently, $f(x)=h\left(g(x;\theta);\phi\right)$.

Given a pre-trained (base) model with parameters $(\theta_b,\phi_b)$, we fine-tune it on $T$ downstream tasks. Each task $t\in\{1,\dots,T\}$ is associated with a dataset $\mathcal{D}_t=\{(x_i^{t},y_i^{t})\}_{i=1}^{|\mathcal{D}_t|}$, obtaining task-specific parameters $\{(\theta_t,\phi_t)\}_{t=1}^T$. In the model-merging setting, especially when tasks have heterogeneous label spaces, standard approaches merge the fine-tuned encoders $\{\theta_t\}_{t=1}^T$ into a shared backbone $\theta_m$, while keeping the task-specific heads $\{\phi_t\}_{t=1}^T$ fixed. The goal is to construct $\theta_m$ solely from the fine-tuned weights, without accessing any additional data, such that the merged model
$h\big(g(x;\theta_m);\phi_t\big)$
performs well on every task $t$.

\subsection{Task Arithmetic and Merging Objective}
Task Arithmetic~\cite{ilharco2022editing_task_arithmetic} defines a \textit{task vector} as the difference between the fine-tuned encoder parameters $\theta_t$ and the base initialization $\theta_b$, namely $\tau_t = \theta_t - \theta_b$. The merged encoder is then formed by a linear combination in parameter space,
\begin{equation}
    \theta_m = \theta_b + \lambda \sum_{t=1}^T \tau_t,
\end{equation}
where $\lambda$ is a scaling hyperparameter. Although this merging is performed in parameter space, the desired behavior is specified in the model's output (logit) space~\cite{ortiz2024task}. Ideally, when $\theta_m$ is paired with the task-specific head $\phi_t$, it should reproduce the predictions of the corresponding fine-tuned model, i.e.
\begin{equation}
    h(g(x; \theta_m); \phi_t) \approx h(g(x; \theta_t); \phi_t), \quad \forall x \in \mathcal{D}_t.
    \label{eq:merging_objective}
\end{equation}
This objective indicates the central challenge of model merging: additive updates to the encoder parameters must induce compatible representations so that fixed, task-specific classifier heads continue to produce correct logits.

\begin{figure*}[h]
    \centering
    \begin{tabular}{@{}c@{\hspace{7pt}}c@{\hspace{7pt}}c@{\hspace{7pt}}c@{\hspace{7pt}}c@{\hspace{7pt}}c@{}}
        \includegraphics[width=0.135\textwidth]{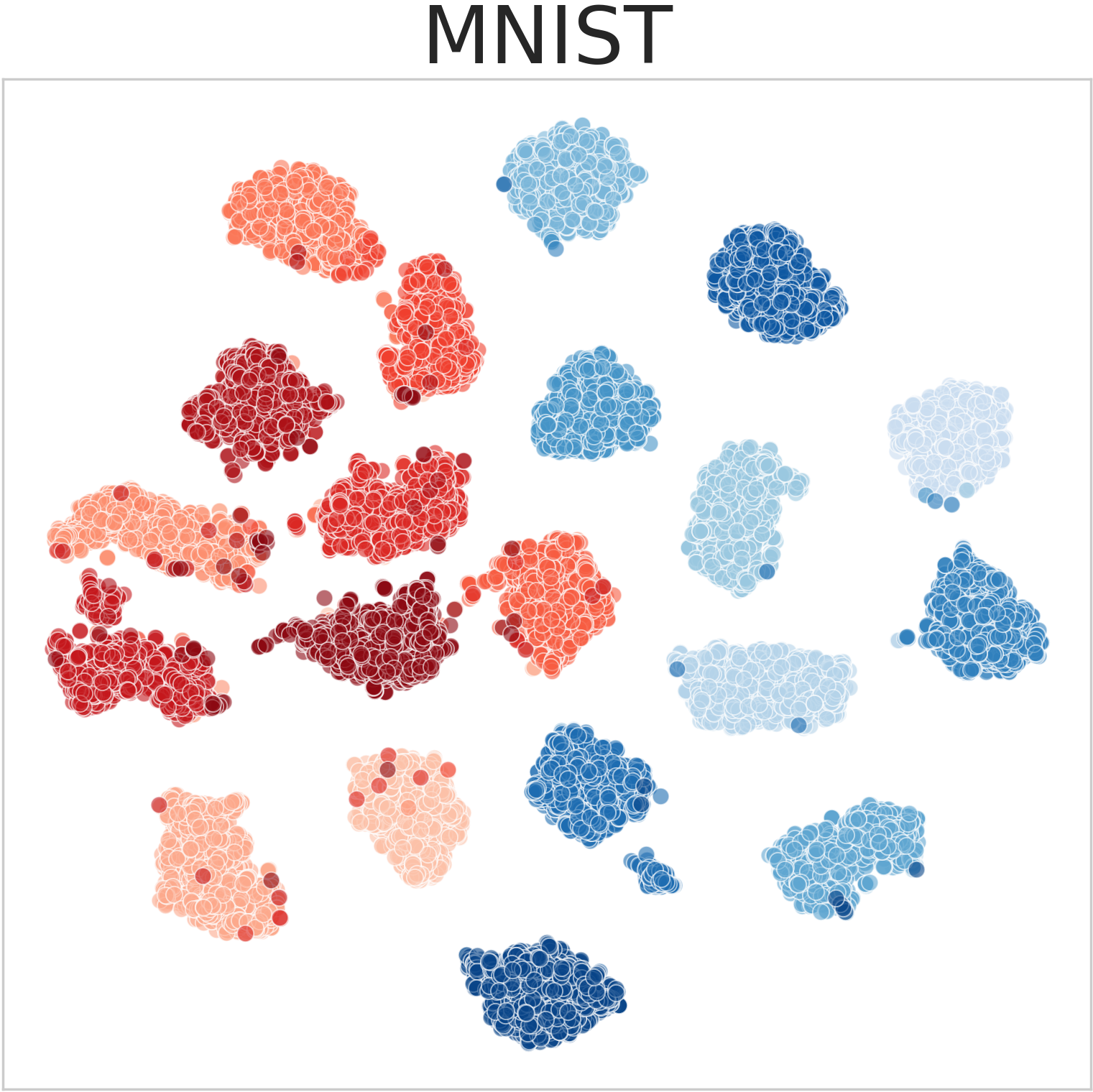} & \includegraphics[width=0.135\textwidth]{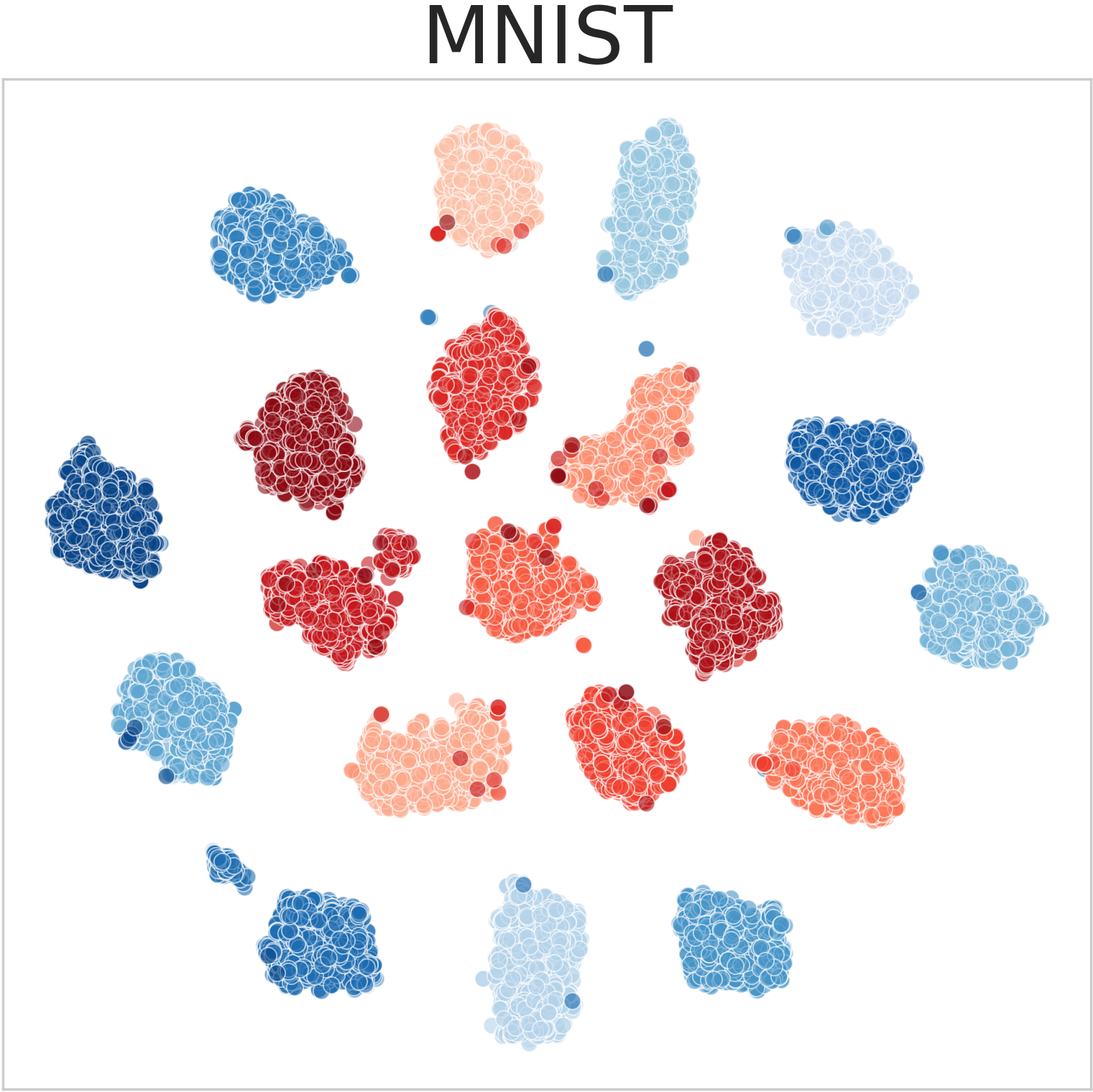} &
        \includegraphics[width=0.135\textwidth]{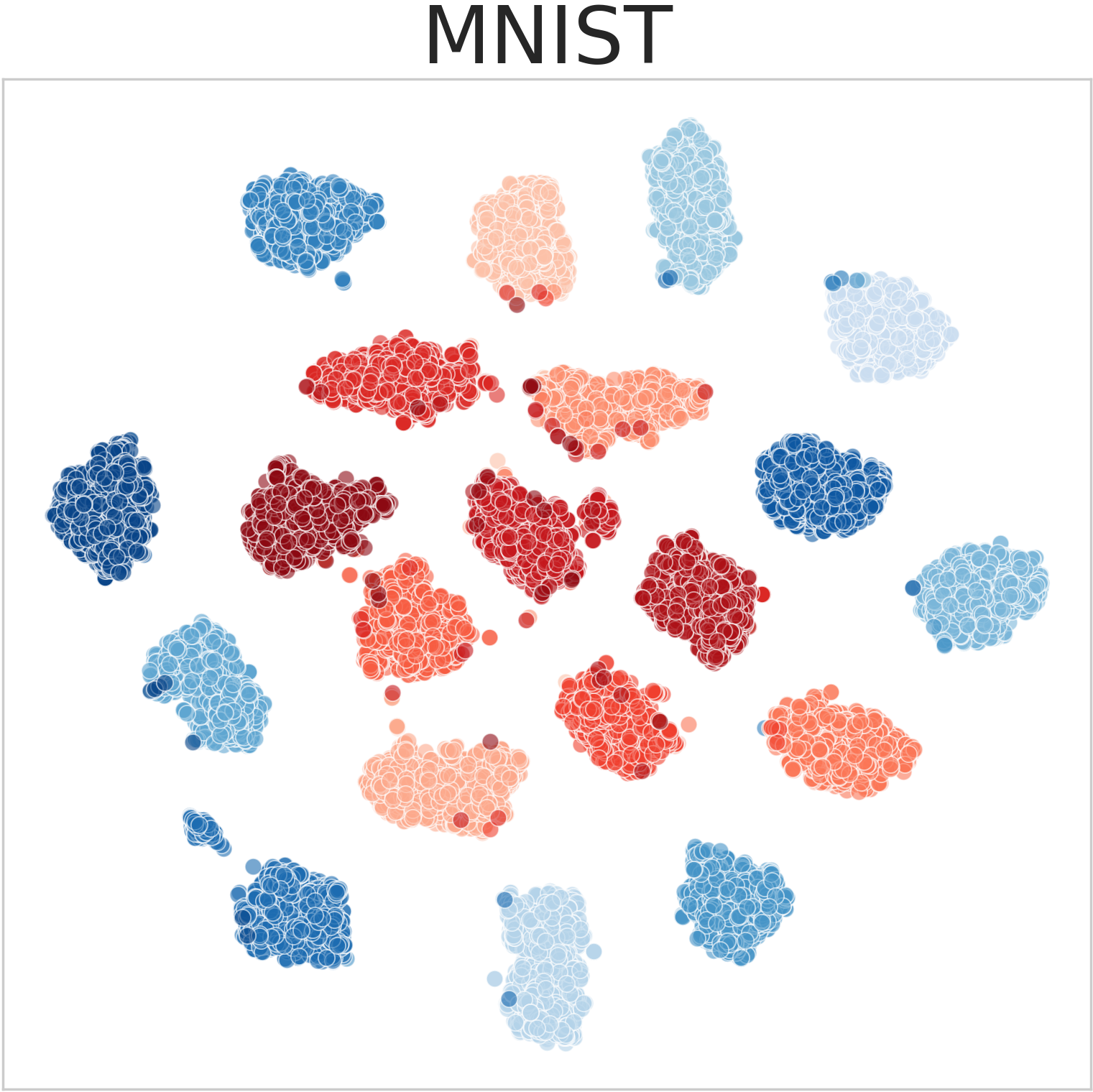} &
        \includegraphics[width=0.135\textwidth]{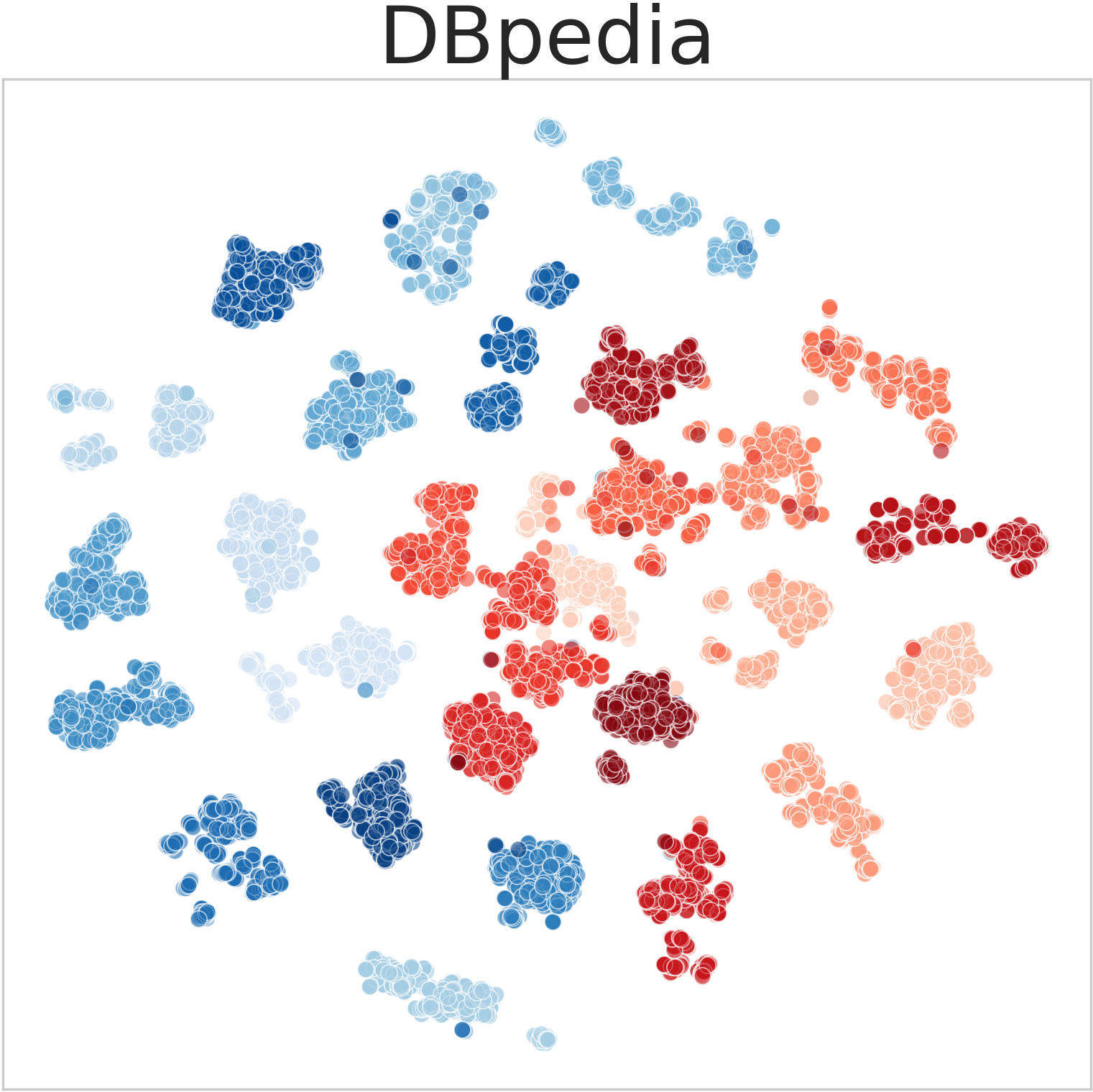} &
        \includegraphics[width=0.135\textwidth]{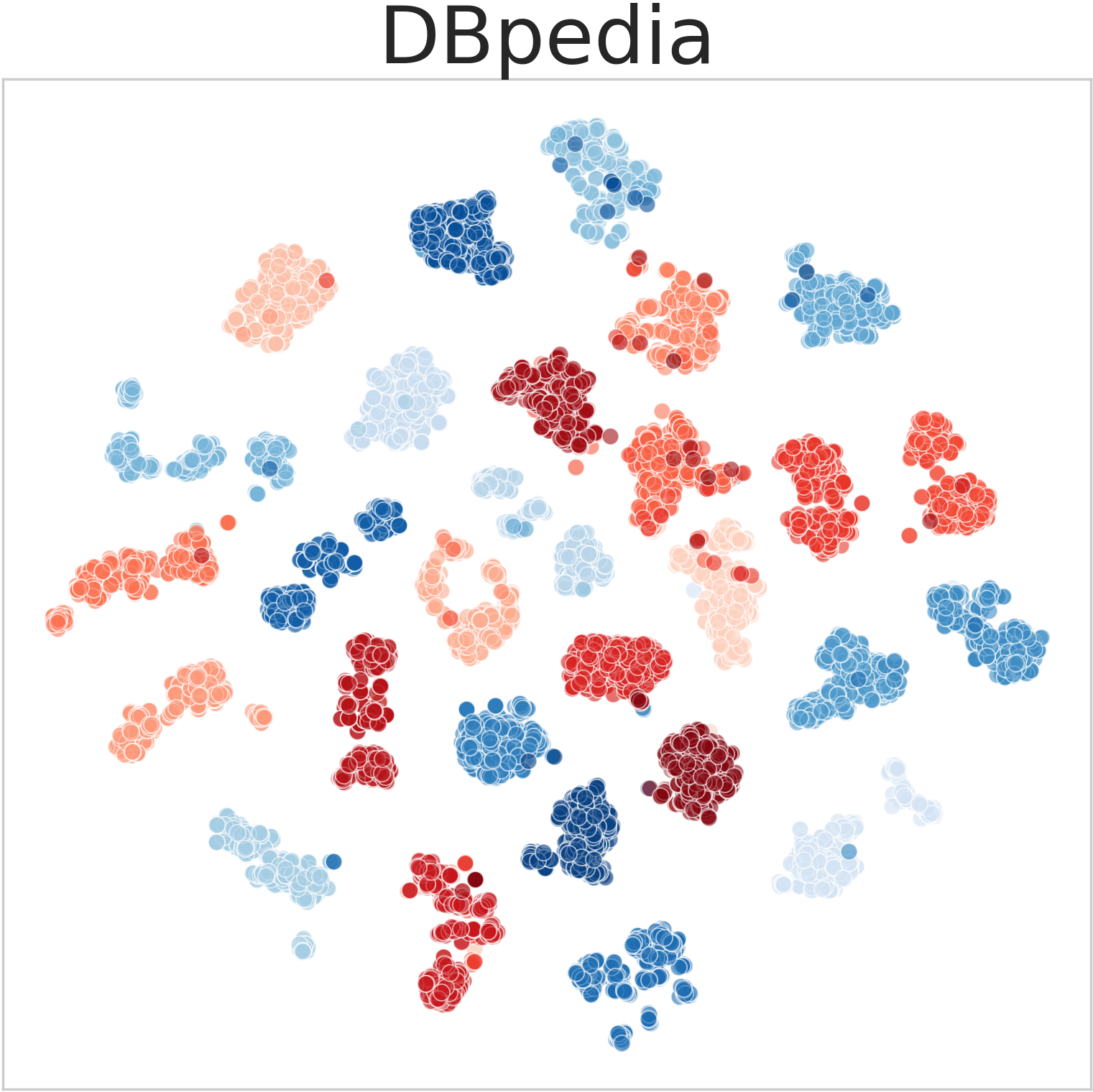} &
        \includegraphics[width=0.135\textwidth]{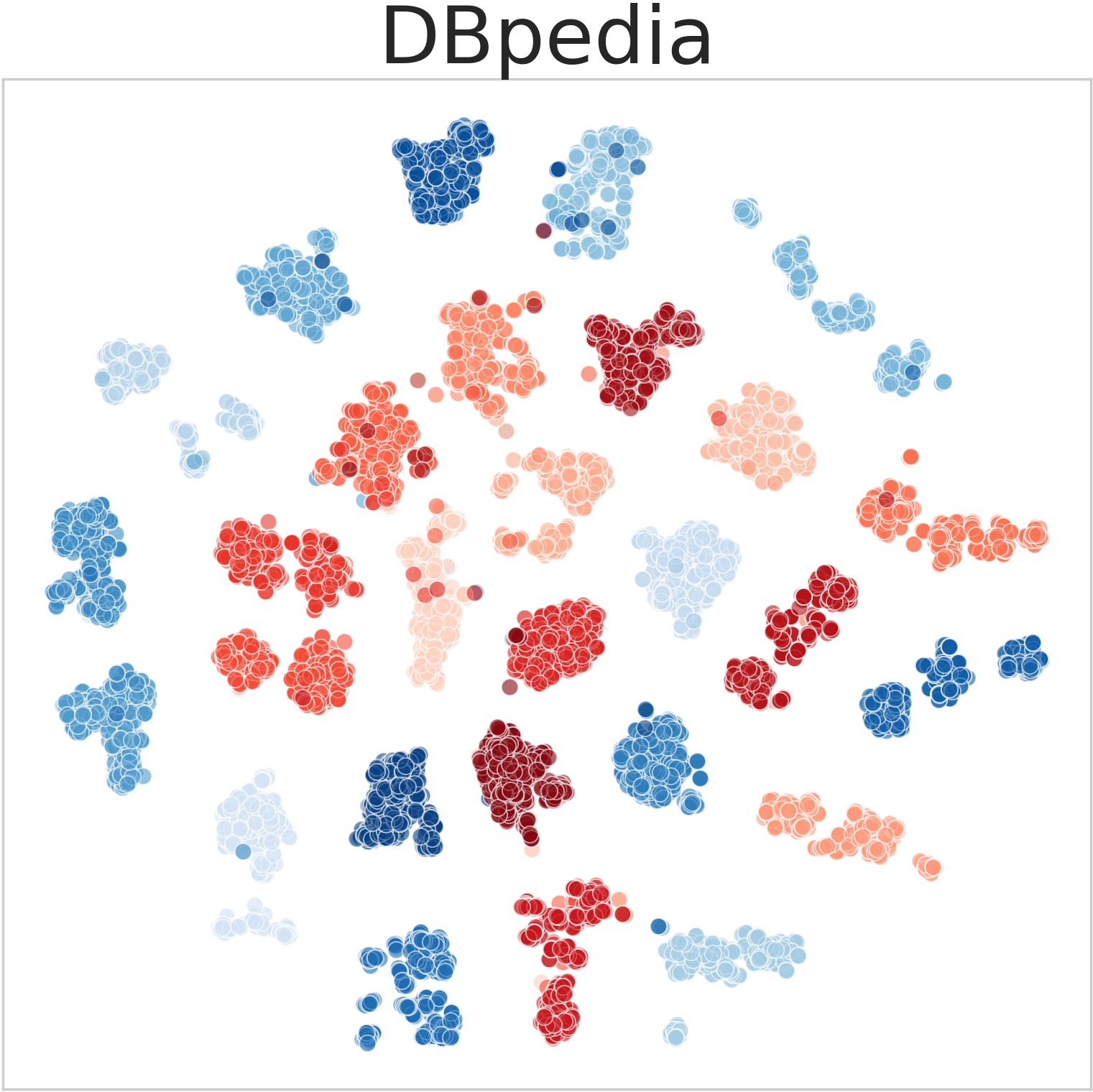} \\
        (a) WA (87.5\%) & (b) TA (97.3\%) & (c) Ties (96.5\%) & (d) WA (55.1\%) & (e) TA (81.5\%) & (f) Ties (83.4\%)
    \end{tabular}
    \caption{T-SNE visualization of representations from Weight Averaging (WA)~\cite{wortsman2022model_model_soup}, Task Arithmetic (TA), and Ties~\cite{yadav2024ties}. Blue points denote fine-tuned representations, while red points denote merged representations, with color shades indicating different class labels and values in parentheses denoting classification accuracy. Notably, despite significant differences in classification accuracy (e.g., WA vs. Ties on DBpedia), all merged representations maintain distinct class-conditional clusters. This visual evidence suggests that discriminative information is preserved,
    i.e., the representation intrinsic for class distinction has not collapsed in the merged model.
    }
    \label{fig:tsne-before}
\end{figure*}

\begin{figure*}[h]
    \centering
    \begin{tabular}{@{}c@{\hspace{7pt}}c@{\hspace{7pt}}c@{\hspace{7pt}}c@{\hspace{7pt}}c@{\hspace{7pt}}c@{}}
        \includegraphics[width=0.135\textwidth]{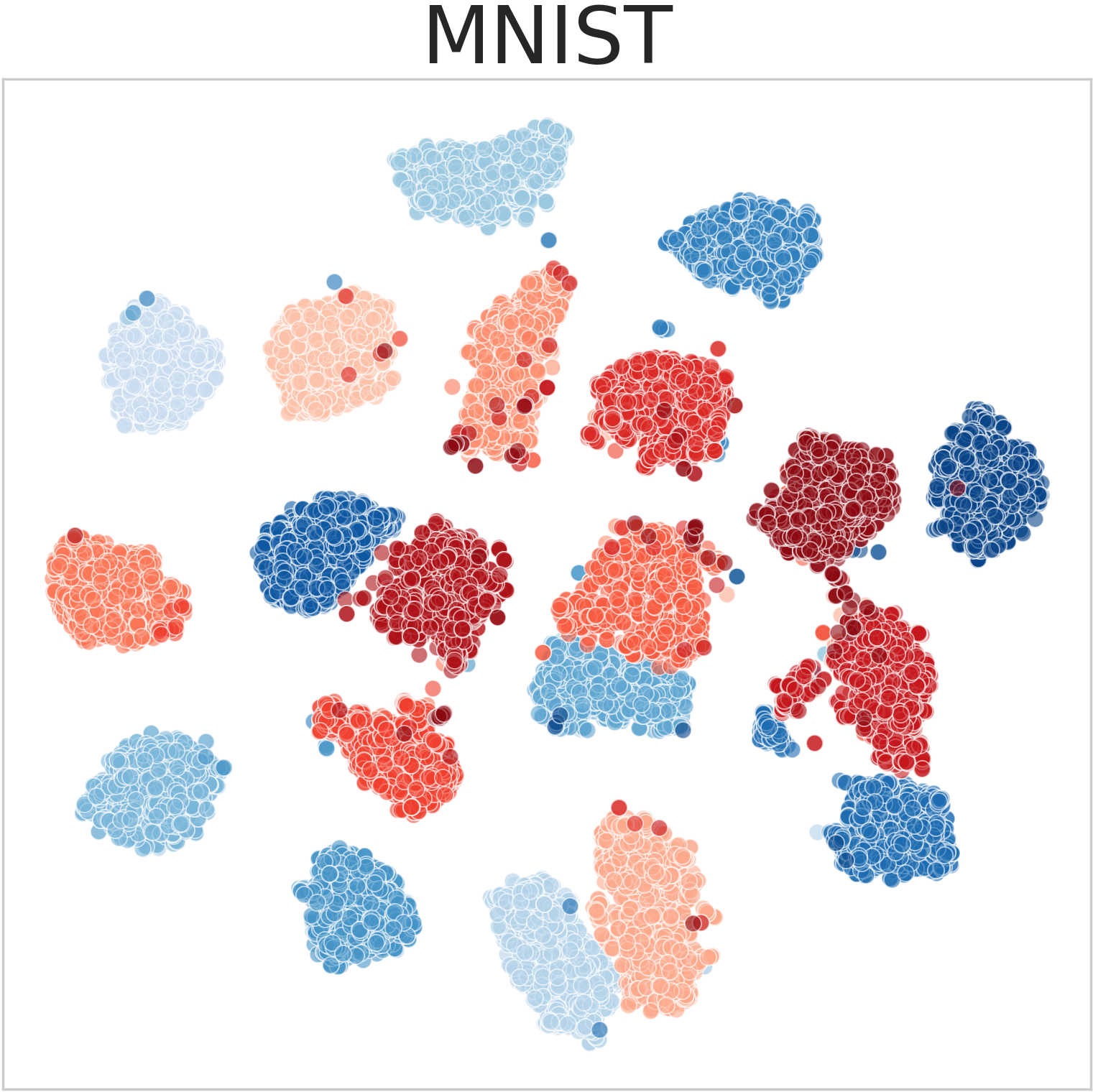} & \includegraphics[width=0.135\textwidth]{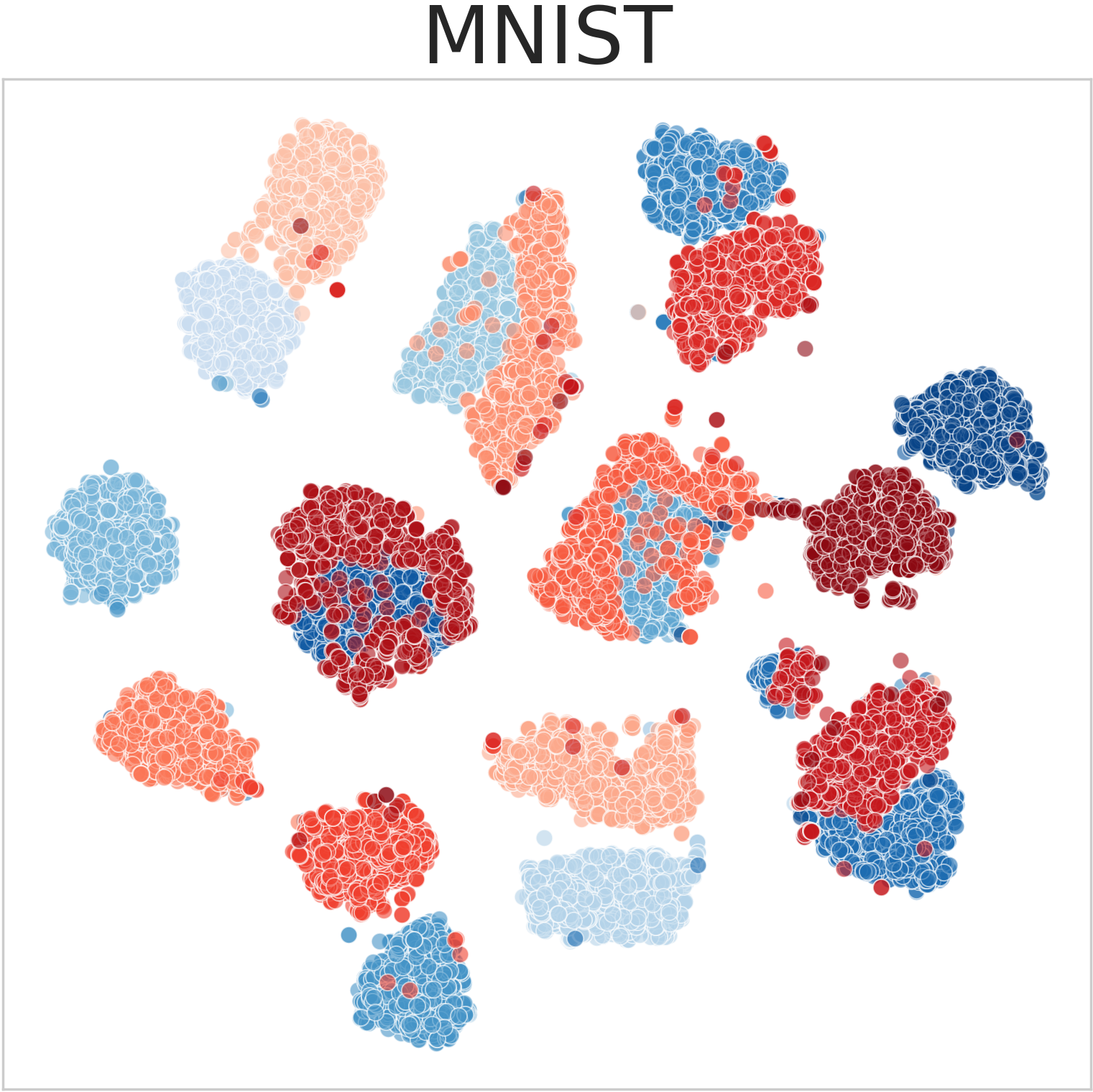} &
        \includegraphics[width=0.135\textwidth]{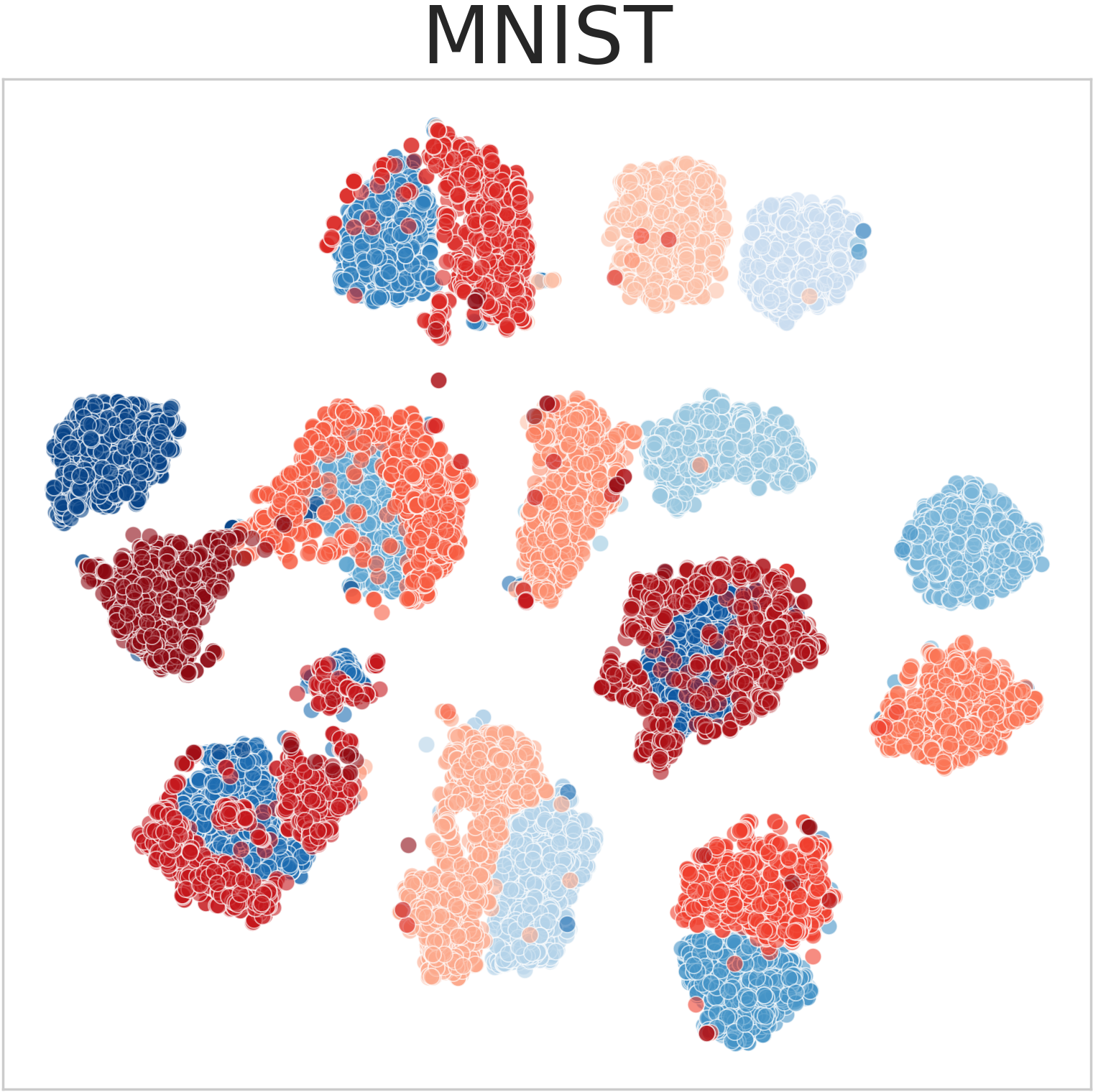} &
        \includegraphics[width=0.135\textwidth]{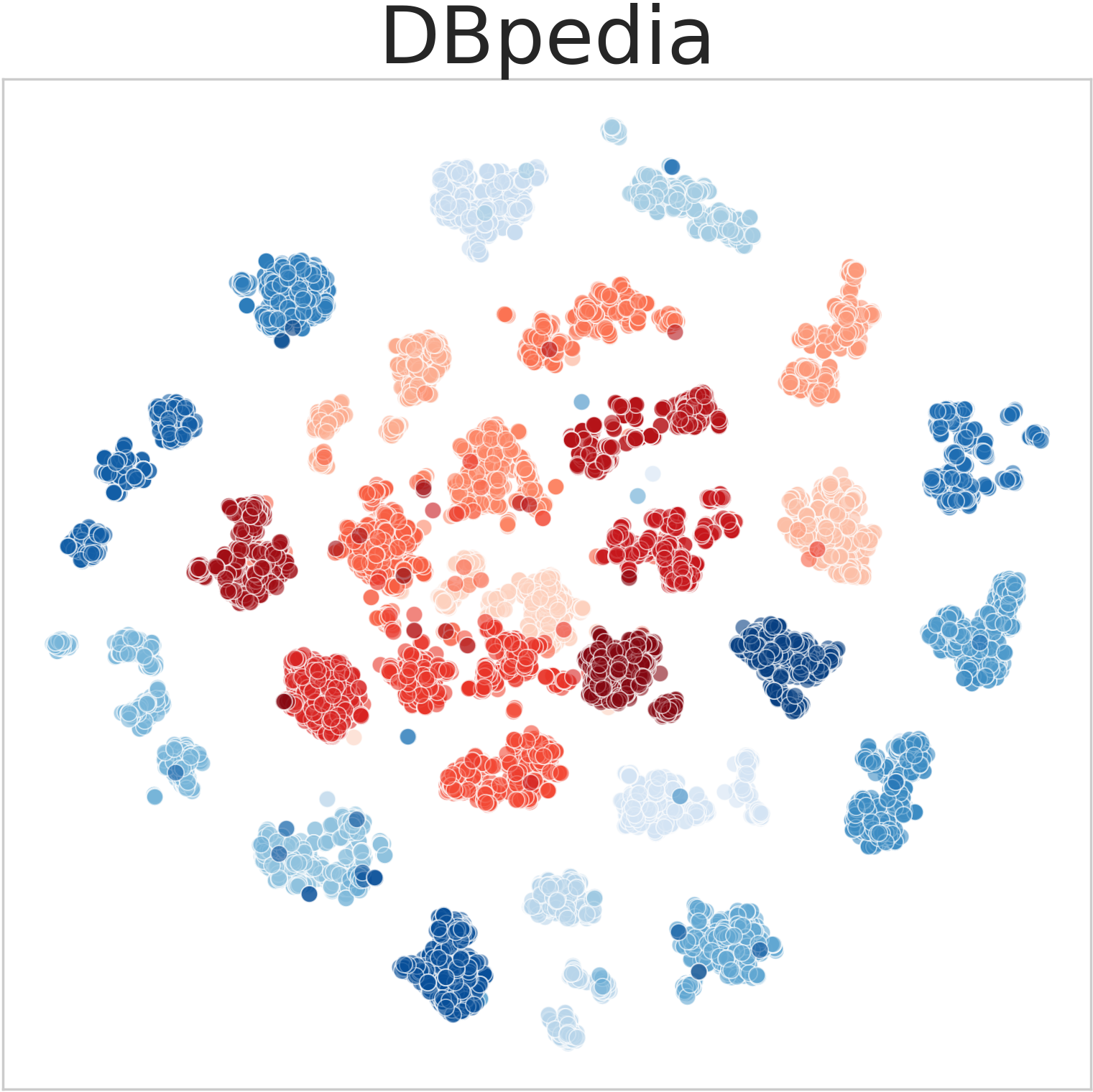} &
        \includegraphics[width=0.135\textwidth]{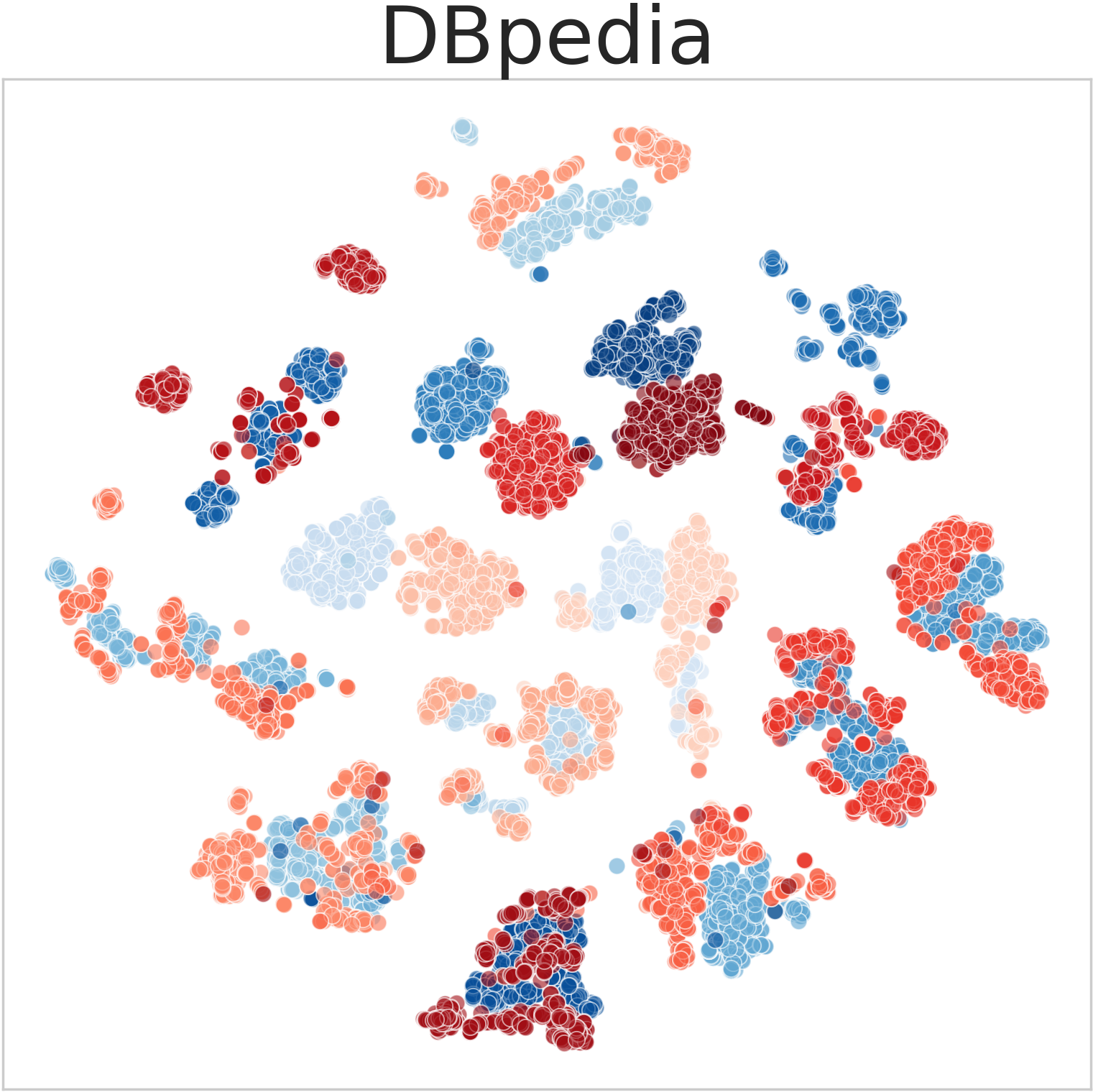} &
        \includegraphics[width=0.135\textwidth]{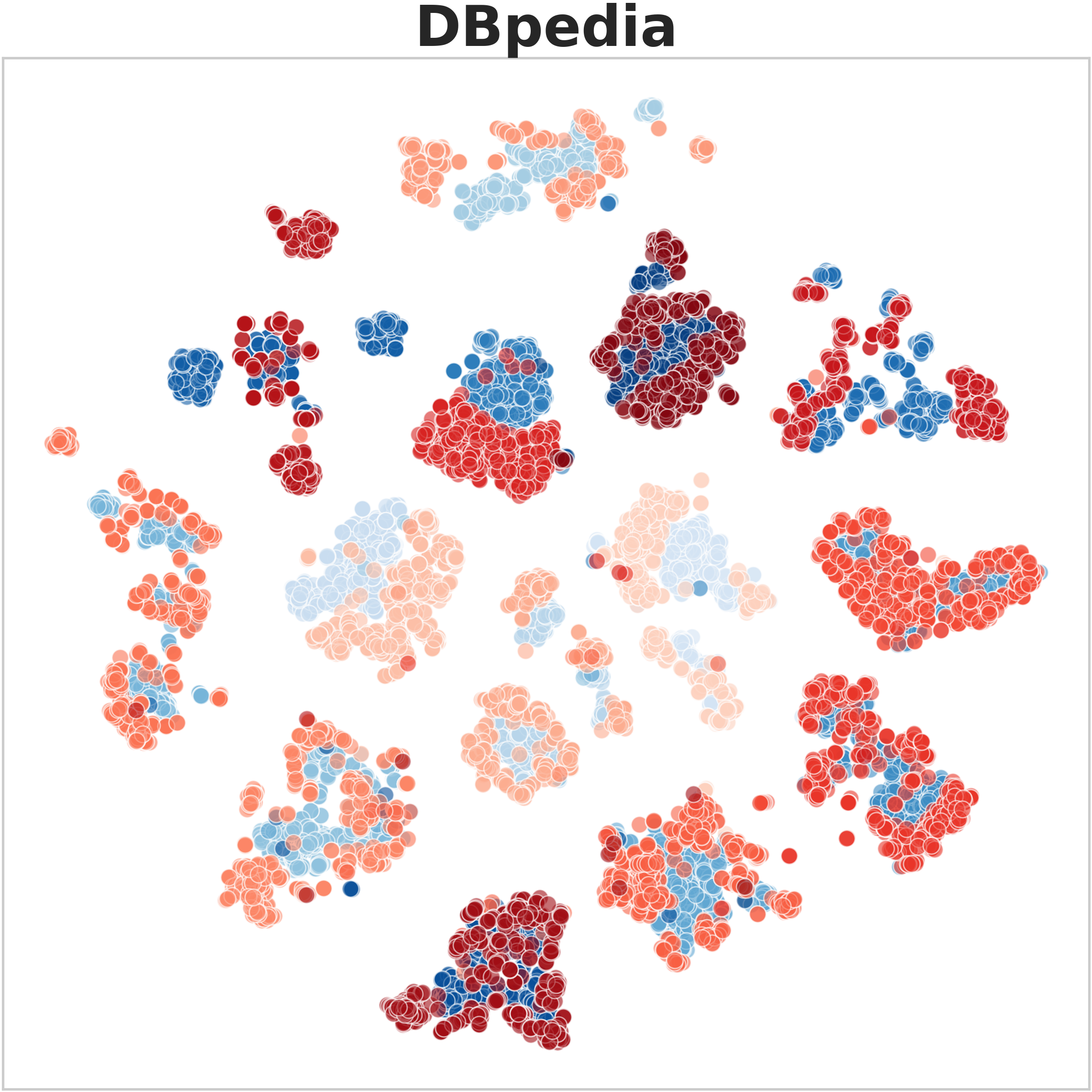} \\
        (a) WA (95.3\%) & (b) TA (98.2\%) & (c) Ties (97.7\%) & (d) WA (97.4\%) & (e) TA (97.6\%) & (f) Ties (98.3\%)
    \end{tabular}
    \caption{
        T-SNE visualization after orthogonal alignment. Merging methods and colors descriptions are as in Figure~\ref{fig:tsne-before}.
    Compared to the unaligned case, the classification accuracy shows a dramatic recovery.}
    \label{fig:tsne-after}
\end{figure*}

\subsection{NTK View: Linear Approximation and Its Limits}
The efficacy of parameter-space merging is often analyzed through the lens of the Neural Tangent Kernel (NTK)~\cite{jacot2018neural}. Specifically, in the infinite-width limit, gradient-based training  falls into the so-called ``lazy training'' (linearized) regime~\cite{chizat2019lazy}, where the network can be well approximated by its first-order Taylor expansion around the initialization $\theta_b$,
\begin{equation}
    g_{\text{lin}}(x; \theta) \triangleq g(x; \theta_b) + (\theta - \theta_b)^\top \nabla_{\theta_b} g(x; \theta_b).
    \label{eq:ntk_taylor}
\end{equation}
Under this linearization, adding task vectors in parameter space translates directly into adding their induced function,
\begin{align}
    g_{\text{lin}}(x; \theta_m) =& g(x; \theta_b) + \lambda\sum_{t=1}^T \tau_t^\top \nabla_{\theta_b} g(x; \theta_b).
    \label{eq:ntk_merging}
\end{align}
In particular, choosing $\lambda=\frac{1}{T}$ obtains $g_{\text{lin}}(x; \theta_m) =\frac{1}{T}\sum_{t=1}^T g_{\text{lin}}(x; \theta_t )$,
i.e., the merged encoder output is the average of the linearized task-specific encoders. Thus, within the NTK/lazy regime, parameter-space merging amounts to aggregating the individual task-induced updates in function space.
However, it is well known that this linear approximation can be inaccurate in realistic settings, even in relatively simple models~\cite{yehudai2019power,ghorbani2019limitations,karp2021local}.  In practice, fine-tuning often moves the parameters far from initialization and the neural network
actively learn task-relevant information, hence violating the conditions under which the linearization holds (i.e., the model operates in a rich or feature-learning regime~\cite{chizat2019lazy,geiger2020disentangling,woodworth2020kernel} rather than a purely lazy kernel regime). This departure creates a non-linear gap; the actual merged representation $g(x;\theta_m)$ can deviate substantially from the sum of the individual task-induced updates, leading to representation shifts that are poorly aligned with the fixed task-specific heads $\phi_t$, which were optimized for the fine-tuned representations.

\subsection{Rethinking Representation Alignment Costs}
To mitigate the performance degradation of parameter-space merging beyond the lazy regime, recent methods such as Surgery~\cite{pmlr-v235-yang24t} introduce auxiliary parameters $\psi$ to explicitly enforce representation alignment. Concretely, they solve an alignment problem of the form
\begin{equation}
    \min_{\psi} \mathbb{E}_{x \sim \mathcal{D}_t} \| R(g(x; \theta_m); \psi) - g(x; \theta_t) \|_2,
\end{equation}
where $R(\cdot; \psi)$ is a learnable function applied to the merged representations. Although this approach can reduce the geometric discrepancy between representations, it does so by expanding the architecture with additional parameters. Importantly, this reliance on extra modules runs counter to the core ethos of model merging, which aims to obtain a unified backbone solely from existing fine-tuned weights, without extra storage (to keep $\psi$) or added inference cost.

In fact, such heavy-handed structural alignment may be unnecessary. Since predictions are governed by the composite map $h\circ g$, the merged encoder need not replicate the exact fine-tuned coordinates $g(x;\theta_t)$; it only needs to produce representations that remain compatible with the fixed head. In particular, if the discrepancy between $g(x;\theta_m)$ and $g(x;\theta_t)$ is well approximated by a simple geometry-preserving transform (e.g., a near-orthogonal rotation) that largely retains class structure, then performance can be recovered by re-aligning the head, rather than forcing the encoder back to the fine-tuned representation space via additional parameters.

\section{Geometric Analysis of Merging Discrepancy}
\label{sec:analysis}

Motivated by the observation that explicit representation matching via auxiliary modules may be overly strong and partially at odds with the goal of model merging to avoid extra storage, we now study the structure of the merging-induced discrepancy. Specifically, we investigate whether the difference between the merged representation $g(x;\theta_m)$ and the fine-tuned representation $g(x;\theta_t)$ can be explained by a geometry-preserving transform, in which case additional alignment parameters $\psi$ as in earlier work~\cite{pmlr-v235-yang24t,wei2025probsurgery} would be avoidable. To this end, we conduct a geometric analysis of merged versus fine-tuned representations. Our qualitative and quantitative results consistently suggest that the discrepancy manifests primarily as an orthogonal misalignment rather than an irreversible loss of task-relevant information.

\subsection{Visualization of Merging Representations}

We use t-SNE to visualize the representation space of the fine-tuned representations $g(\cdot; \theta_t)$ and the merged representations $g(\cdot; \theta_m)$ for several merging methods. 
To more clearly assess class-conditional separation,
we incorporate the ground-truth label information into the visualization process, as in Figure~\ref{fig:tsne-before}. More visualization in Appendix~\ref{sec:app:t-sne}.

The visualization reveals a distinct decoupling between classification accuracy and representational structure. As shown in Figure~\ref{fig:tsne-before}, even merged models with suboptimal performance exhibit well-separated clusters that mirror the geometry of the fine-tuned targets. This disconnect implies that the essential task-discriminative information remains intact after merging. Therefore, we hypothesize that the observed performance drops do not arise from a loss of intrinsic classification capability, but rather from a geometric drift where these otherwise healthy clusters have shifted away from the optimal regions defined by the static classifier heads $\phi_t$.

\begin{figure}[t]
  \centering
    \begin{tabular}{@{}c@{\hspace{6pt}}c@{}}
        \includegraphics[width=0.39\columnwidth]{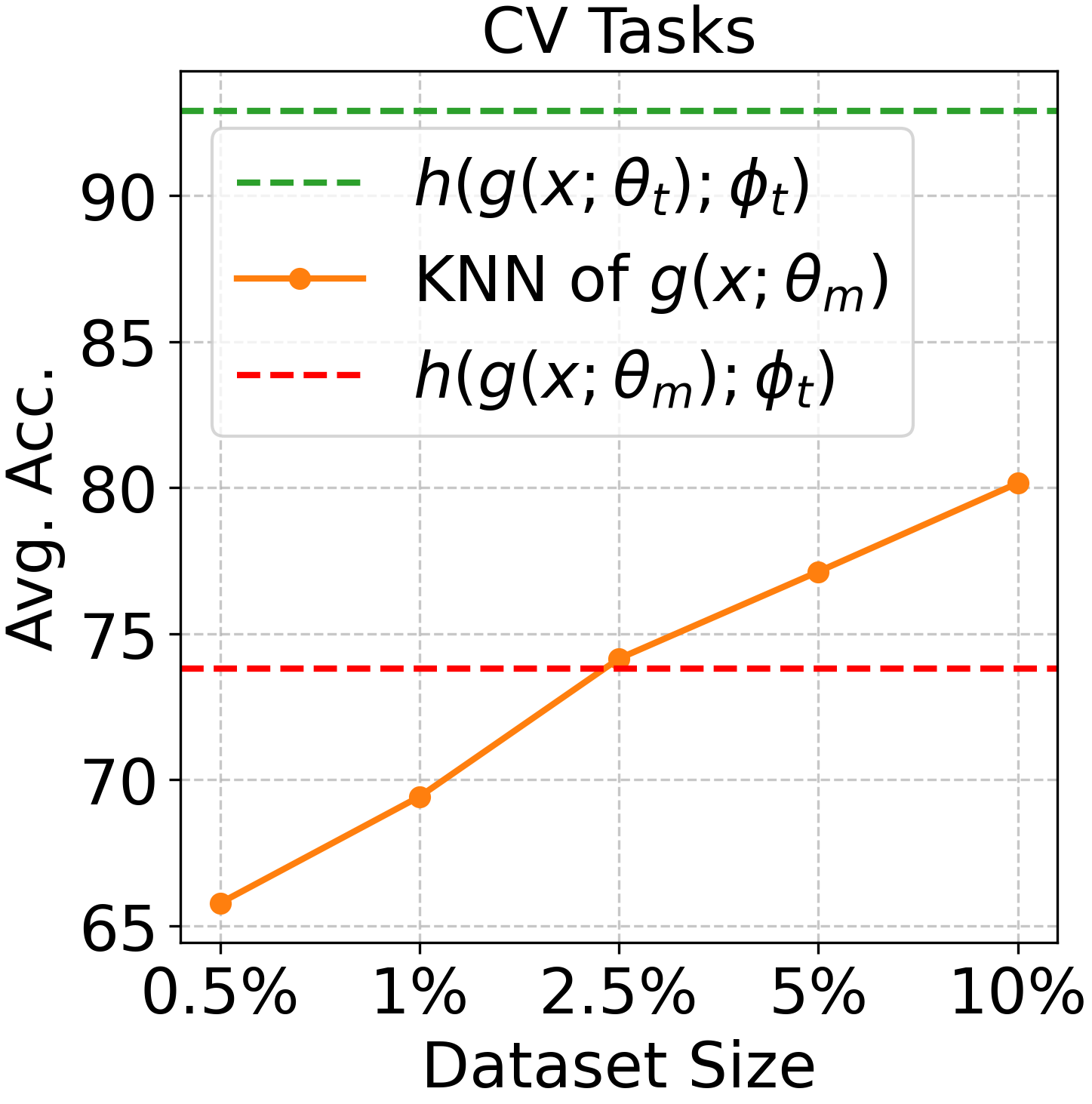} & \includegraphics[width=0.39\columnwidth]{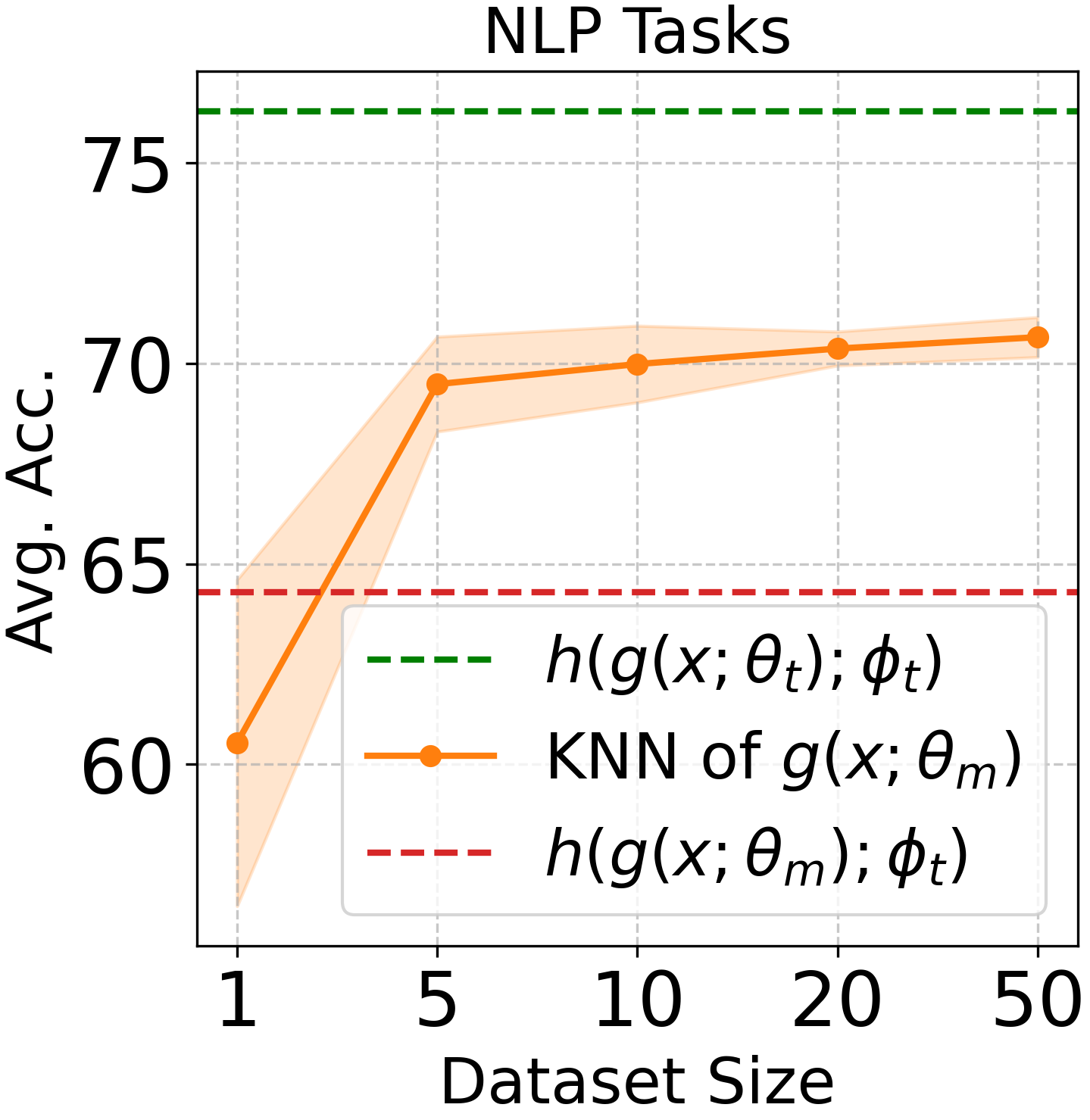} \\
    \end{tabular}
    \caption{Performance comparison of merging representations using KNN. We compare the upper-bound fine-tuned model ($h(g(x; \theta_t); \phi_t)$), the direct inference of the merged model constructed via Task Arithmetic ($h(g(x; \theta_m); \phi_t)$), and the KNN evaluation of the merging representations. The x-axis represents the number of support samples $K$ (or the dataset ratio for CV tasks with large class counts). Interestingly, KNN evaluation (orange line) significantly outperforms the direct inference baseline (red dashed line) with minimal support samples (e.g., 5-shot for NLP). This empirical evidence validates that the static classifier head significantly hinders the model's potential, rendering it inferior to a direct non-parametric evaluation of the representations.}
\label{fig:few-shot}
\end{figure}

\subsection{KNN Evaluation of Merging Representations}
\label{sec:knn}
To empirically verify whether the merging representations retain sufficient task-relevant information, despite the performance drop observed with the static classifier, we employ a \textit{K-Nearest Neighbors (KNN)} evaluation on $g(\cdot, \theta_m)$. Unlike the static classifier $\phi_t$, which requires the merging representations to be aligned with a fixed weight matrix, KNN is a non-parametric method that relies solely on the local geometry (Euclidean distances) of the representation space.

Specifically, for each task $t$, we construct a support set by sampling $K$ labeled examples per class from the dataset. A test sample $x$ is classified by assigning it to the class that minimizes the average distance to these support samples. Formally, let $\{x^{t,c}_i\}_{i=1}^K$ denote the set of $K$ support samples for class $c$ in task $t$. The prediction is formulated as:
\begin{equation}
    \begin{aligned}
        d_c(x) & = \frac{1}{K} \sum_{i=1}^{K} \| g(x;\theta_m) - g(x^{t,c}_i;\theta_m) \|_2, \\
        \hat{y} & = \arg \min_{c \in \mathcal{Y}} d_c(x).
    \end{aligned}
    \label{eq:knn}
\end{equation}
As illustrated in Figure~\ref{fig:few-shot}, the KNN evaluation effectively recovers the classification accuracy. This finding is pivotal: since a simple nearest-neighbor probe can successfully retrieve the correct labels, it confirms that the merged encoder $g(\cdot; \theta_m)$ preserves the task-specific discriminative information. Consequently, the standard performance degradation is not due to information loss in the representations, but is primarily attributable to the geometric misalignment between the encoder and the static classifier head.

\begin{table}[t]
    \centering
    \renewcommand{\arraystretch}{1.2}
    \resizebox{0.875\columnwidth}{!}{
    \begin{tabular}{l|l}
            \toprule
            \multicolumn{2}{c}{$G(\mathbf{x}) = \mathbf{A}\mathbf{x} + \mathbf{b}$, where $\mathbf{A} \in \mathbb{R}^{n \times n}, \mathbf{b} \in \mathbb{R}^{n}$} \\
            \toprule
            Transformation & Constraints \\
            \midrule
            Affine Transform & None \\
            Linear Transform & $\mathbf{b}=\mathbf{0}$ \\
            Invertible Linear Transform & $\det(\mathbf{A}) \neq 0, \mathbf{b} = \mathbf{0}$ \\
            Similarity Transform & $\mathbf{A}^\top \mathbf{A} = s\mathbf{I} ~ (s>0), \mathbf{b} = \mathbf{0}$ \\
            Orthogonal Transform & $\mathbf{A}^\top \mathbf{A} = \mathbf{I}, \mathbf{b}=\mathbf{0}$ \\
            Scaling Transform & $\mathbf{A}_{ij} = 0 ~ (\forall i \neq j), \mathbf{b} = \mathbf{0}$ \\
            \bottomrule
        \end{tabular}
    }
    \caption{Definitions of different transformation classes used in our analysis. Here, $\mathbf{x}$ denotes the merging representation and $G(\mathbf{x})$ denotes the target fine-tuned representation. The constraints impose different types of geometric regularization.}
    \label{tab:transformation}
\end{table}

\begin{figure*}[h]
    \centering
    \begin{tabular}{@{}c@{\hspace{2pt}}c@{}}
    \includegraphics[width=0.58\linewidth]{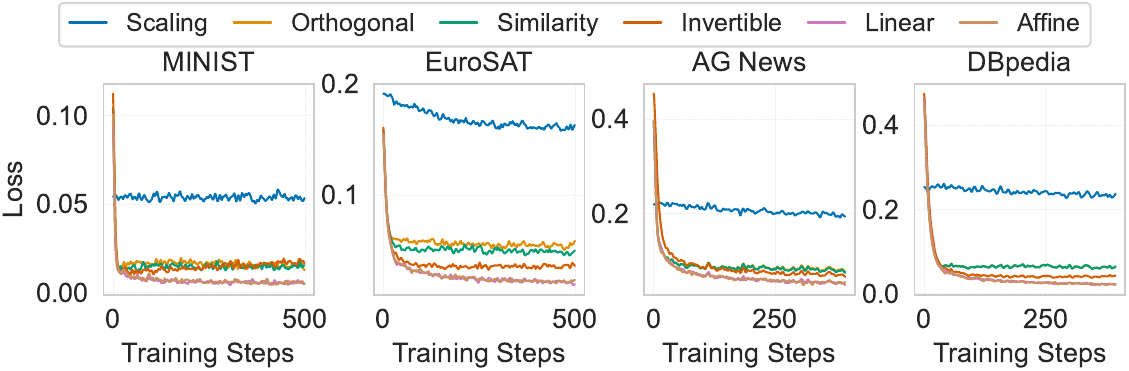} &
    \raisebox{-5pt}{\includegraphics[width=0.38\linewidth]{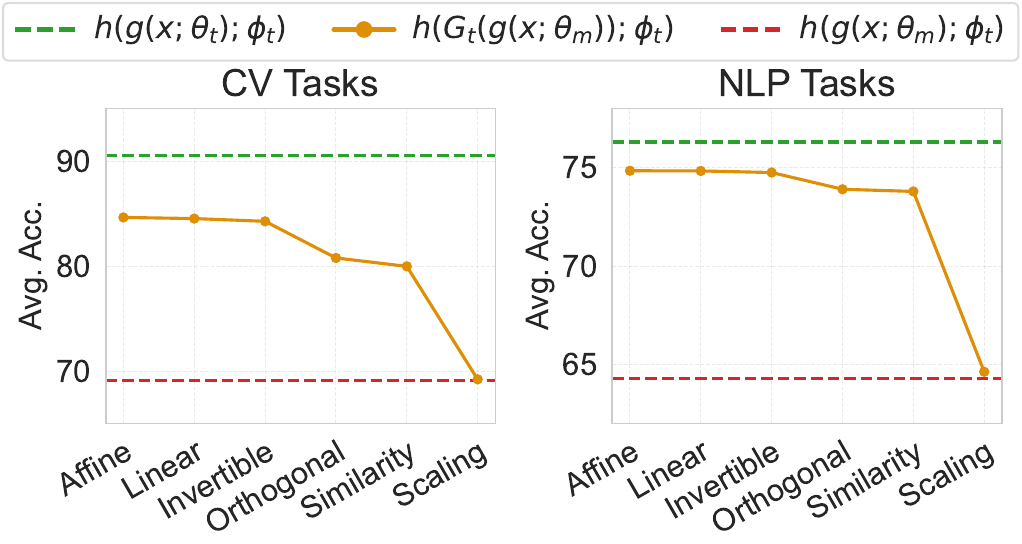}} \\
    (a) Training loss via different transformation. & (b) Performance via different transformation.
\end{tabular}
    \caption{Analysis of alignment trainging loss and recovered performance under varying transformation constraints. The transition to an orthogonal transformation yields the most significant performance leap, recovering a substantial portion of the accuracy. While further relaxing the constraints (to Invertible, Linear, or Affine) offers marginal additional gains, these non-rigid deformations provide negligible benefits relative to the increased degrees of freedom. This suggests that the misalignment is largely rotational, identifying orthogonal alignment as the optimal trade-off between performance recovery and geometric structural preservation.}
    \label{fig:loss}
\end{figure*}

\begin{figure*}[h]
    \centering
    \begin{tabular}{c@{\hspace{0.003\linewidth}}c}
        \includegraphics[width=0.68\linewidth]{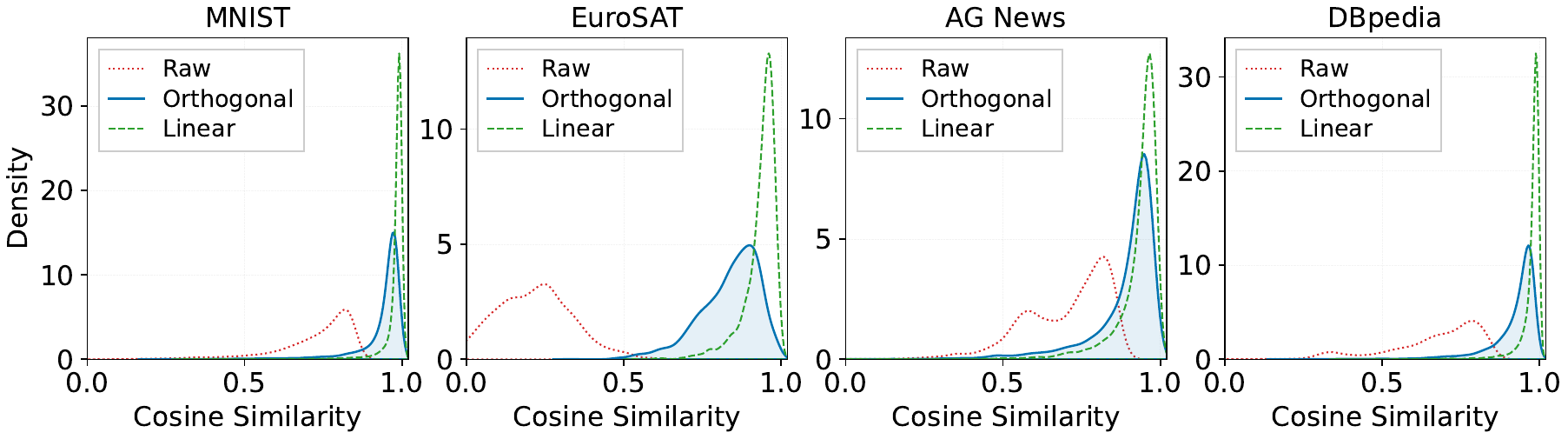} & 
        \raisebox{0.268cm}{\includegraphics[width=0.30\linewidth]{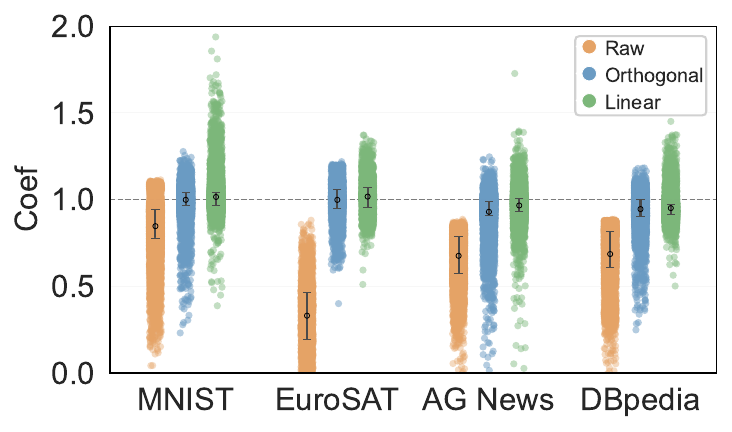}} \\
        (a) Kernel density estimation of cosine similarity distributions. & (b) Distribution of coefficients. \\
    \end{tabular}
    \caption{Quantitative Verification of Orthogonality. The legends Raw, Orthogonal, and Linear represent the alignment results between merging and fine-tuned representations under different constraints. (a) The orthogonal alignment drastically shifts the distribution towards 1.0, matching the unconstrained Linear alignment. (b) The orthogonal alignment naturally recovers the regression coefficients to approximately 1.0, indicating that the magnitude scale is inherently preserved. Collectively, these results demonstrate that the orthogonal transformation significantly reduces the geometric discrepancy between the merging and fine-tuned representations.}
    \label{fig:cos-coef}
\end{figure*}

\subsection{Hypothesis Validation via Transformation Alignment}
\label{sec:Re-evaluation}
The success of the KNN evaluation suggests that the merged representations and the fine-tuned representations share the same intrinsic geometry, yet differ in their spatial alignment. To characterize this misalignment, we hypothesize that there exists a specific transformation mapping the former to the latter. We validate this by formulating an alignment objective with varying degrees of constraints on the transformation function. Formally, we seek to align the merging representations $g(\cdot; \theta_m)$ with the fine-tuned representations $g(\cdot; \theta_t)$ by optimizing:
\begin{equation}
    \min_{G_t} \mathbb{E}_{x \sim \mathcal{D}_t} \| G_t(g(x; \theta_m)) - g(x; \theta_t) \|_2,
    \label{eq:align}
\end{equation}
where $G_t(\cdot)$ denotes the transformation function for task $t$, instantiated according to the families listed in Table~\ref{tab:transformation}. We employ unlabeled samples to optimize the parameters of $G_t$, while keeping the model parameters $\theta_m$ and $\theta_t$ frozen. This setup serves strictly as an analytical probe to determine the nature of the misalignment.

Taking the Orthogonal Transformation as a primary case study, we solve for the optimal rotation matrix $\mathbf{A}_t$ by minimizing the alignment error subject to an orthogonality regularization:
\begin{equation}
    \min_{\mathbf{A}_t} \sum_{x \in \mathcal{D}_t} \| \mathbf{A}_t g(x; \theta_m) - g(x; \theta_t) \|_2 + \alpha \| \mathbf{A}_t^\top \mathbf{A}_t - \mathbf{I} \|_2,
    \label{eq:objective_mlp_orth}
\end{equation}
where $\alpha$ is a hyperparameter balancing the alignment fidelity and the orthogonality condition. By tuning $\alpha$, we enforce $\mathbf{A}_t$ to converge towards a rigid rotation.

The empirical results of loss and performance via different transformations are shown in Figure~\ref{fig:loss}, corroborated by the distinct cluster recovery visualized in Figure~\ref{fig:tsne-after}, demonstrate that an orthogonal transformation is sufficient to effectively rectify the representation drift between merged and fine-tuned representations. Since orthogonal mappings are rigid transformations that preserve the intrinsic geometric structure, this finding supports our hypothesis that the merging representations retain necessary task-relevant information. Consequently, enforcing stricter representational equivalence (via Linear or Affine constraints) or introducing additional parameters and structures offers diminishing returns and is unnecessary for performance recovery.

\begin{table*}[h]
    \centering
    \resizebox{\linewidth}{!}{
    \begin{tabular}{l|cccccccc|c}
    \toprule
        \textbf{Method}  & \textbf{SUN397} & \textbf{Cars} & \textbf{RESISC45} & \textbf{EuroSAT} & \textbf{SVHN} & \textbf{GTSRB} & \textbf{MNIST} & \textbf{DTD} & \textbf{Avg.} \\
        \midrule
        Pre-trained Model &62.3 &59.7 &60.7 &45.5 &31.4 &32.6 &48.5 &43.8 &48.0  \\
        Fine-tuned Model & 75.3 & 77.7 & 96.1 & 99.7 & 97.5 & 98.7 & 99.7 & 79.4 & 90.5 \\
        Traditional MTL  &73.9 &74.4 &93.9 &98.2 &95.8 &98.9 &99.5 &77.9 &88.9  \\
        \midrule
        Fisher Merging~\citep{matena2022fishermerging}  &68.6 &69.2 &70.7 &66.4 &72.9 &51.1 &87.9 &59.9 &68.3  \\
        RegMean~\citep{jin2022dataless_regmean} &65.3 &63.5 &75.6 &78.6 &78.1 &67.4 &93.7 &52.0 &71.8  \\
        Concrete TA~\citep{tang2023concrete} &62.5 &61.1 &76.0 &95.7 &91.0 &81.9 &98.5 &51.9 &77.3  \\
        Concrete AM~\citep{tang2023concrete}&67.8 &70.0 &87.5 &96.0 &91.6 &96.7 &98.7 &63.8 &84.0  \\
        TW AdaMerging~\citep{yangadamerging}&58.0 &53.2 &68.8 &85.7 &81.1 &84.4 &92.4 &44.8 &71.1  \\
        \midrule
        Weight Averaging & 65.3 & 63.4 & 71.4 & 71.7 & 64.2 & 52.8 & 87.5 & 50.1 & 65.8 \\
        \quad w/ Surgery~\citep{pmlr-v235-yang24t} &67.6 &64.6 &85.8 &96.8 &76.9 &82.9 &97.8 &67.3 &80.0 \\
        \quad w/ ProbSurgery~\citep{wei2025probsurgery} &70.7 & 70.1 & 94.0 & 99.6 & 83.4 & 98.7 & 99.3 & 78.1 & \textbf{86.7} \\
        \rowcolor{gray!15} \quad w/ MOMA &66.6 & 53.5 & 87.5 & 97.6 & 90.6 & 89.2 & 97.1 & 71.8 & 81.7 \\
        \midrule
        Task Arithmetic~\citep{ilharco2022editing_task_arithmetic} & 55.2 & 54.9 & 66.7 & 78.9 & 80.2 & 69.7 & 97.3 & 50.4 & 69.1 \\
        \quad w/ Surgery~\citep{pmlr-v235-yang24t} &63.8 & 59.9 & 83.3 & 97.9 &  87.0 & 87.0 & 98.6 & 69.4 & 80.9 \\
        \quad w/ ProbSurgery~\citep{wei2025probsurgery} &67.0 & 67.0 & 94.1 & 99.8 &  91.2 & 98.8 & 99.4 & 79.0 & 87.0  \\
        \rowcolor{gray!15} \quad w/ MOMA &72.7 & 70.0 & 93.5 & 99.4 &  96.7 & 96.6 & 99.5 & 77.3 & \textbf{88.2} \\
        \midrule
        Ties~\citep{yadav2024ties}& 65.0 & 64.4 & 74.8 & 77.4 & 81.2 & 69.3 & 96.5 & 54.5 & 72.9 \\
        \quad w/ Surgery~\citep{pmlr-v235-yang24t}&69.8 & 66.1 & 87.3 & 97.5 & 86.7 & 87.6 & 98.5 & 71.6 & 83.1 \\
        \quad w/ ProbSurgery~\citep{wei2025probsurgery} &71.5 & 70.6 & 94.4 & 99.7 & 90.6 & 98.9 & 99.4 & 78.9 & 88.0 \\
        \rowcolor{gray!15} \quad w/ MOMA &72.6 &69.7 &93.5 &99.3 &96.6 &96.8 &99.4 &77.1 &\textbf{88.1} \\
    \bottomrule    
    \end{tabular}
    }
    \caption{Comparison on CV datasets via ViT-B/32. MOMA achieves state-of-the-art (SOTA) performance, particularly with Task Arithmetic and Ties. Although ProbSurgery holds a slight edge on Weight Averaging, MOMA matches or exceeds this strict alignment method without introducing any additional parameters or added inference cost. Results on ViT-B/16 are shown in Appendix~\ref{sec:app:vit-16}.}
    \label{tab:main-results-cv}
\end{table*}

\subsection{Quantitative Verification of Orthogonality}

To quantitatively verify our hypothesis that the discrepancy between the merging representations and the fine-tuned representations is nearly an orthogonal transformation, we employ a metric analysis following \citet{zhouemergence}. Specifically, we examine two geometric indicators: cosine similarity to assess directional alignment, and the coefficient to quantify magnitude scaling consistency.
We compute these metrics between the fine-tuned representations $g(x; \theta_t)$ and merging representations $g(x; \theta_m)$ under three distinct alignment conditions: (i) \textit{Raw}, representing the original merging representations; (ii) \textit{Orthogonal}, aligned via the optimal orthogonal transformation derived in Eq.~\ref{eq:objective_mlp_orth}; and (iii) \textit{Linear}, aligned via the flexible unconstrained linear transformation. Formally, for a pair of representation vectors $u$ and $v$, these metrics are defined as $\mathrm{cos}(u, v) = u^\top v / (\|u\| \|v\|)$ and $\mathrm{coef}(u, v) = u^\top v / \|v\|^2$.

As visualized in Figure~\ref{fig:cos-coef}, the gap between merged and fine-tuned representations can be almost rectified by a simple rotation. This empirical evidence supports our hypothesis: the representational drift is primarily a rigid geometric transformation, leaving the intrinsic feature structure intact.

\section{Masked Orthogonal Matrix Alignment}
Since the core objective of model merging is to recover the predictive function (i.e., output logits) of fine-tuned models rather than strictly reconstructing their representation, rectifying this orthogonal misalignment is sufficient.
Crucially, the linearity of the classifier heads allows us to absorb this orthogonality directly into the classifier parameters, thereby avoiding any architectural expansion.

Specifically, let the classifier head for task $t$ be denoted as $h(\cdot; \phi_t)$, parameterized by weights $\mathbf{W}_t$, such that the logit calculation is governed by the inner product between the representation and the weights. Applying an orthogonal transformation $\mathbf{A}_t$ to the merging representations $g(x; \theta_m)$ modifies this operation to $\langle \mathbf{A}_t g(x; \theta_m), \mathbf{W}_t \rangle$. By the inner-product-preserving property of orthogonal transformation, this is mathematically equivalent to:
\begin{equation}
    \langle \mathbf{A}_t g(x; \theta_m), \mathbf{W}_t \rangle = \langle g(x; \theta_m), \mathbf{A}_t^\top \mathbf{W}_t \rangle.
\end{equation}
Consequently, we can replace the original weights $\mathbf{W}_t$ with the aligned weights $\mathbf{A}_t^\top \mathbf{W}_t$. This operation achieves the desired alignment without introducing any additional parameters or computational overhead during inference.

While orthogonal alignment is theoretically sound, enforcing such a rigid geometric constraint can be optimizationally challenging and prone to underfitting. Prior studies~\cite{wanglocalizing} have demonstrated that aggregated multi-task vectors can be effectively pruned without compromising representation capability. Motivated by this, we introduce a pruning mechanism to the aggregated multi-task vector. By selectively masking the parameters, we increase the degrees of freedom within the optimization landscape, thereby facilitating the convergence of the orthogonal transformation.

We define the aggregated multi-task vector as $\tau_{\text{agg}} = \sum_{t=1}^{T}\tau_t$, and learn a mask $\mathbf{M}$,
yielding the pruned task vector $\tau_m = \mathbf{M} \odot \tau_{\text{agg}}$. To enable gradient-based optimization for the binary mask, we employ a sigmoid reparameterization strategy. Specifically, we optimize a real-valued parameter $\mathbf{S}$ and derive the mask via $\mathbf{M} = \sigma(\mathbf{S})$. 
The joint optimization objective is formulated as:
\begin{align}
     \min_{\{\mathbf{A}_t\}_{t=1}^T, \mathbf{S}} \! & \sum_{t=1}^T \mathbb{E}_{x \sim \mathcal{D}_t} \|\mathbf{A}_t g(x; \theta_b \!  +  \! \lambda \sigma(\mathbf{S}) \!  \odot \! \tau_{\text{agg}}) \! - \! g(x; \theta_t)\|_2 \notag\\
    & + \alpha \sum_{t=1}^T \|\mathbf{A}_t^\top\mathbf{A}_t - \mathbf{I}\|_2,
    \label{eq:objective}
\end{align}
where $\odot$ denotes the Hadamard product.
Finally, the learned mask is absorbed in $\tau_m$, and the orthogonal matrices $\mathbf{A}_t$ are absorbed into their respective task classifier heads $\mathbf{W}_t$ via matrix multiplication.
Consequently, our method incurs zero additional parameter storage overhead at deployment.

We also explore an evaluation protocol grounded in our orthogonality analysis, where fine-tuning aligned classifier heads on few-shot data proves sufficient to bridge the performance gap (see Appendix~\ref{app:sec:aligned-c}).

\begin{table}[h]
    \centering
    \resizebox{\linewidth}{!}{
    \begin{tabular}{@{}l@{\hspace{5pt}}c@{\hspace{5pt}}c@{\hspace{5pt}}c@{\hspace{5pt}}c@{\hspace{5pt}}c@{\hspace{5pt}}c@{}}
    \toprule
        \textbf{Method}  & \textbf{AG News} & \textbf{Yelp} & \textbf{Amazon} & \textbf{Yahoo} & \textbf{DBpedia} & \textbf{Avg.} \\
        \midrule
        Pre-trained Model &24.9 &19.5 &19.7 &9.9 &8.8 &16.5 \\
        Fine-tuned Model & 91.6  & 60.4  & 57.9  & 72.7  & 98.8  & 76.3 \\
        Traditional MTL & 90.6 & 59.1 & 55.6 & 71.3 & 98.5 & 75.0 \\
        \midrule
        Weight Averaging  & 79.2  & 49.8  & 45.0  & 50.3  & 55.1  & 55.8 \\
        \quad w/ Surgery &90.3 &58.0 &54.2 &70.8 &98.4 &74.3 \\
        \quad w/ ProbSurgery &91.4 &62.1 &55.8 &72.4 &98.1 &\textbf{76.0} \\
        \rowcolor{gray!15} \quad w/ MOMA &88.8 &58.0 &54.0 &68.3 &98.8 &73.6 \\
        \midrule
        Task Arithmetic & 82.9 & 55.8 & 48.4 & 53.1 & 81.5 & 64.3 \\
        \quad w/ Surgery &89.8 &58.4 &55.4 &70.3 &98.0 &74.4 \\
        \quad w/ ProbSurgery &90.7 &60.4 &56.4 &72.3 &98.7 &75.7 \\
        \rowcolor{gray!15} \quad w/ MOMA &91.5 &60.8 &57.7 &72.9 &98.8 &\textbf{76.3} \\
        \midrule
        Ties & 83.3  & 55.7  & 47.0  & 55.9  & 83.4  & 65.1 \\
        \quad w/ Surgery & 90.4 & 58.8 & 56.5 & 71.5 & 98.7 & 75.2 \\
        \quad w/ ProbSurgery & 90.9 & 59.6 & 56.3 & 71.7 & 98.8 & 75.5 \\
        \rowcolor{gray!15} \quad w/ MOMA &91.4 &60.7 &57.7 &73.0 &98.9 &\textbf{76.3} \\
    \bottomrule
    \end{tabular}
    }
    \caption{Comparison on NLP tasks. The results exhibit trends consistent with the CV tasks in Table~\ref{tab:main-results-cv}.}
    \label{tab:main-results-nlp}
\end{table}

\section{Experiments}
\label{sec:Experiments}
\subsection{Experimental Setting}
\paragraph{Datasets}
Following~\cite{pmlr-v235-yang24t,wei2025probsurgery}, we evaluate on eight image and five text classification datasets. Details are in Appendix~\ref{sec:app:data}.

\paragraph{Comparable Baselines} We compare MOMA against specialized methods like Fisher Merging~\cite{{matena2022fishermerging}} and TW AdaMerging ~\cite{yangadamerging}. 
We also evaluate MOMA integrated into Weight Averaging, Task Arithmetic, and Ties, benchmarking against alignment strategies like Surgery~\cite{pmlr-v235-yang24t} and ProbSurgery~\cite{wei2025probsurgery}. See Appendix~\ref{sec:app:baseline} for details.

\paragraph{Implementation Details}
We use ViT-B/16, 32~\cite{radford2021learning} for CV and bert-base-uncased~\cite{devlin2018bert} for NLP, adopting fine-tuned checkpoints from~\cite{pmlr-v235-yang24t}. All experiments are conducted on NVIDIA A100 GPUs. See Appendix~\ref{sec:app: implementation} for more details.

\subsection{Main Results}
Main Results are shown in Table~\ref{tab:main-results-cv} and Table~\ref{tab:main-results-nlp}. 
MOMA surpasses complex competitors like Surgery and ProbSurgery to achieve SOTA accuracy.
By doing so with zero additional parameters, we validate that rigid orthogonal alignment is sufficient for effective model merging.

\begin{table}[h]
\centering
\label{tab:comparison}
\resizebox{0.85\linewidth}{!}{
\begin{tabular}{l|c|c}
\toprule
\textbf{Method} & \textbf{CV (Avg.)} & \textbf{NLP (Avg.)} \\
\midrule
Task Arithmetic & 69.1 & 64.3 \\
\quad w/ Orthogonal & 76.4 & 72.3 \\
\quad w/ Orthogonal + Masking & 88.2 & 76.3 \\
\quad w/ Linear + Masking & 89.9 & 76.5 \\
\bottomrule
\end{tabular}
}
\caption{Ablation study of MOMA and constraint analysis. Notably, ``Orthogonal'' and ``Linear'' represent the geometric transformation constraints applied to the alignment, while ``Orthogonal + Masking'' constitutes our proposed MOMA framework.}
\label{tab:ablation}
\end{table}
\subsection{Ablation Study}
Table~\ref{tab:ablation} details the contribution of each component.
Introducing Orthogonal transformation significantly boosts the baseline ($69.1\% \to 76.4\%$ on CV), while the full MOMA framework (adding Masking) yields a dramatic leap to $88.2\%$ (CV) and $76.3\%$ (NLP). To probe the performance limit, we evaluate the looser Linear transformation (which allows non-rigid distortion). While it achieves slightly higher accuracy, the marginal gain is negligible compared to MOMA's massive improvement over the baseline. This empirically verifies that the degradation is driven predominantly by rotational drift, making MOMA's orthogonal alignment sufficient to recover the majority of performance.

\subsection{Representation Learned by MOMA}
\begin{figure}[h]
    \centering
    \begin{tabular}{@{}c@{\hspace{2pt}}c@{}}
        \includegraphics[width=0.49\columnwidth]{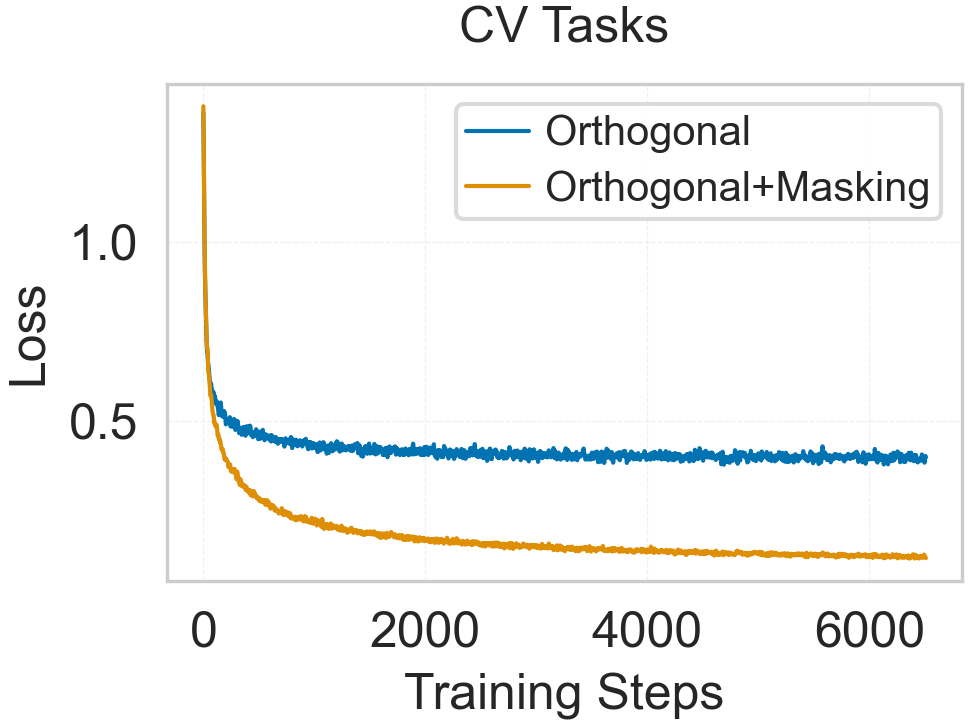} & \includegraphics[width=0.49\columnwidth]{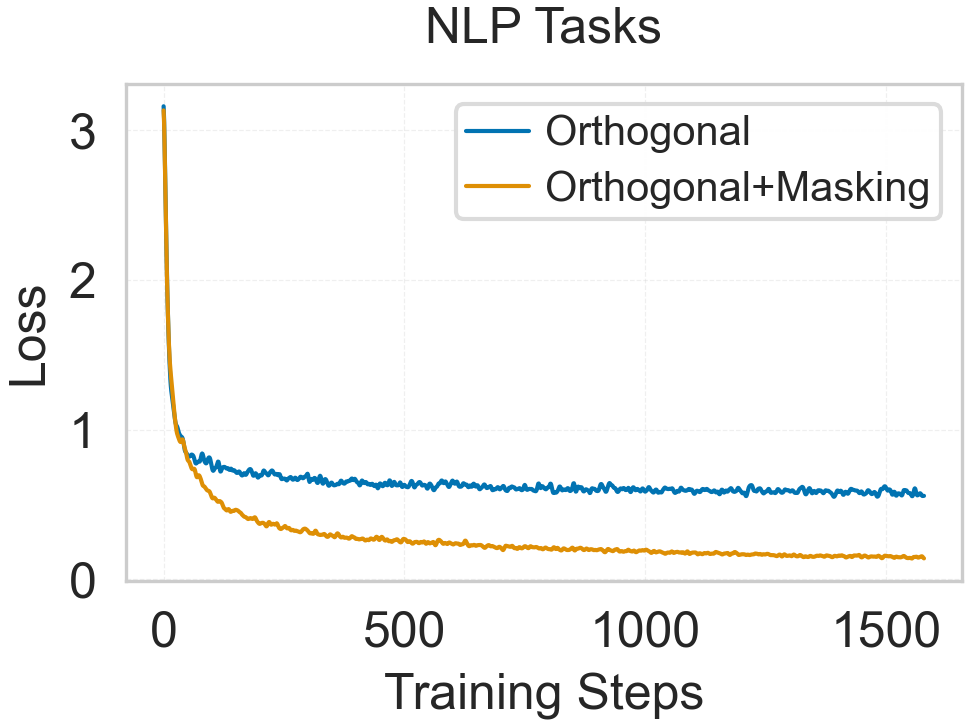} \\
    \end{tabular}
    \caption{Training loss of MOMA (Orthogonal + Masking).}
    \label{fig:loss MOMA}
\end{figure}

To better understand the mechanism behind MOMA's performance, we analyze the training loss and the geometric properties of the aligned representations.
\paragraph{Loss of MOMA}
Figure~\ref{fig:loss MOMA} visualizes the alignment loss during training. 
Compared to using Orthogonal alone, the full MOMA converges to a significantly lower loss value on both CV and NLP tasks. This indicates that by masking non-essential weights, MOMA effectively relaxes the optimization landscape, granting the transformation the necessary degrees of freedom to achieve a much tighter and more accurate fit.

\begin{figure}[h]
    \centering
    \begin{tabular}{c@{\hspace{0.0015\linewidth}}c}
        \includegraphics[width=0.60\linewidth]{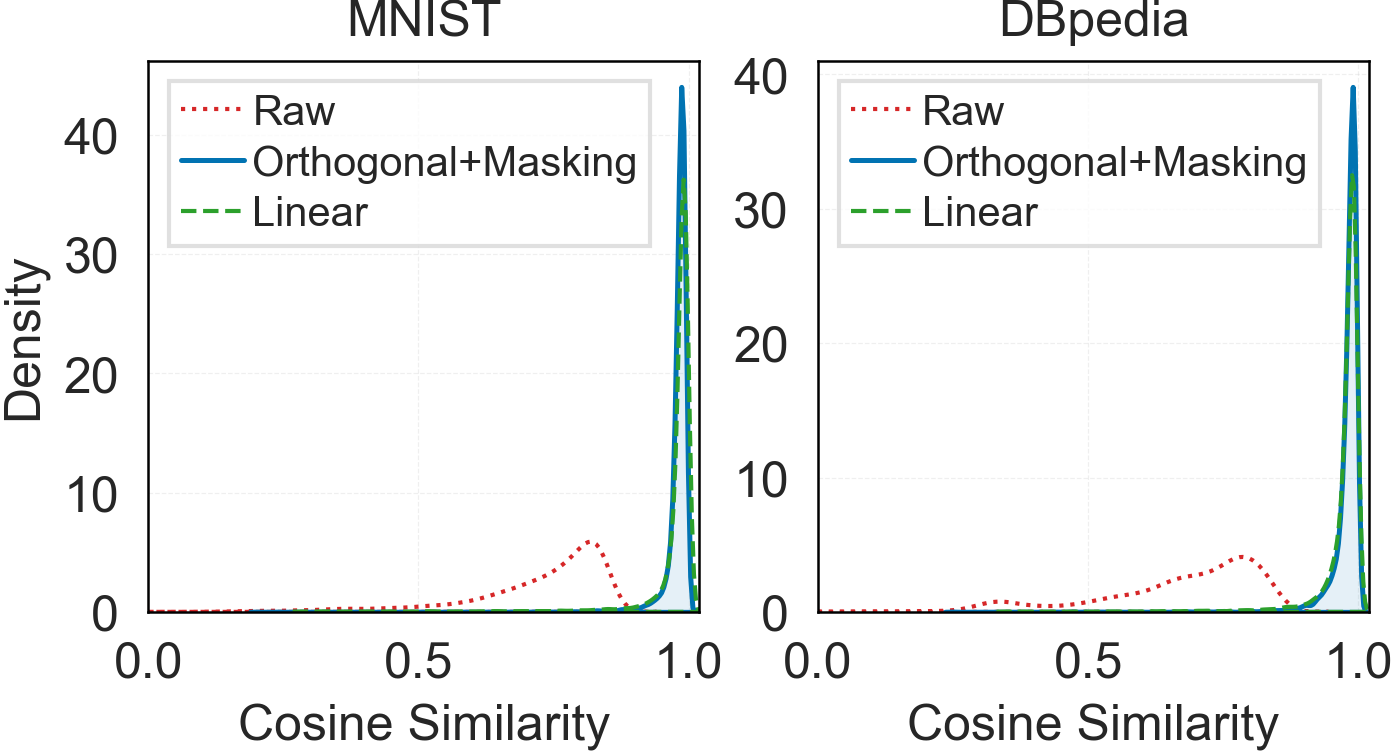} & 
        \raisebox{0.11cm}{\includegraphics[width=0.37\linewidth]{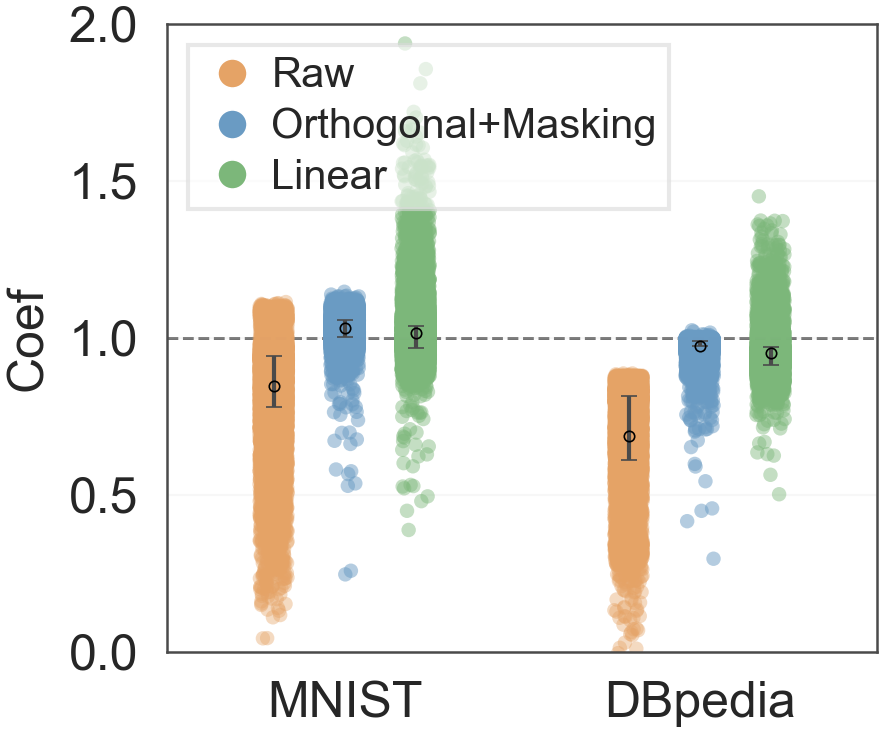}} \\
    \end{tabular}
    \caption{Geometric analysis of the learned representations.}
\label{fig:moma_geometry}
    \label{fig:x}
\end{figure}

\paragraph{Geometric Analysis}
Figure~\ref{fig:moma_geometry} further evaluates the geometry of the aligned representations. 
As shown in Figure~\ref{fig:moma_geometry} (a), the cosine similarity distribution of MOMA matches that of the Linear alignment (the empirical upper bound) almost exactly. 
Similarly, Figure~\ref{fig:moma_geometry} (b) demonstrates that MOMA recovers the regression coefficients to unity, correcting the scale shift observed in the \textit{Raw} baseline.
These results validate that MOMA attains the high geometric precision of non-rigid transformations (Linear) while strictly adhering to the parameter-efficiency and structural preservation of rigid orthogonal mappings.

\subsection{Parameter Analysis}
\begin{figure}[!ht]
    \centering
    \setlength{\tabcolsep}{2pt}  
    \begin{tabular}{cc}
        \includegraphics[width=0.49\linewidth]{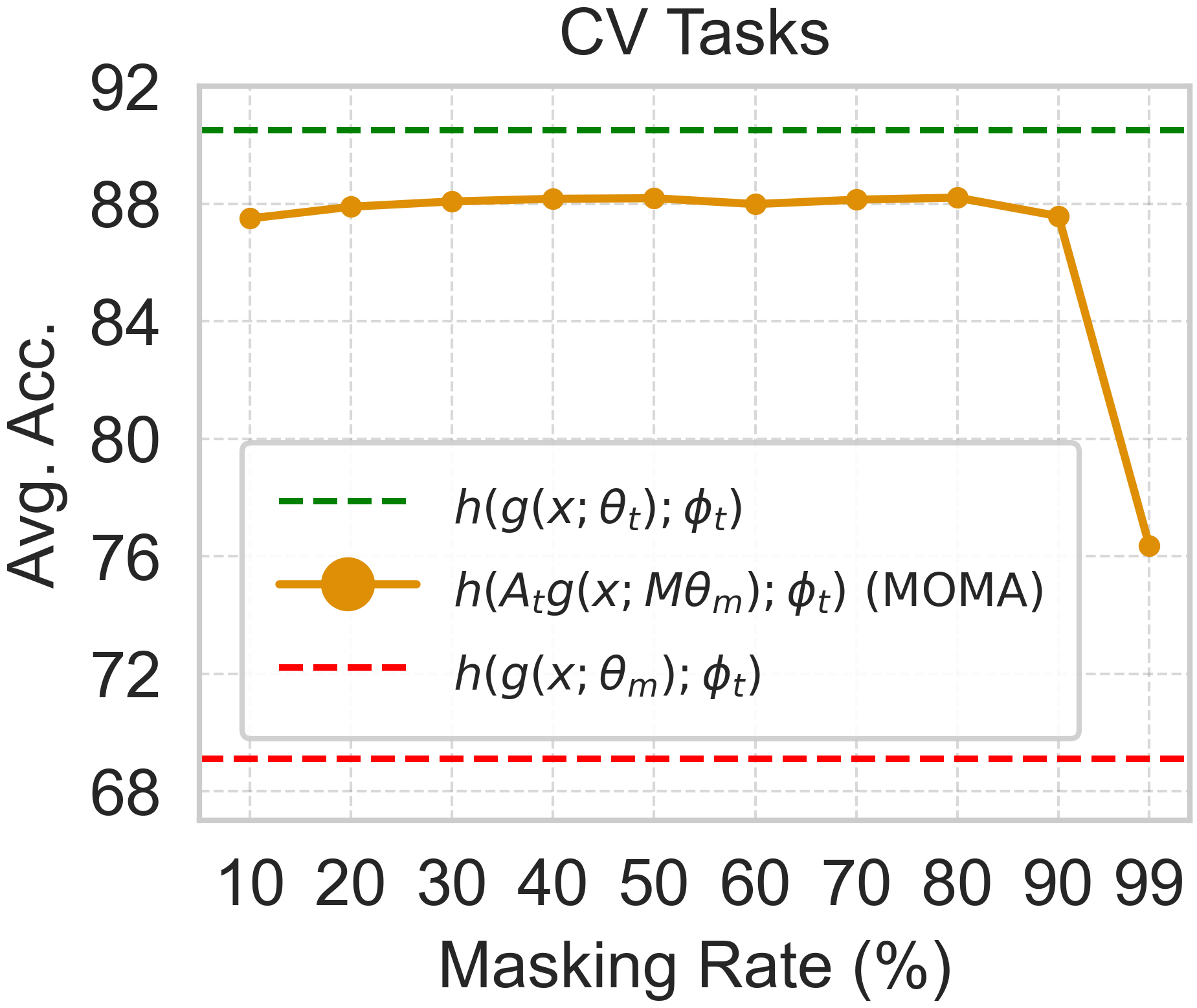} & 
        \includegraphics[width=0.49\linewidth]{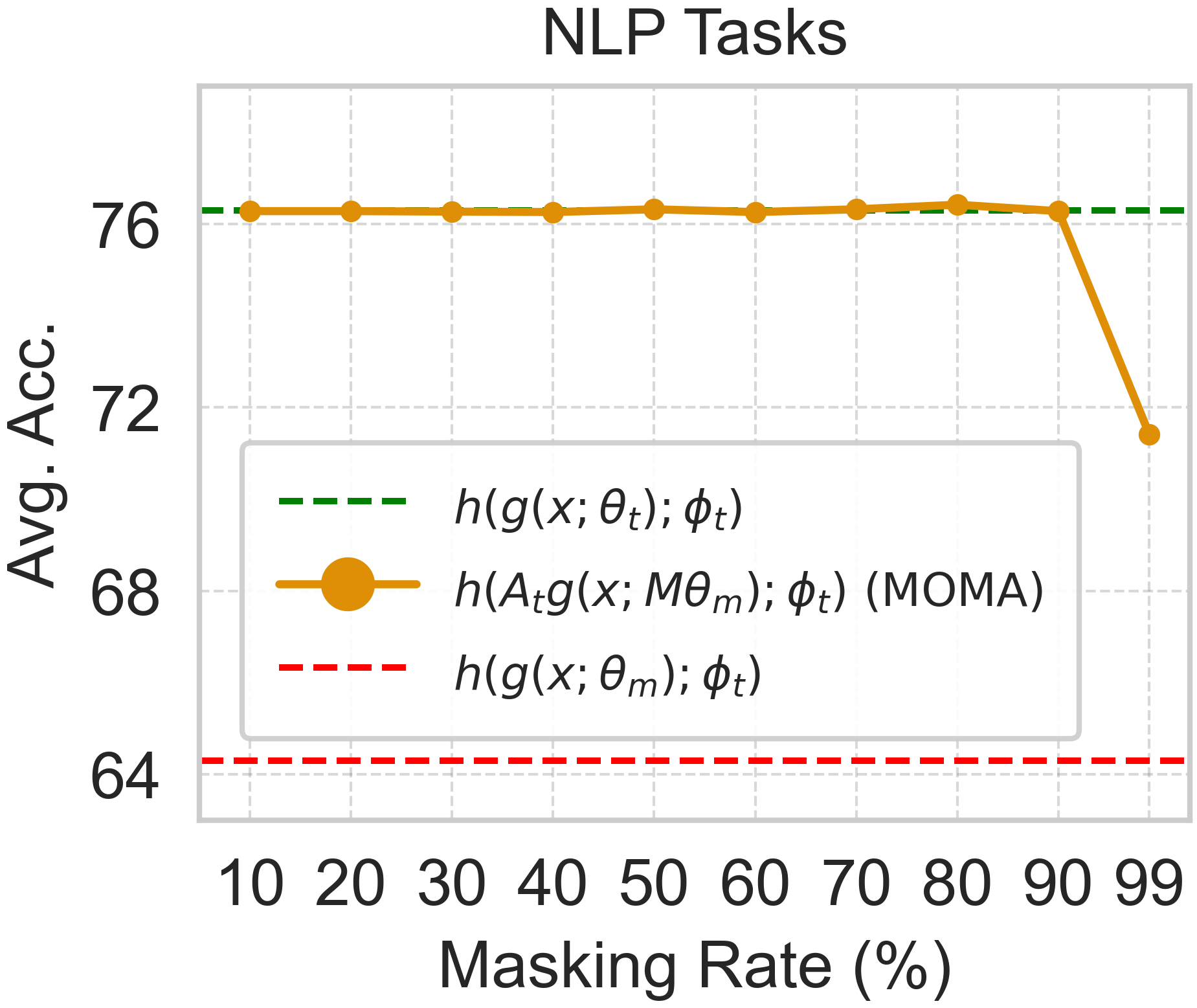} \\
    \end{tabular}
    \caption{Sensitivity analysis of the masking rate.}
    \label{fig:masking_sensitivity}
\end{figure}

\paragraph{Masking Rate}
Figure~\ref{fig:masking_sensitivity} illustrates the test accuracy via the masking rate.
MOMA exhibits remarkable robustness across a wide range of sparsity levels (10\%--90\%) on both CV and NLP tasks. The performance remains stable and consistently matches the fine-tuned upper bound, only degrading under extreme sparsity ($>90\%$). This observation aligns with the findings in DARE~\cite{yu2024language_dare}, which demonstrated that removing up to 90\% of delta parameters results in minimal performance degradation. 

\begin{table}[!ht]
\centering
\resizebox{0.85\linewidth}{!}{
\begin{tabular}{lc|c|c}
\toprule
\textbf{Method} & \textbf{Ratio} & \textbf{CV (Avg.)} & \textbf{NLP (Avg.)} \\
\midrule
Task Arithmetic & - & 69.1 & 64.3 \\
\quad w/ MOMA & 10\% & 83.2 & 75.0 \\
\quad w/ MOMA & 50\% & 87.3 & 76.1 \\
\quad w/ MOMA & 100\% & 88.2 & 76.3 \\
\bottomrule
\end{tabular}
}
\caption{Data efficiency analysis. We evaluate MOMA using varying ratios of the unlabeled test dataset for alignment.}
\label{tab:data_efficiency}
\end{table}
\paragraph{Data Efficiency}
Table~\ref{tab:data_efficiency} examines the impact of available data size on orthogonal alignment.
Empirically, using just 10\% of the test set is sufficient to boost CV performance from the baseline $69.1\%$ to $83.2\%$.
While increasing the data ratio to $100\%$ yields further improvements ($88.2\%$), the ability to achieve SOTA results with minimal data confirms that MOMA is well-suited for scenarios with limited access to test samples.

\section{Related Works}
\label{sec:related}
Model merging combines homologous models into a single backbone without accessing training data~\cite{zheng2025decouple,jang2024model_Stock,daheim2023model_uncertainty,panigrahi2023task,marczak2024magmax,ortiz2024task,yangadamerging}. In this work, we analyze several representative model merging methods.

Foundational methods like Weight Averaging~\cite{wortsman2022model_model_soup} and Task Arithmetic~\cite{ilharco2022editing_task_arithmetic} manipulate parameters directly via averaging or task vectors. To improve upon uniform merging, subsequent works optimize coefficients~\cite{qiu2025superpose,zhang2025variational,dai2025leveraging}. Fisher Merging~\cite{matena2022fishermerging} weighs parameters by information content, while RegMean~\cite{jin2022dataless_regmean} and AdaMerging~\cite{yangadamerging} minimize prediction divergence, sometimes utilizing unlabeled data.

To alleviate parameter collision, recent studies propose sparsifying task vectors~\cite{miyano2025adaptive,zimmer2023sparse,marczak2024magmax}. Ties~\cite{yadav2024ties} resolves sign conflicts and prunes redundancy, while DARE~\cite{yu2024language_dare} employs random dropping to reduce interference while preserving task knowledge. TALL-masks~\cite{wanglocalizing} and Model Tailor~\cite{zhumodel} further refine this by learning masks to localize task-relevant information.

To address the limitations of parameter-space operations, Surgery~\cite{pmlr-v235-yang24t} and ProbSurgery~\cite{wei2025probsurgery} seek to align models in the representation space. These methods introduce additional trainable parameters to explicitly minimize the distance between merged and fine-tuned representations.

\section{Conclusion}
This paper presents MOMA, a model merging framework that addresses the geometric misalignment of merging representation and task-specific classifier. By identifying that the misalignment is primarily due to an orthogonal transformation, we devised a method to rectify this drift via learnable orthogonal matrices and a mask matrix. Leveraging the mathematical properties of orthogonal transformations, MOMA absorbs these new parameters into the existing model weights without performance loss. Our approach achieves superior results compared to existing baselines while maintaining the core advantages of model merging: zero additional parameters and zero added inference cost.



\bibliography{custom}
\bibliographystyle{icml2026}

\appendix

\onecolumn
\section*{Appendix}
The Appendix is organized as follows:
\begin{itemize}
    \item Appendix A presents t-SNE visualizations of the merged representations. The results demonstrate that representations from different merging methods all form distinct clusters, indicating that they effectively preserve task-relevant discriminative information.

    \item Appendix B provides supplementary details regarding the experimental setup and implementation.

    \item Appendix C reports the performance of MOMA on the ViT-B/16 backbone. The results confirm that MOMA matches or surpasses restricted alignment methods (e.g., Surgery and ProbSurgery) while maintaining a zero-parameter overhead.

    \item Appendix D introduces Aligned-C, an evaluation protocol that mitigates misalignment by fine-tuning classifier heads on few-shot data. Results demonstrate that even with minimal examples, Aligned-C significantly outperforms standard evaluation baselines.
\end{itemize}

\section{Visualization of Merging Representation}
\label{sec:app:t-sne}
\begin{figure*}[h]
    \centering
    \begin{tabular}{@{}c@{\hspace{2pt}}c@{\hspace{2pt}}c@{\hspace{2pt}}c@{\hspace{2pt}}c@{\hspace{2pt}}c@{}}
        \includegraphics[width=0.16\linewidth]{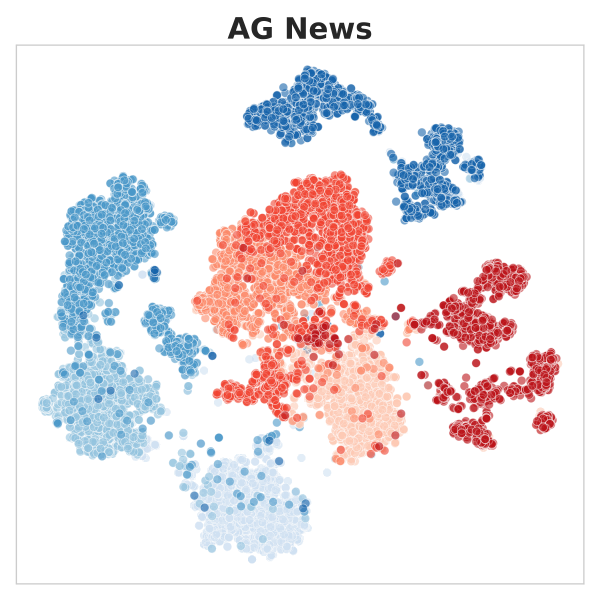} &\includegraphics[width=0.16\linewidth]{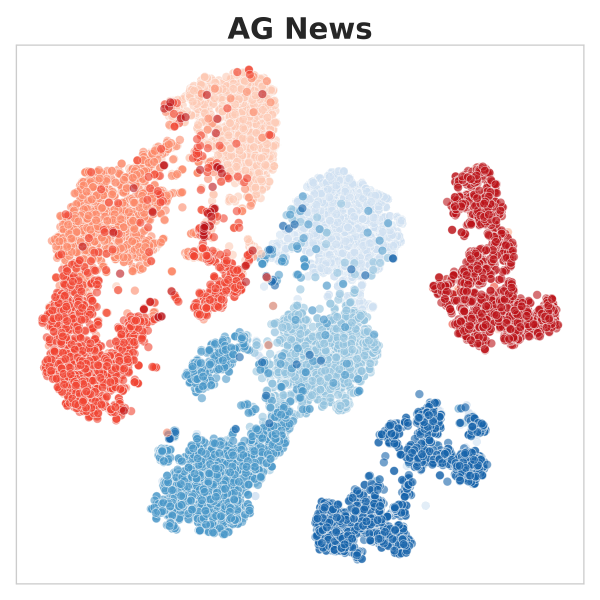} &  \includegraphics[width=0.16\linewidth]{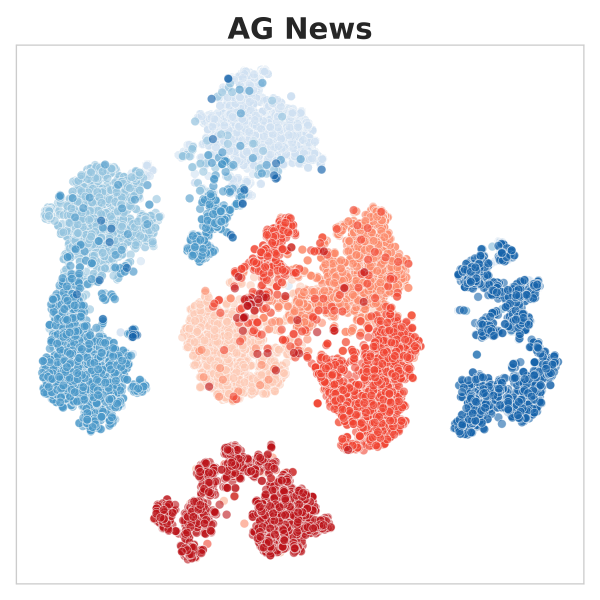} &
        \includegraphics[width=0.16\linewidth]{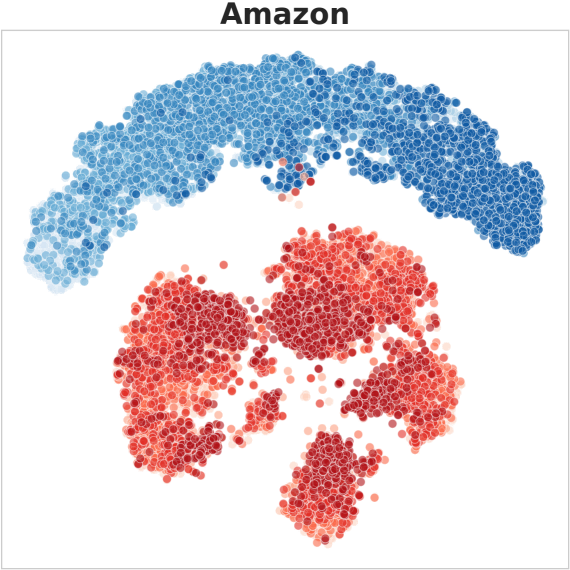} &\includegraphics[width=0.16\linewidth]{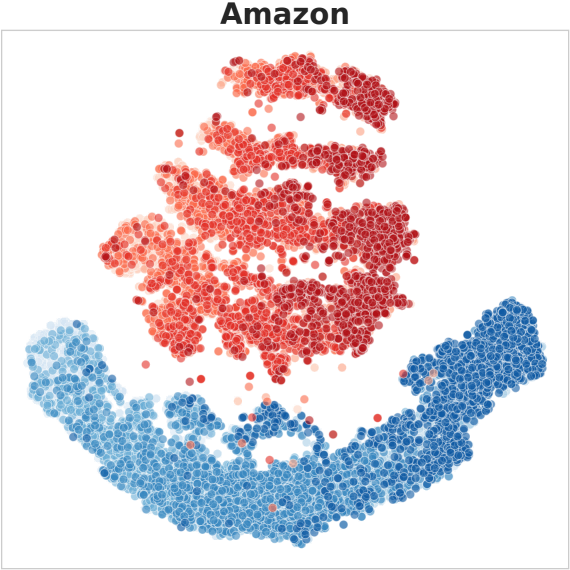} &  \includegraphics[width=0.16\linewidth]{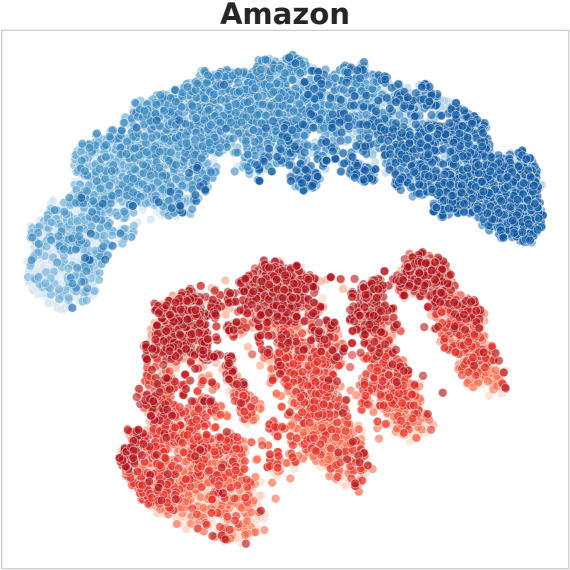} \\
        Weight Averaging & Task Arithmetic & Ties  & Weight Averaging & Task Arithmetic & Ties\\
         \includegraphics[width=0.16\linewidth]{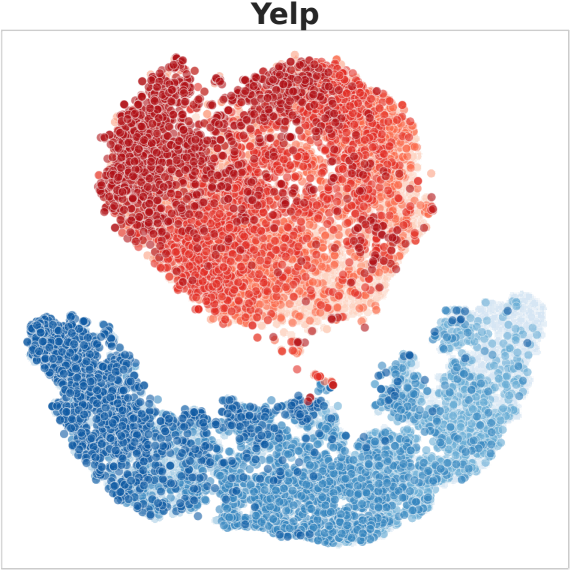} &\includegraphics[width=0.16\linewidth]{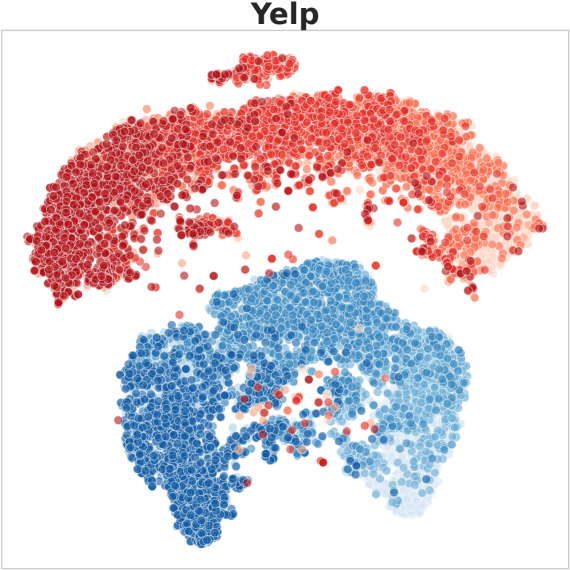} &  \includegraphics[width=0.16\linewidth]{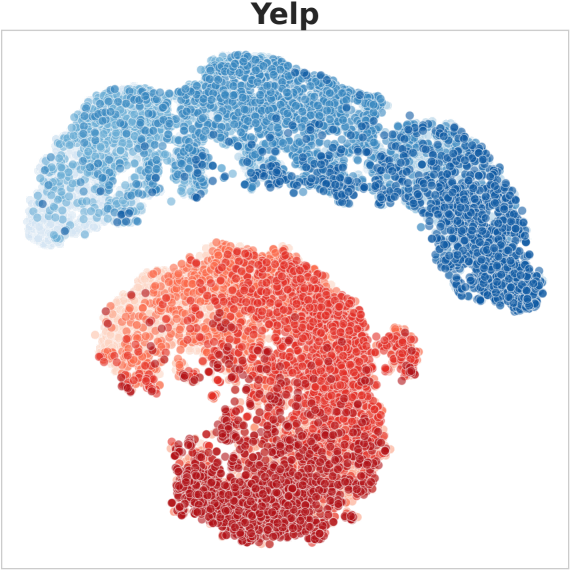} & 
         \includegraphics[width=0.16\linewidth]{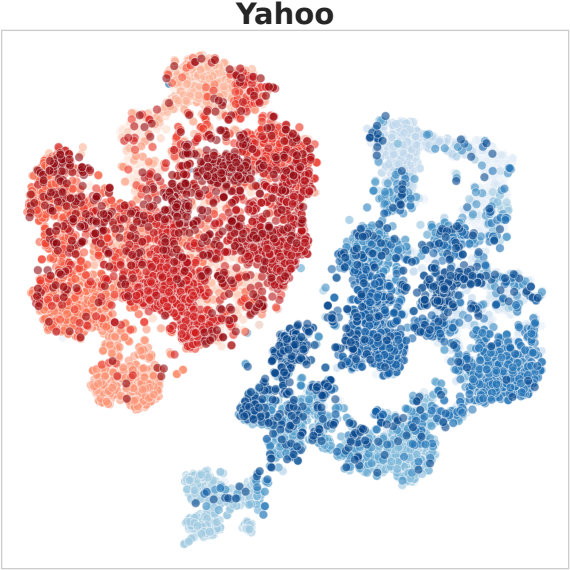} &\includegraphics[width=0.16\linewidth]{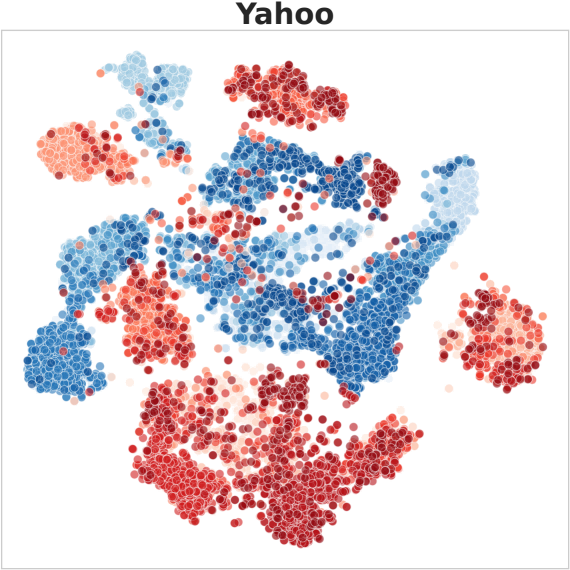} &  \includegraphics[width=0.16\linewidth]{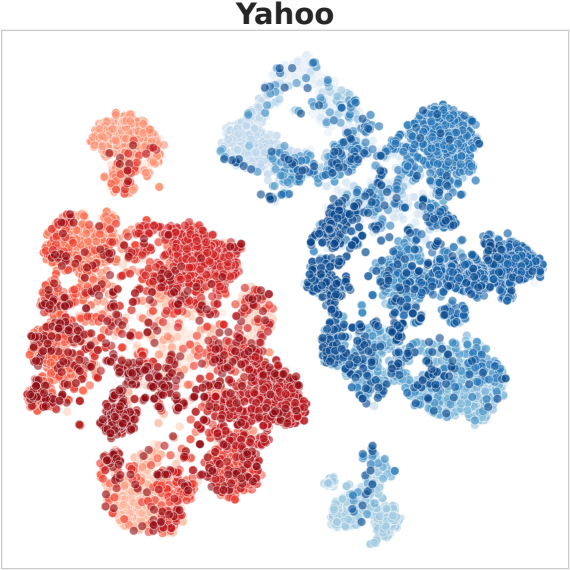} \\
         Weight Averaging & Task Arithmetic & Ties & Weight Averaging & Task Arithmetic & Ties \\
    \end{tabular}
    \caption{T-SNE visualization of merging representations on NLP tasks.}
    \label{fig:visualization-other}
\end{figure*}

\begin{figure*}[h]
    \centering
    \begin{tabular}{@{}c@{\hspace{2pt}}c@{\hspace{2pt}}c@{\hspace{2pt}}c@{\hspace{2pt}}c@{\hspace{2pt}}c@{}}
        \includegraphics[width=0.16\linewidth]{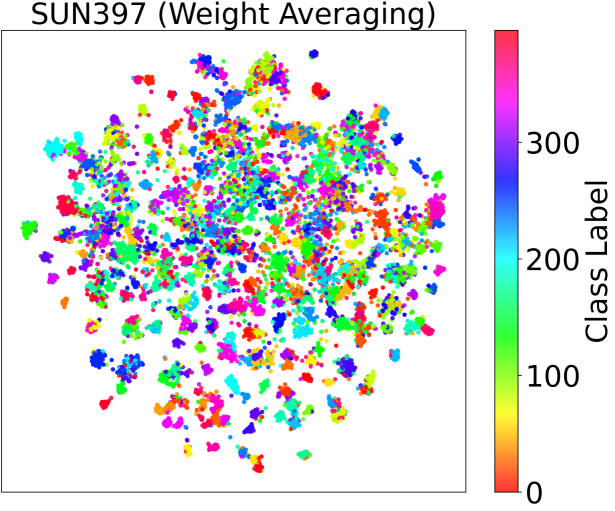} & \includegraphics[width=0.16\linewidth]{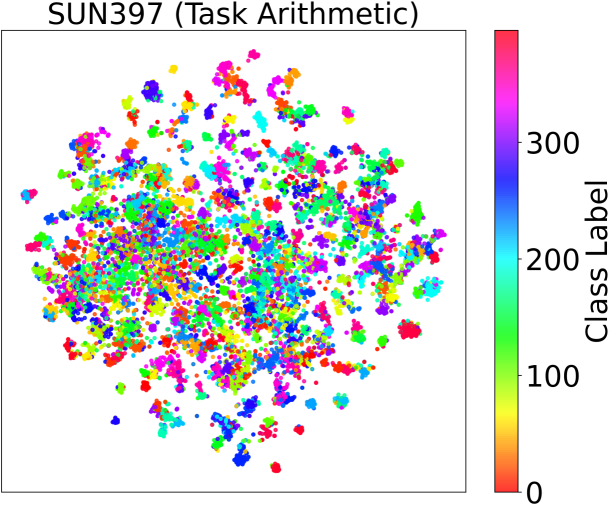} & \includegraphics[width=0.16\linewidth]{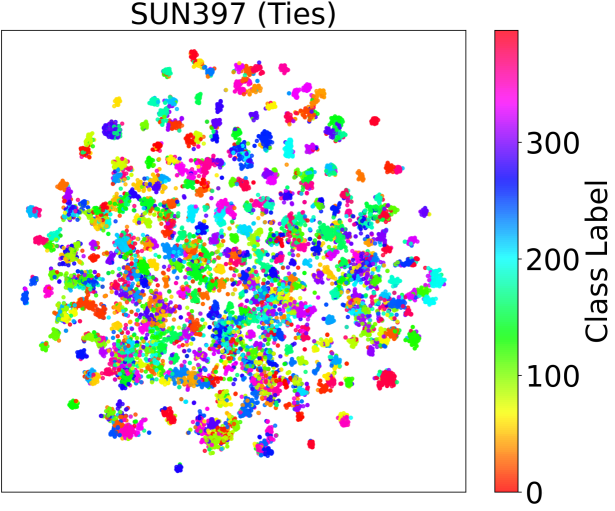} &
        \includegraphics[width=0.16\linewidth]{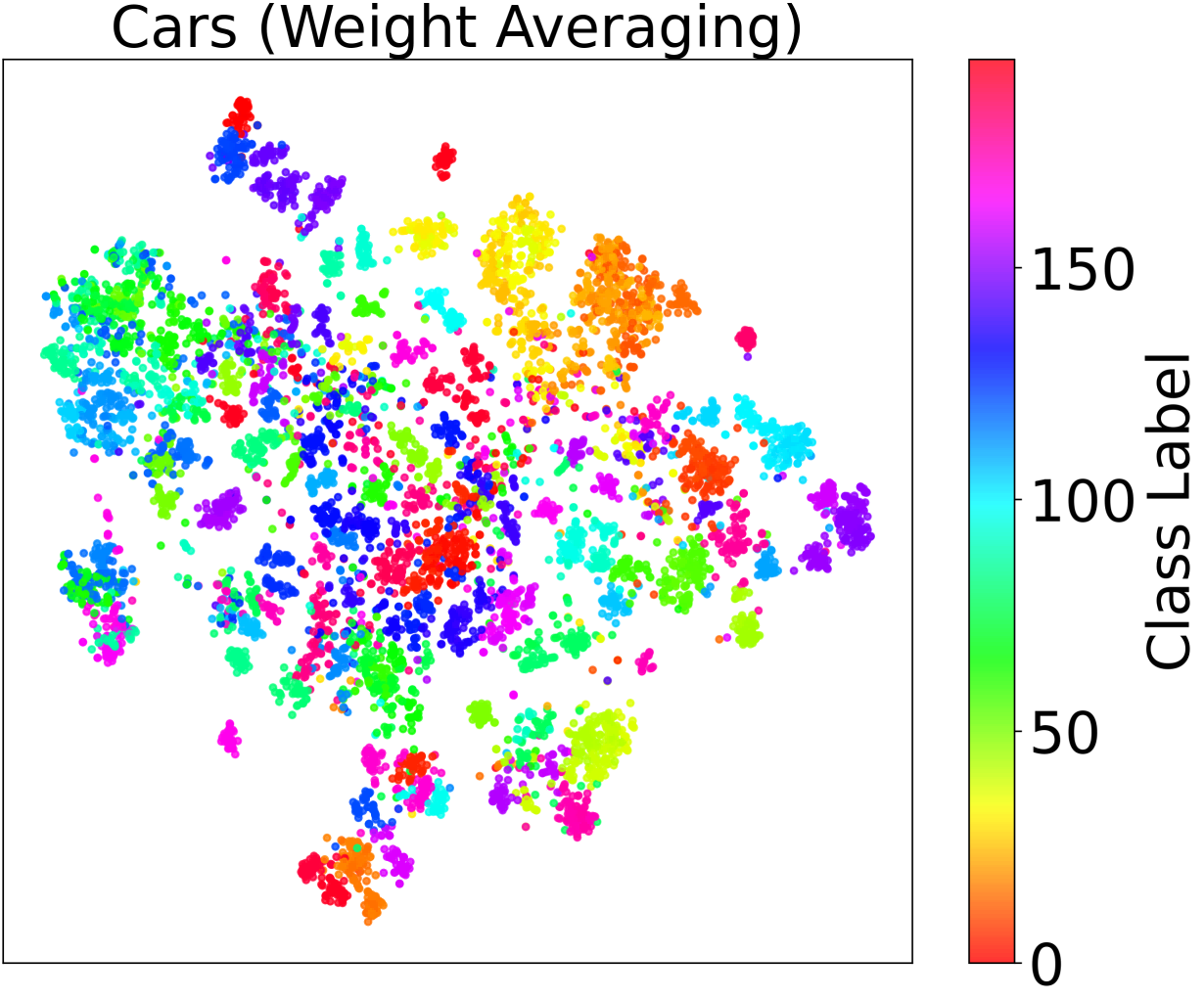} & \includegraphics[width=0.16\linewidth]{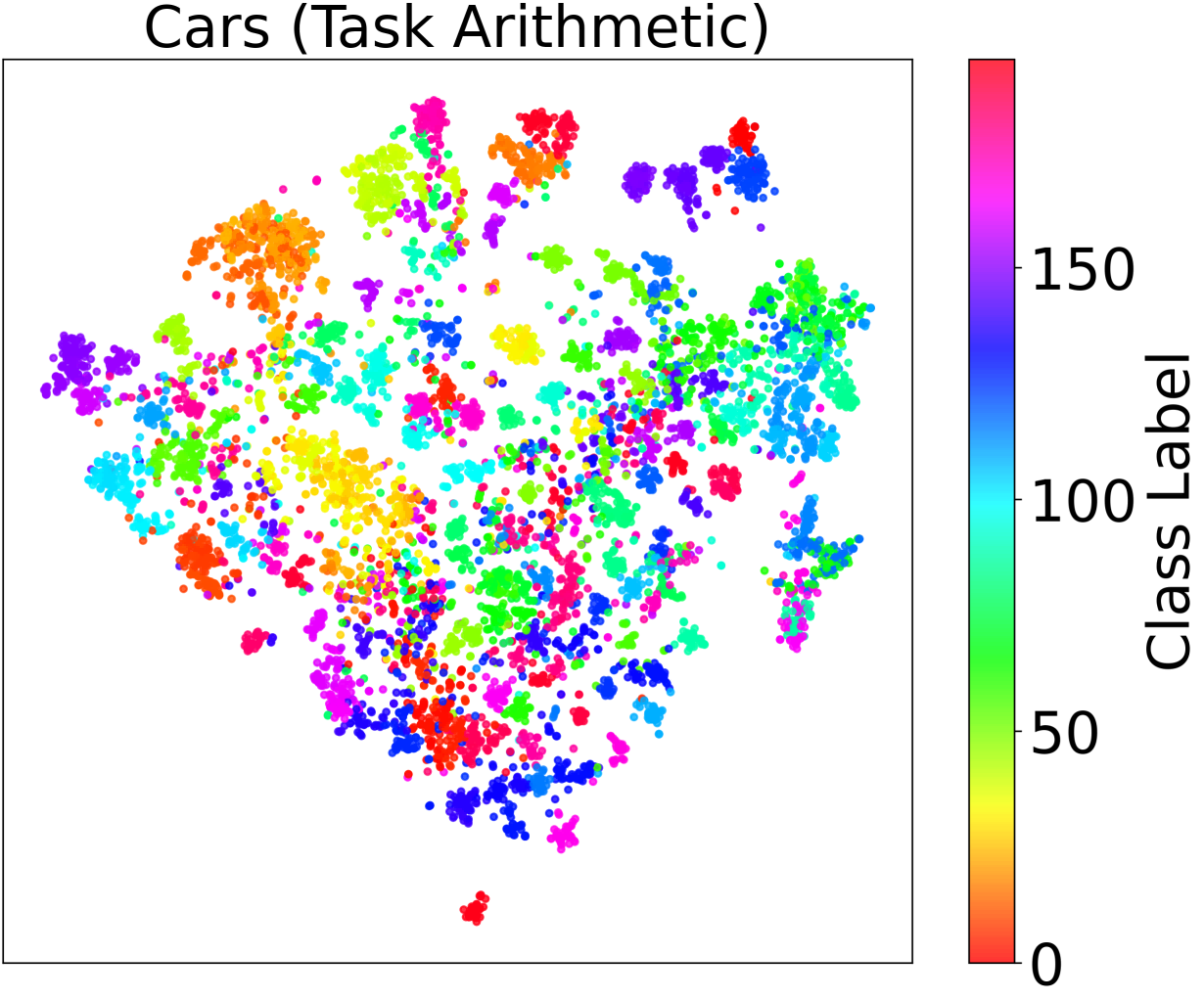} & \includegraphics[width=0.16\linewidth]{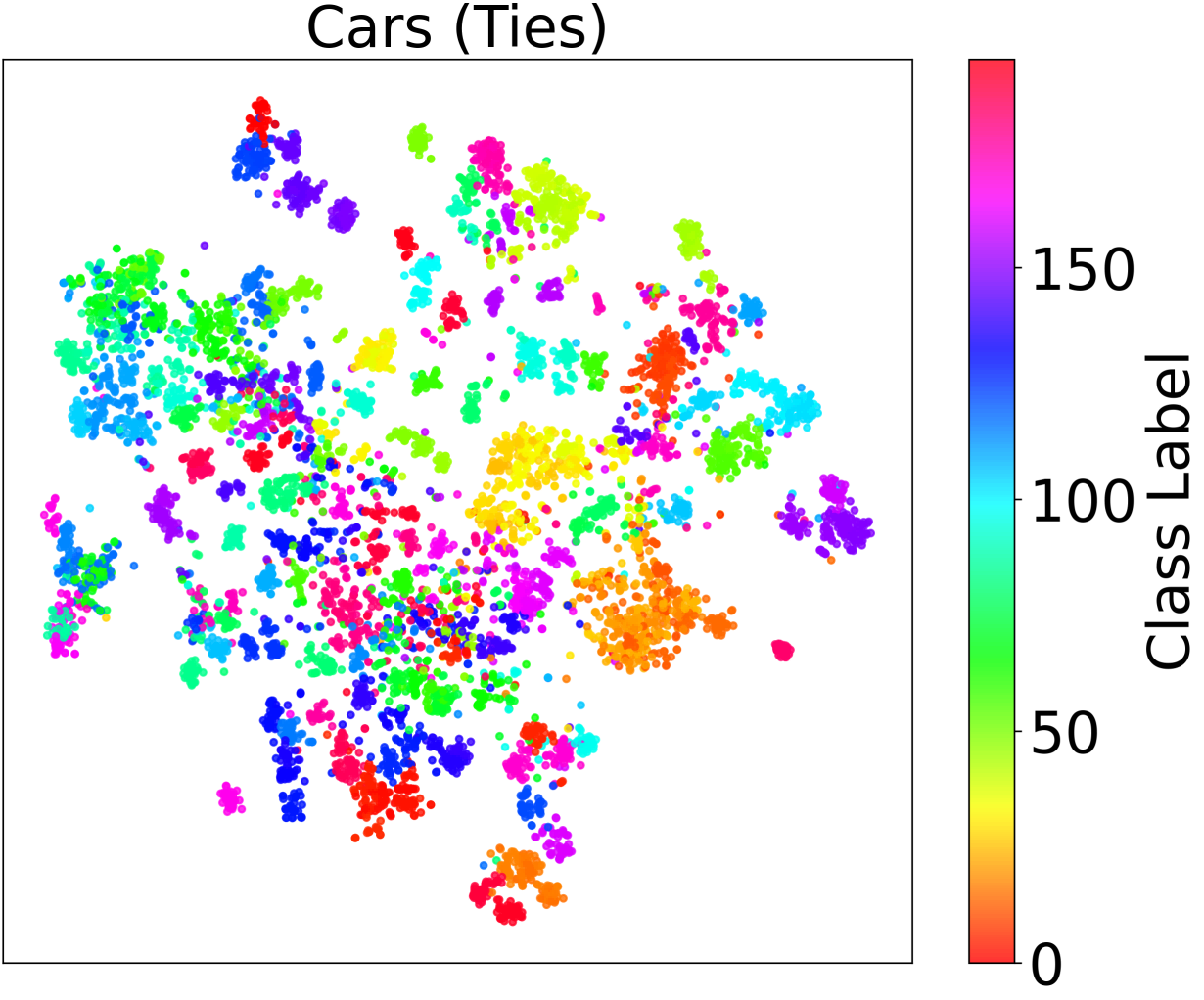}\\
        \includegraphics[width=0.16\linewidth]{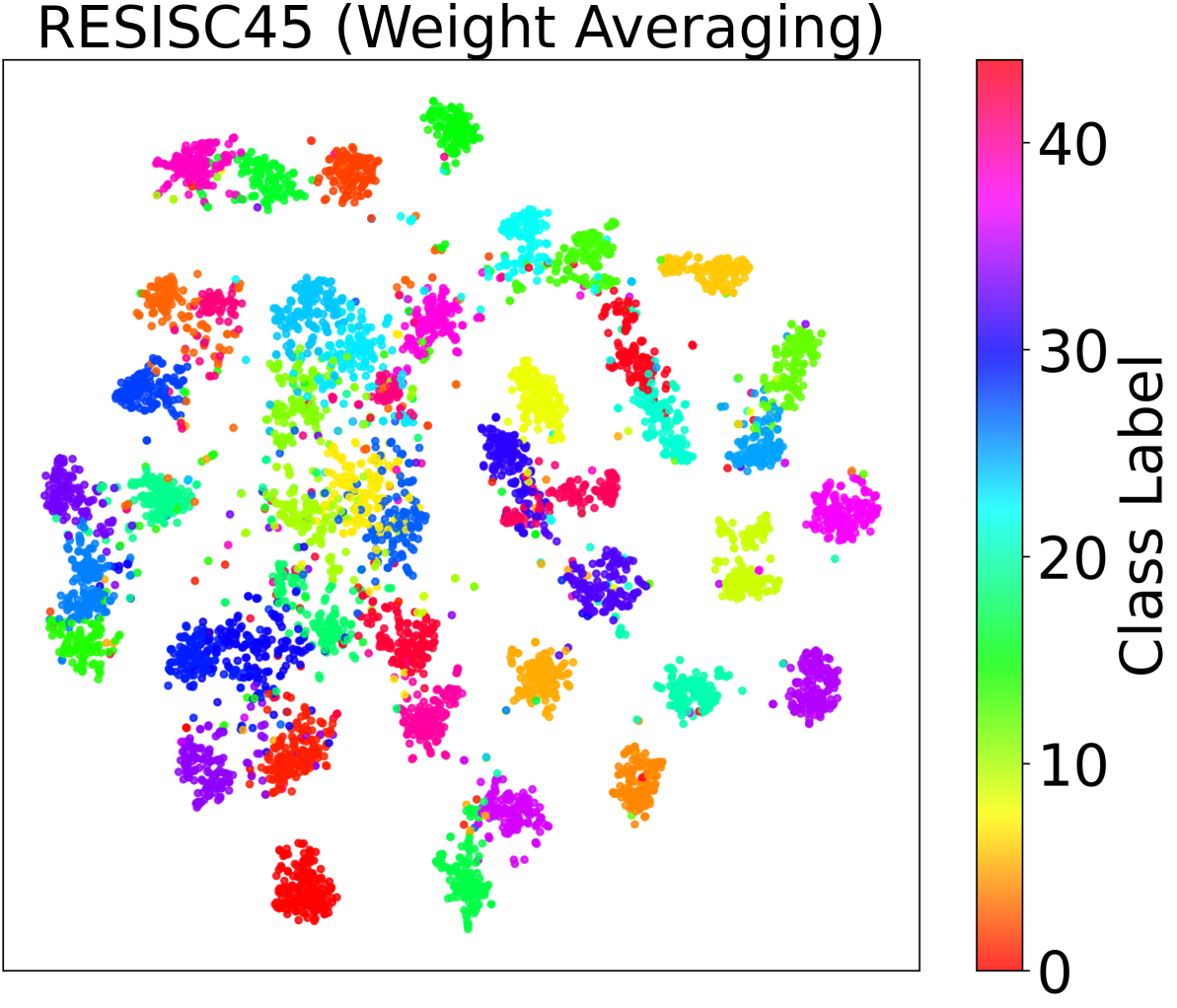} & \includegraphics[width=0.16\linewidth]{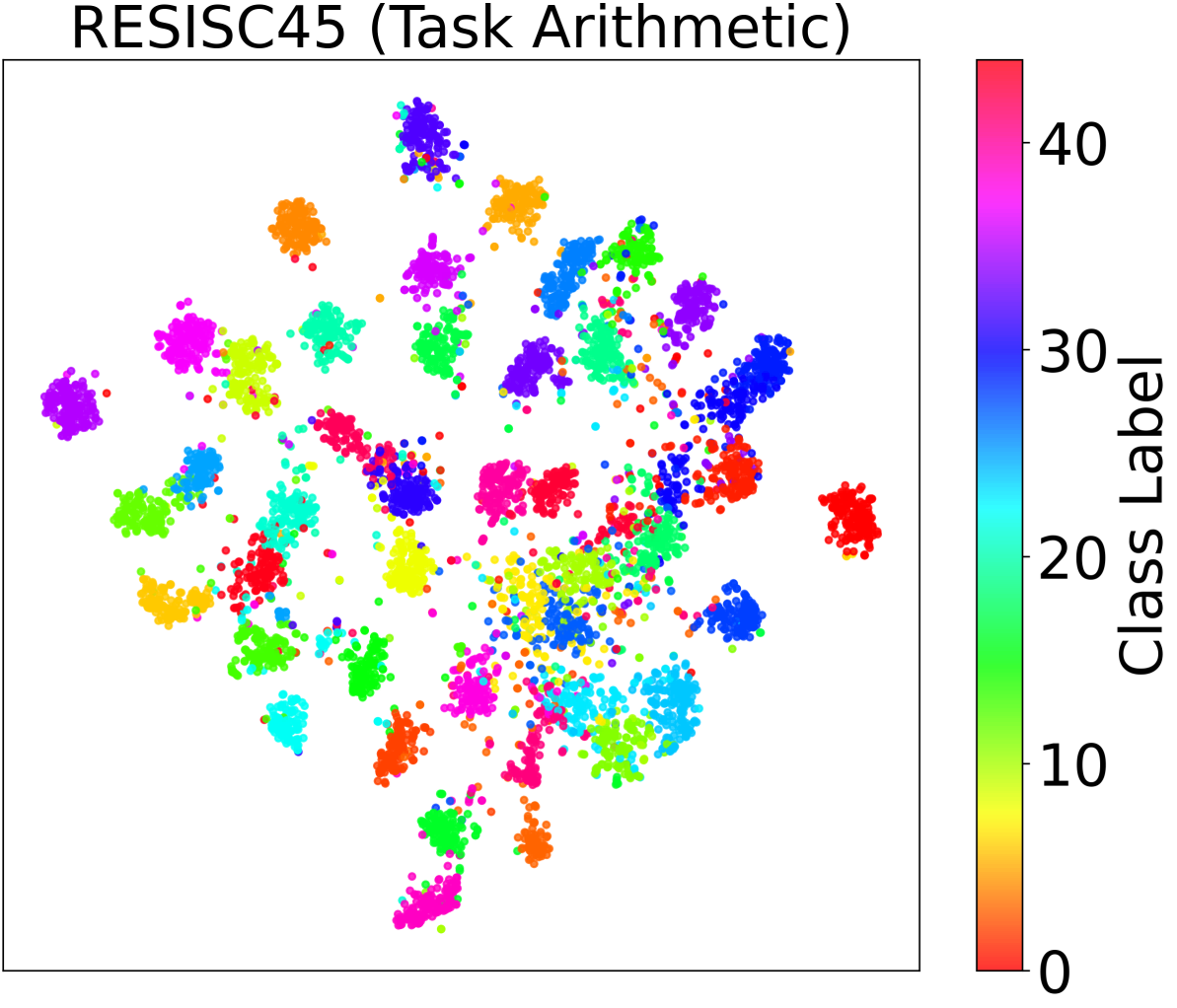} & \includegraphics[width=0.16\linewidth]{cv/ta_RESISC45_tsne.png} &
        \includegraphics[width=0.16\linewidth]{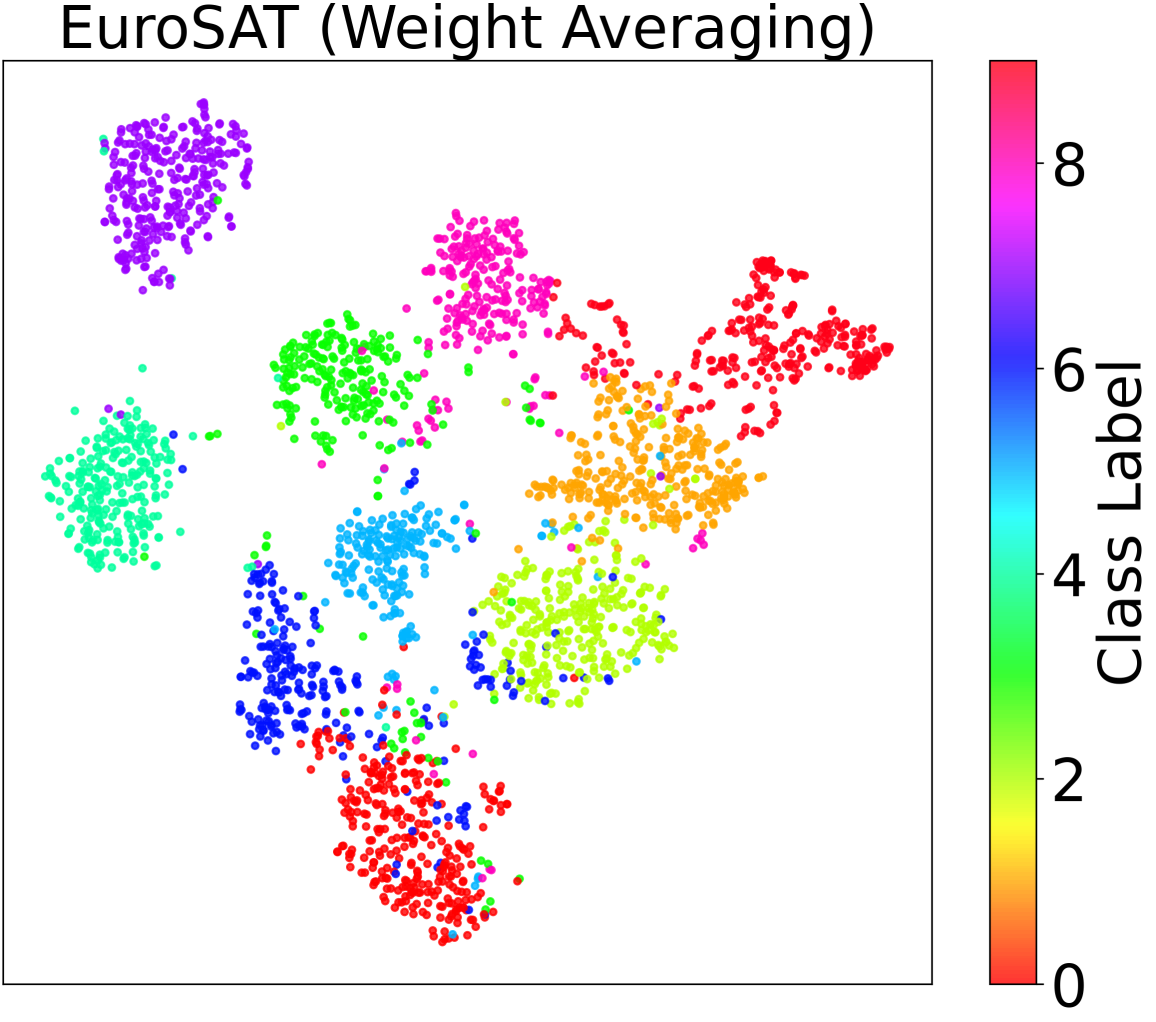} & \includegraphics[width=0.16\linewidth]{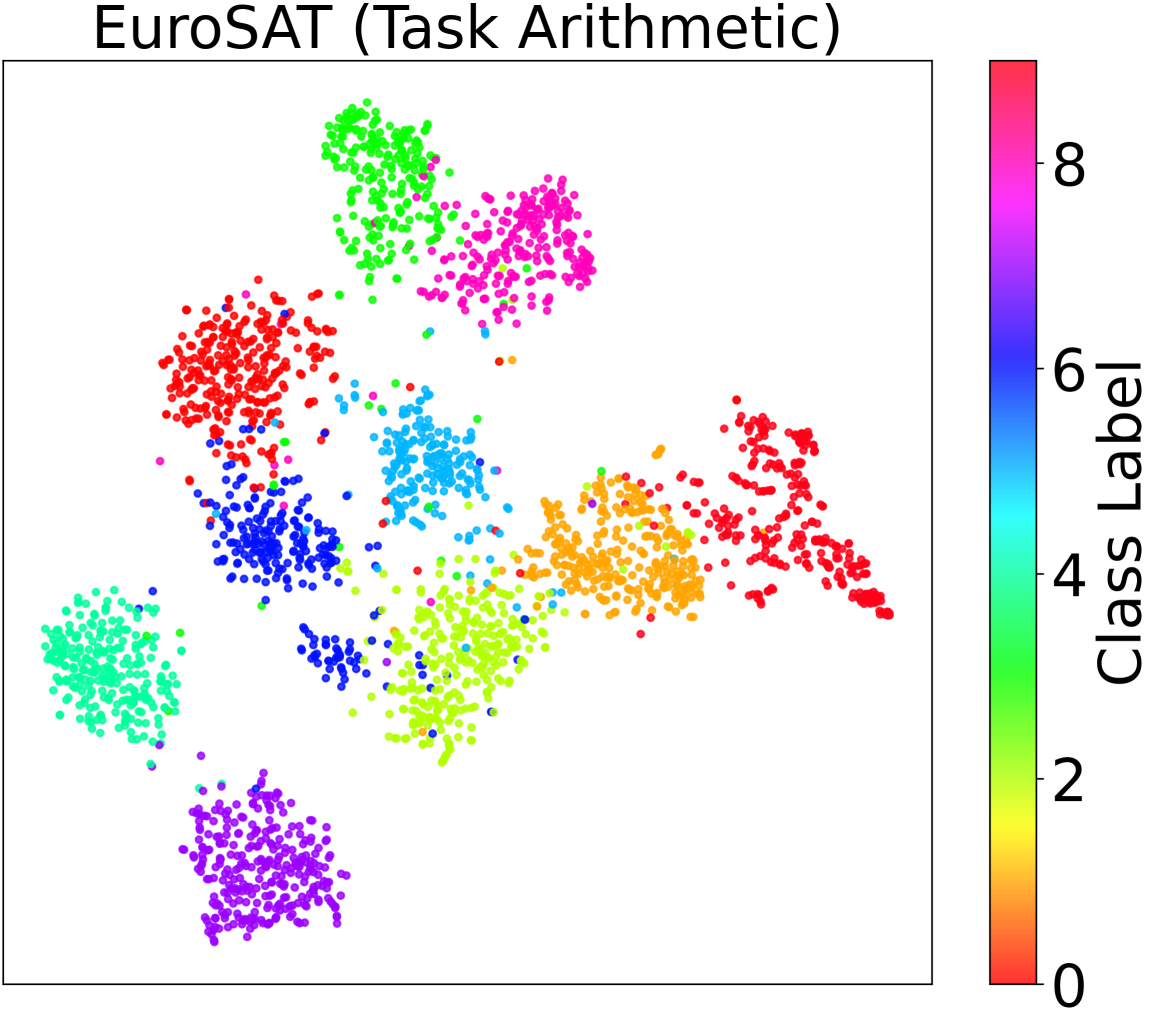} & \includegraphics[width=0.16\linewidth]{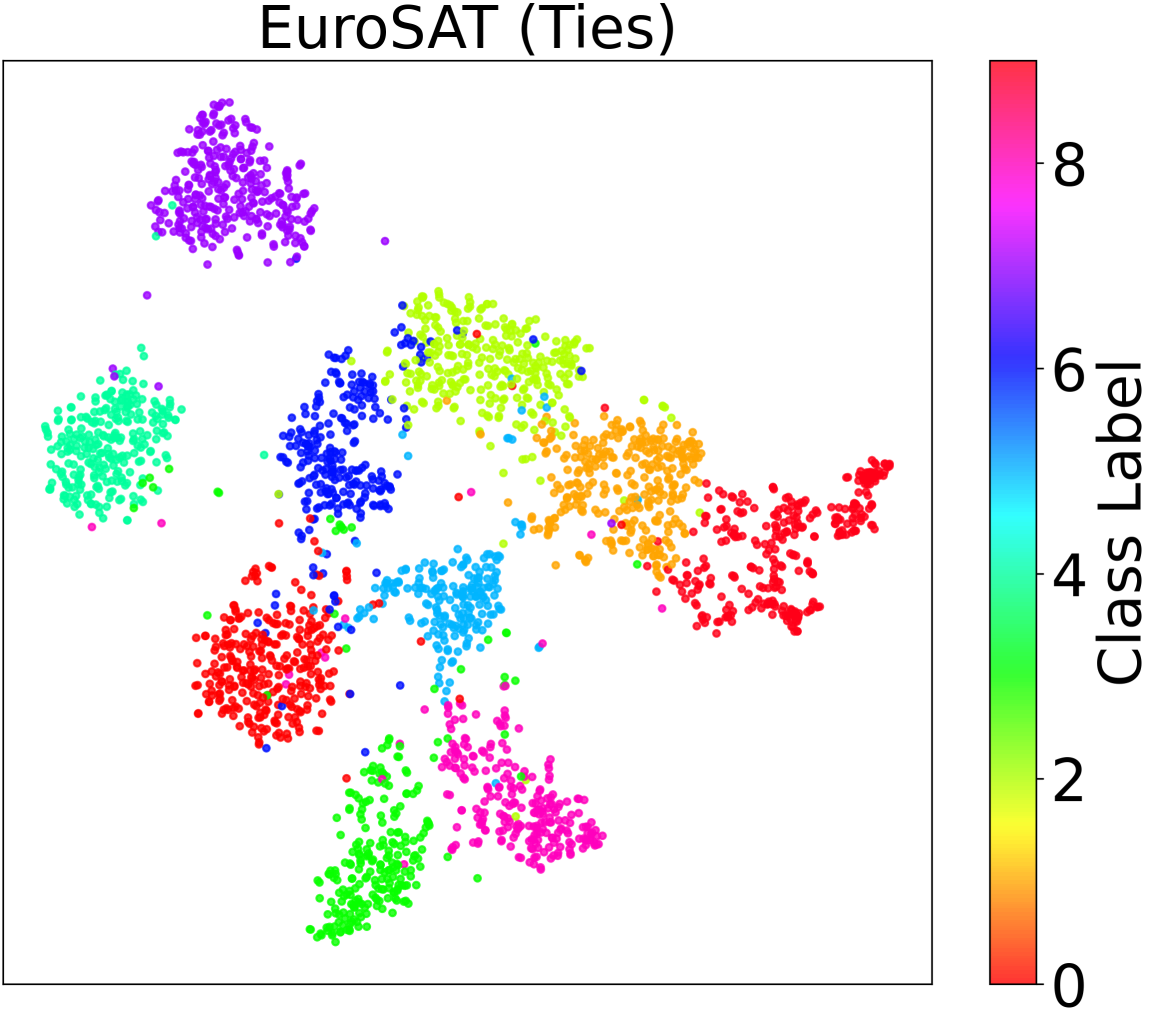}\\
        \includegraphics[width=0.16\linewidth]{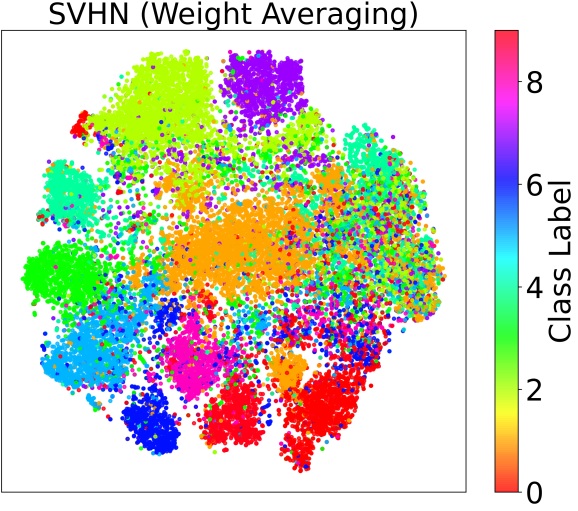} & \includegraphics[width=0.16\linewidth]{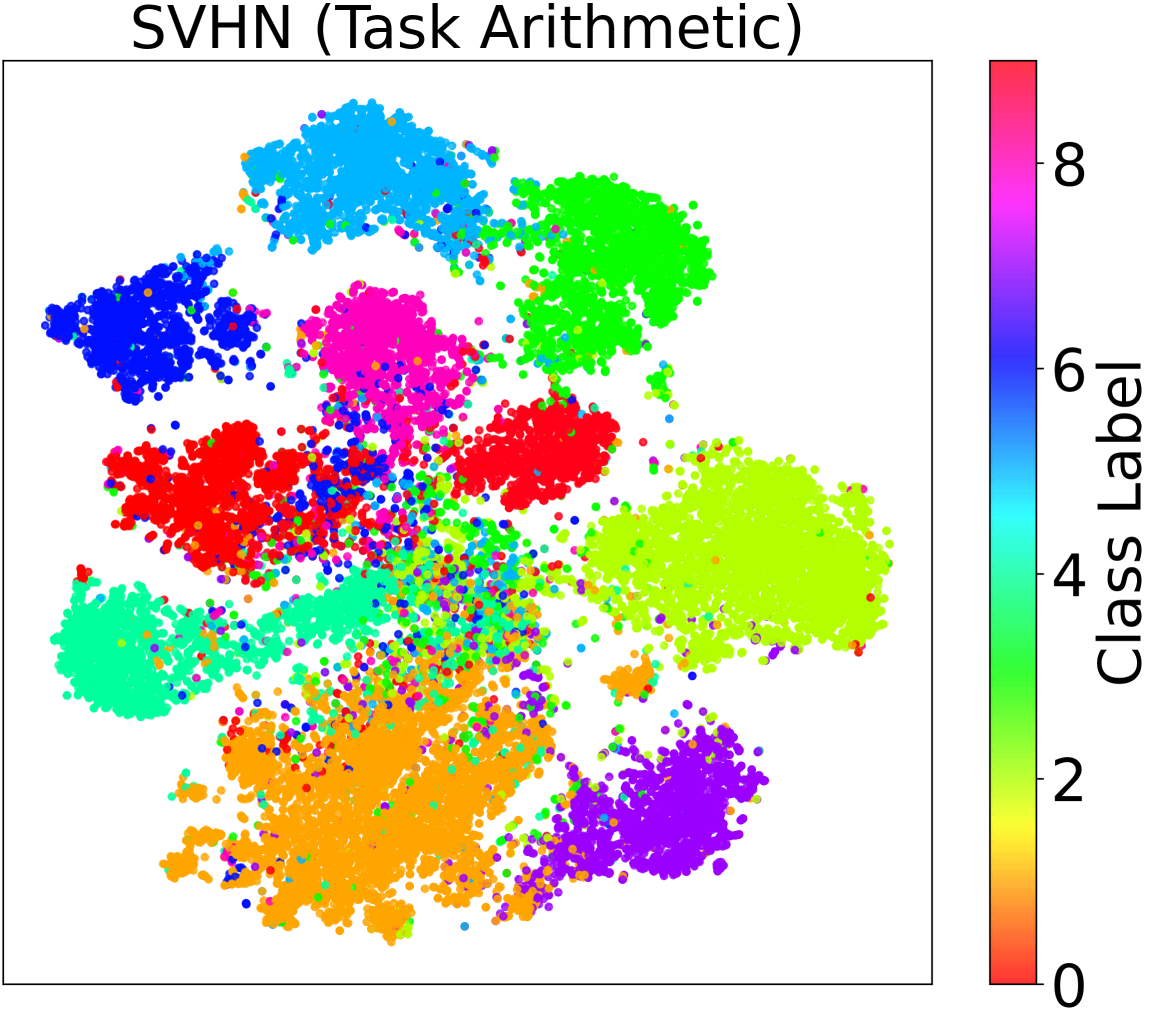} & \includegraphics[width=0.16\linewidth]{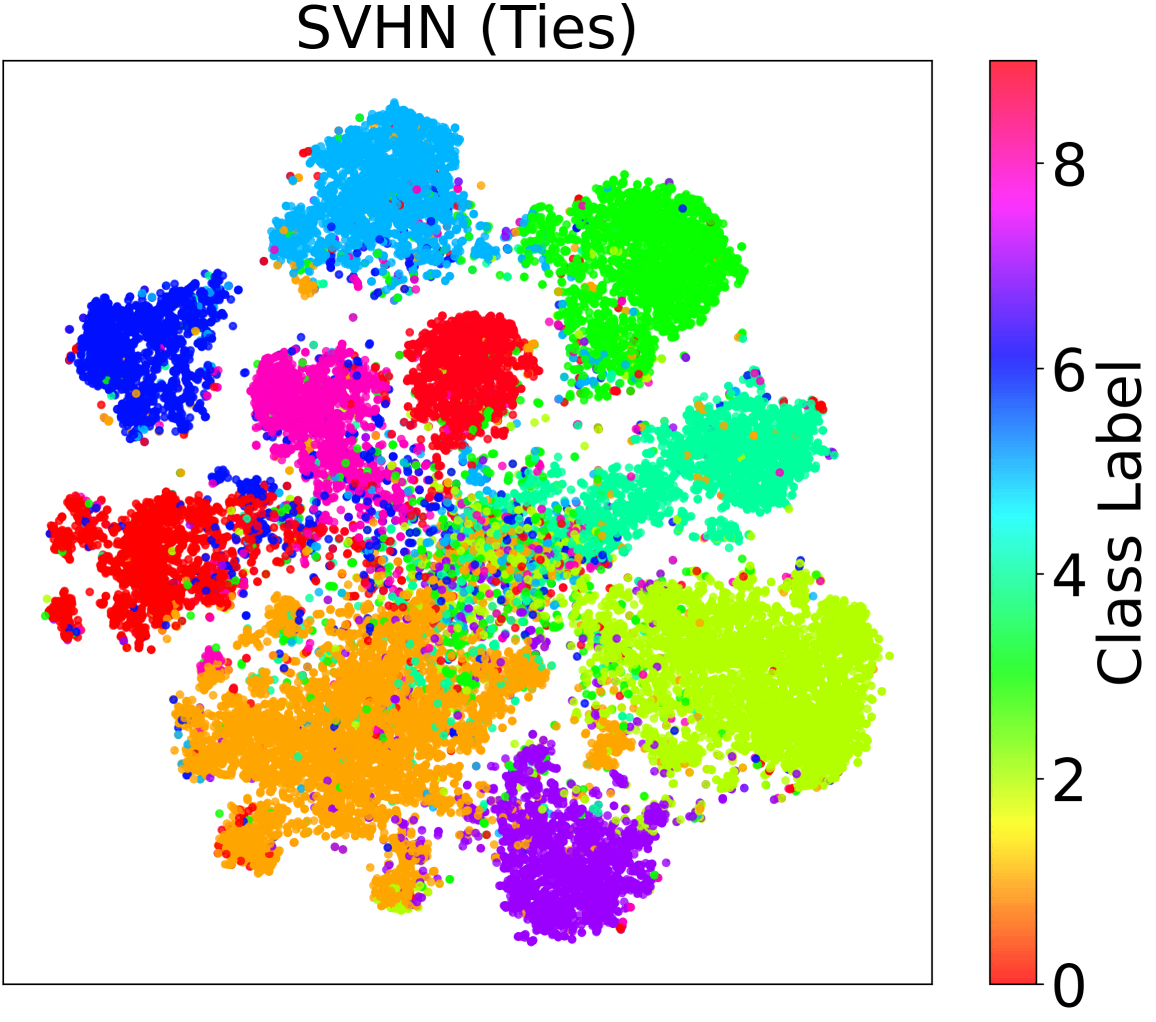} &
        \includegraphics[width=0.16\linewidth]{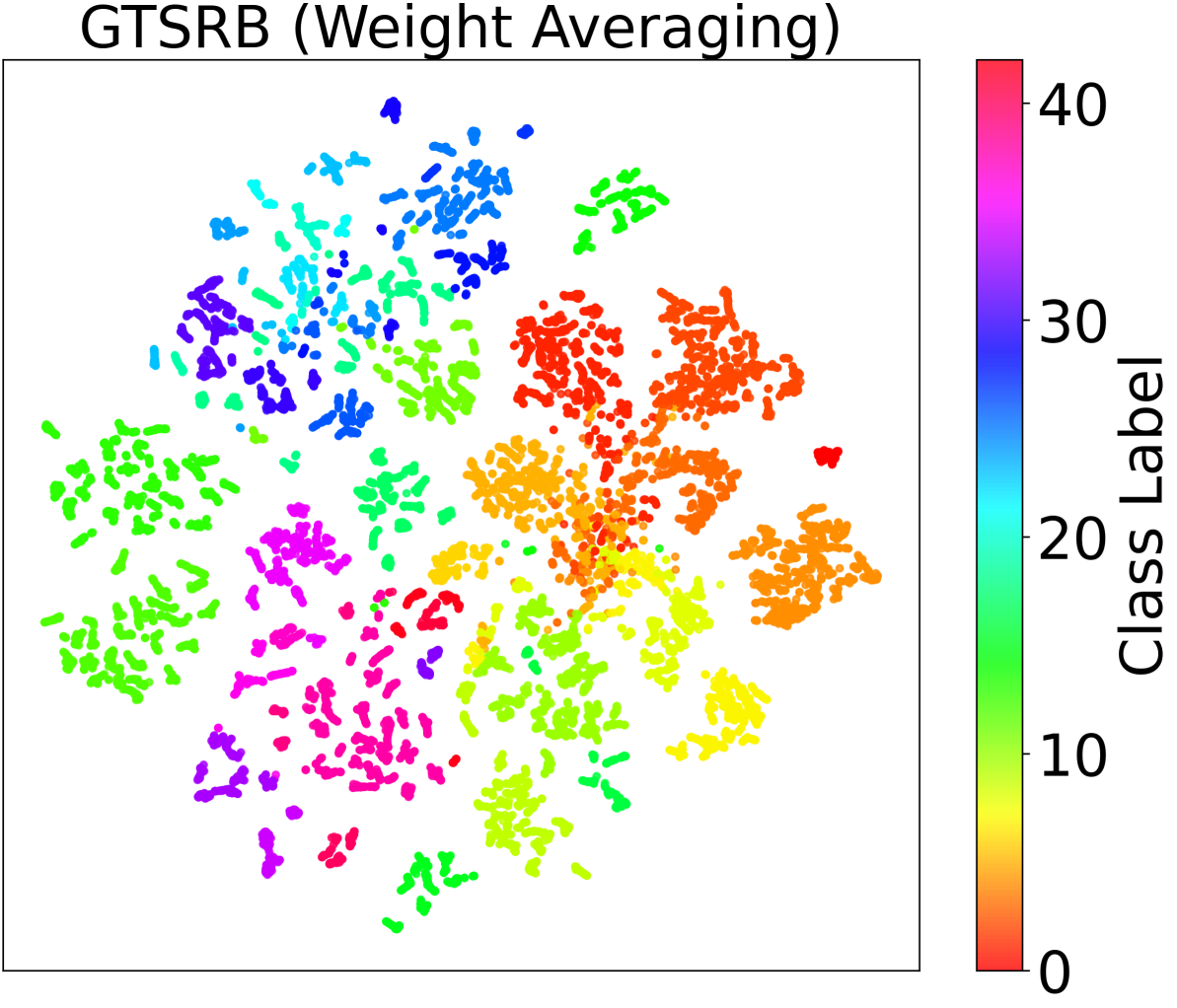} & \includegraphics[width=0.16\linewidth]{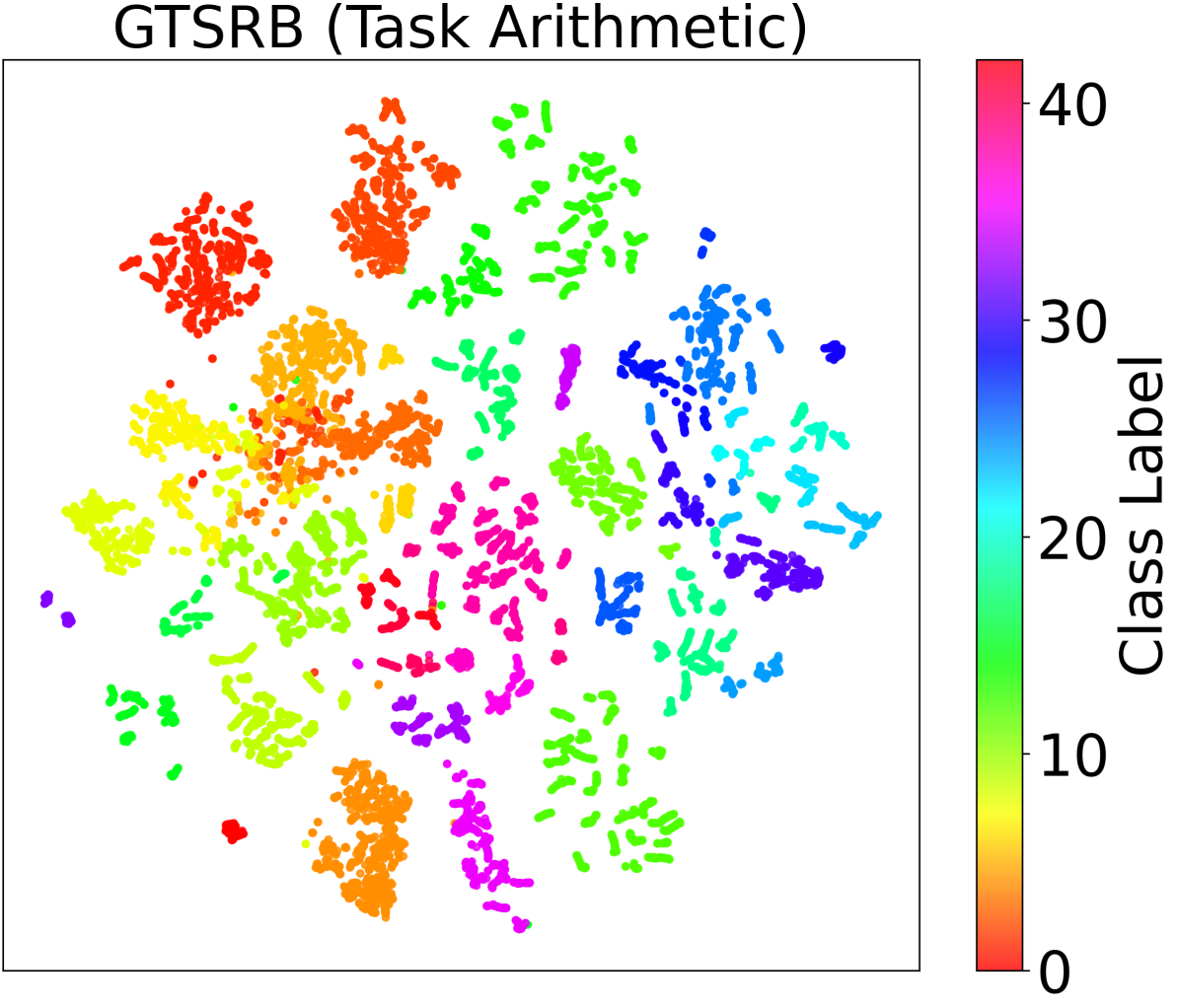} & \includegraphics[width=0.16\linewidth]{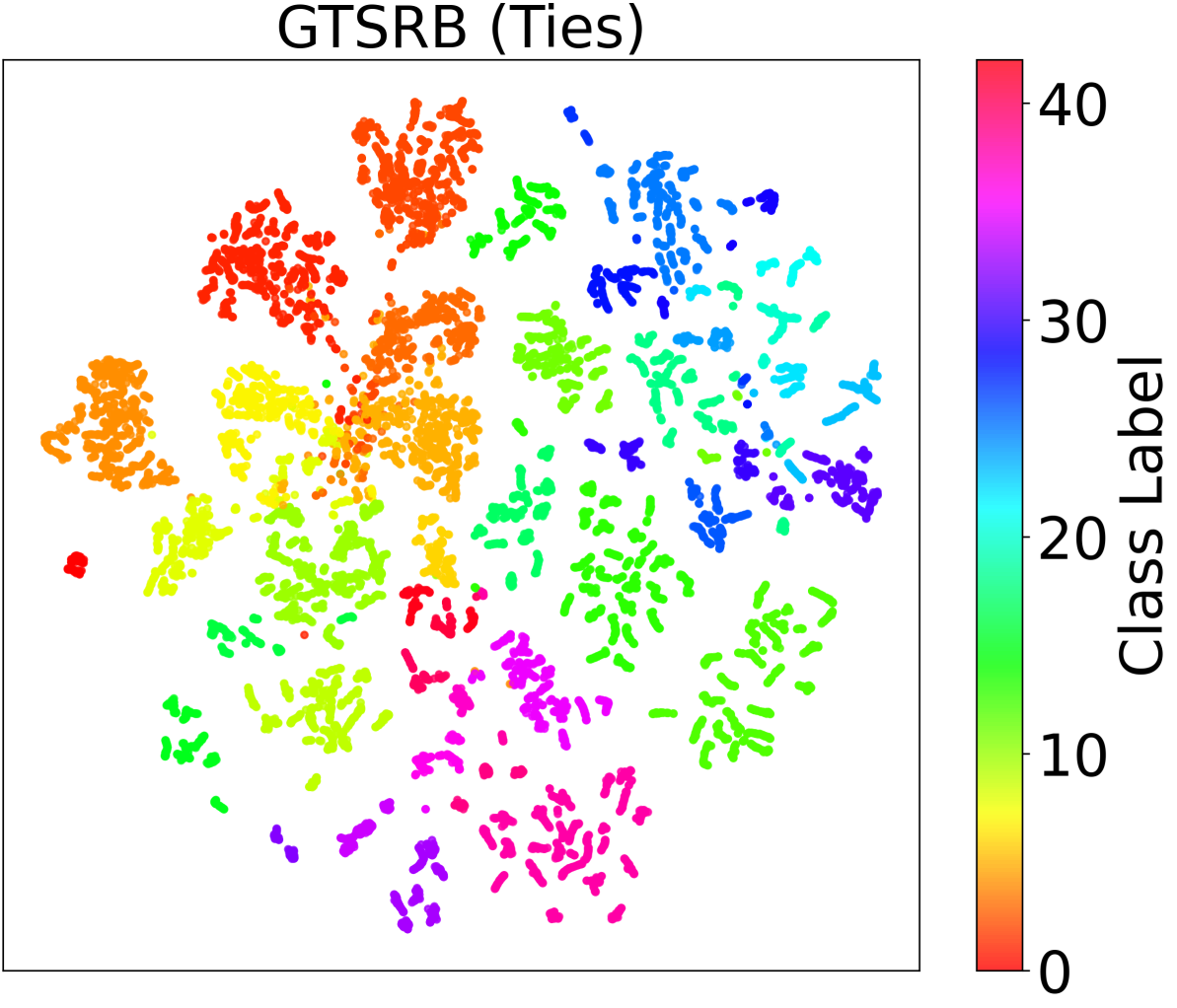}\\
        \includegraphics[width=0.16\linewidth]{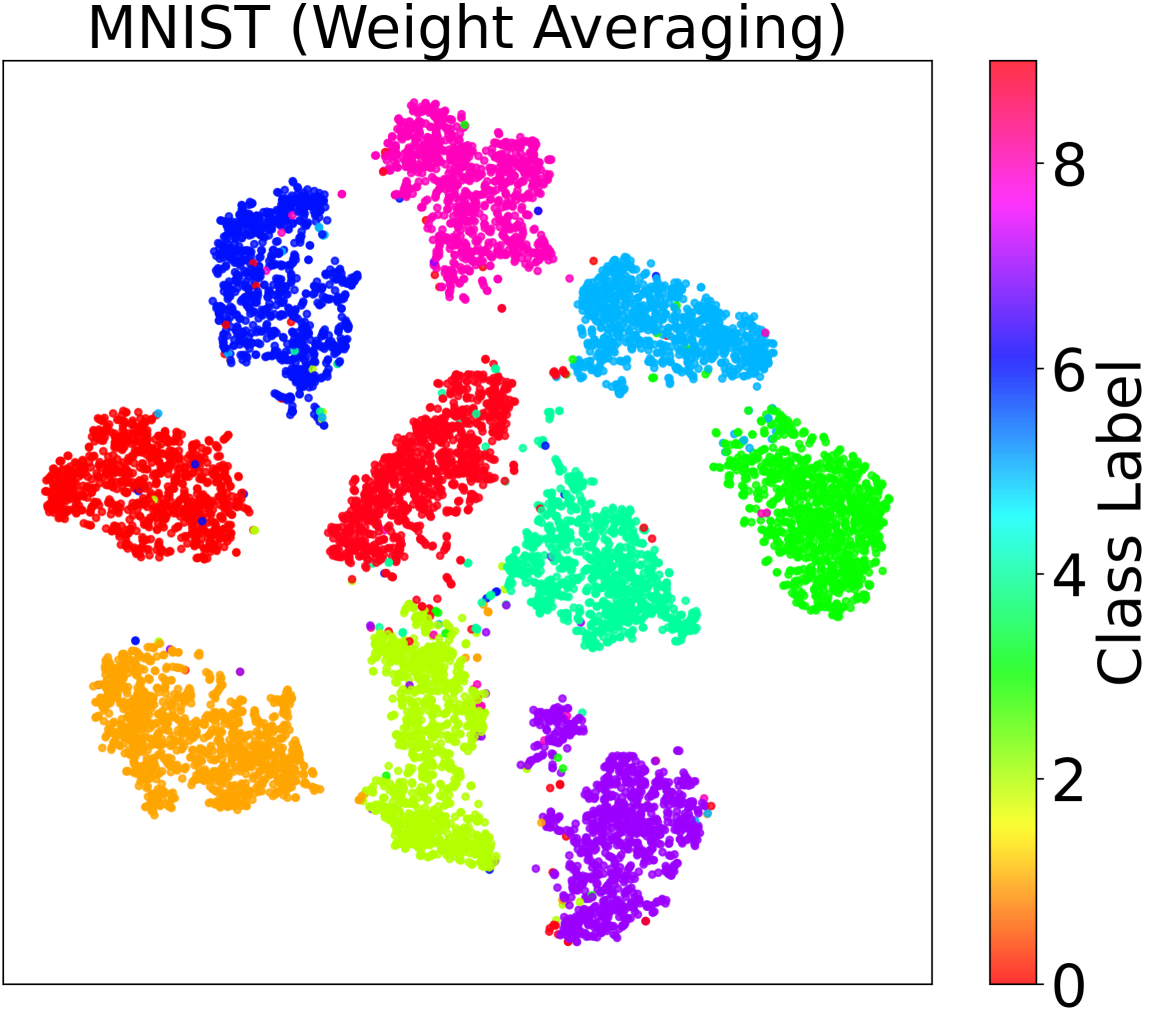} & \includegraphics[width=0.16\linewidth]{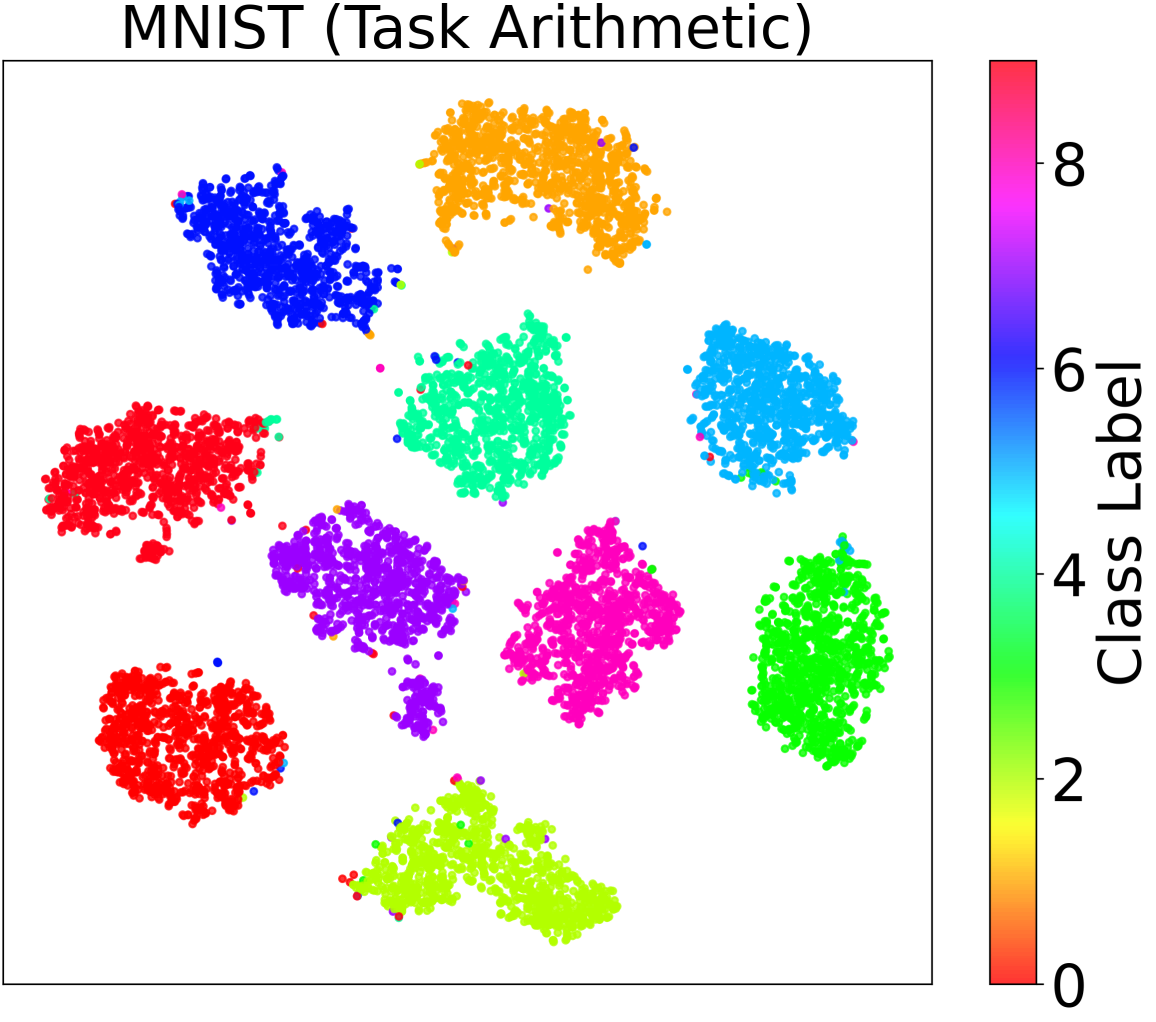} & \includegraphics[width=0.16\linewidth]{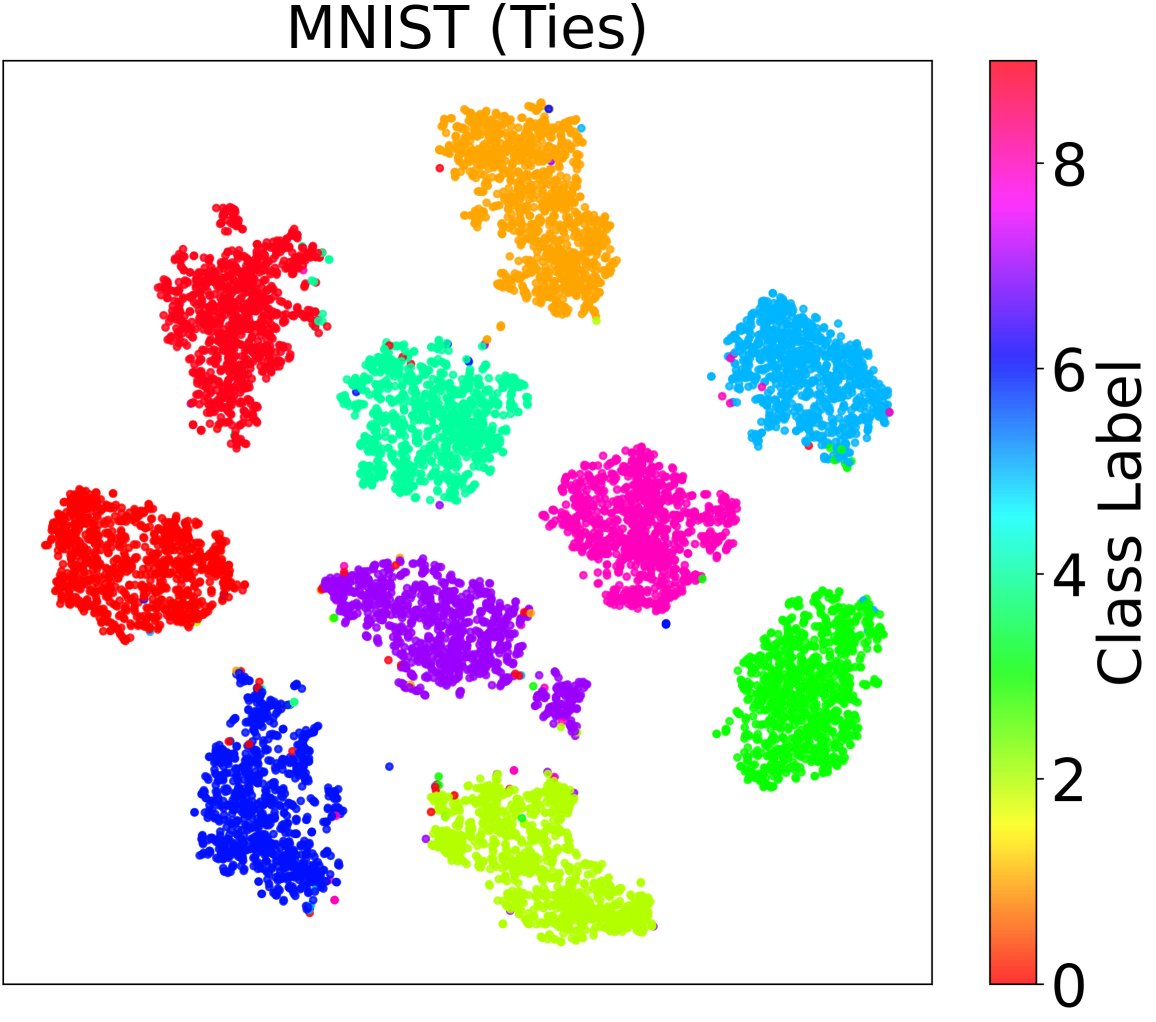} &
        \includegraphics[width=0.16\linewidth]{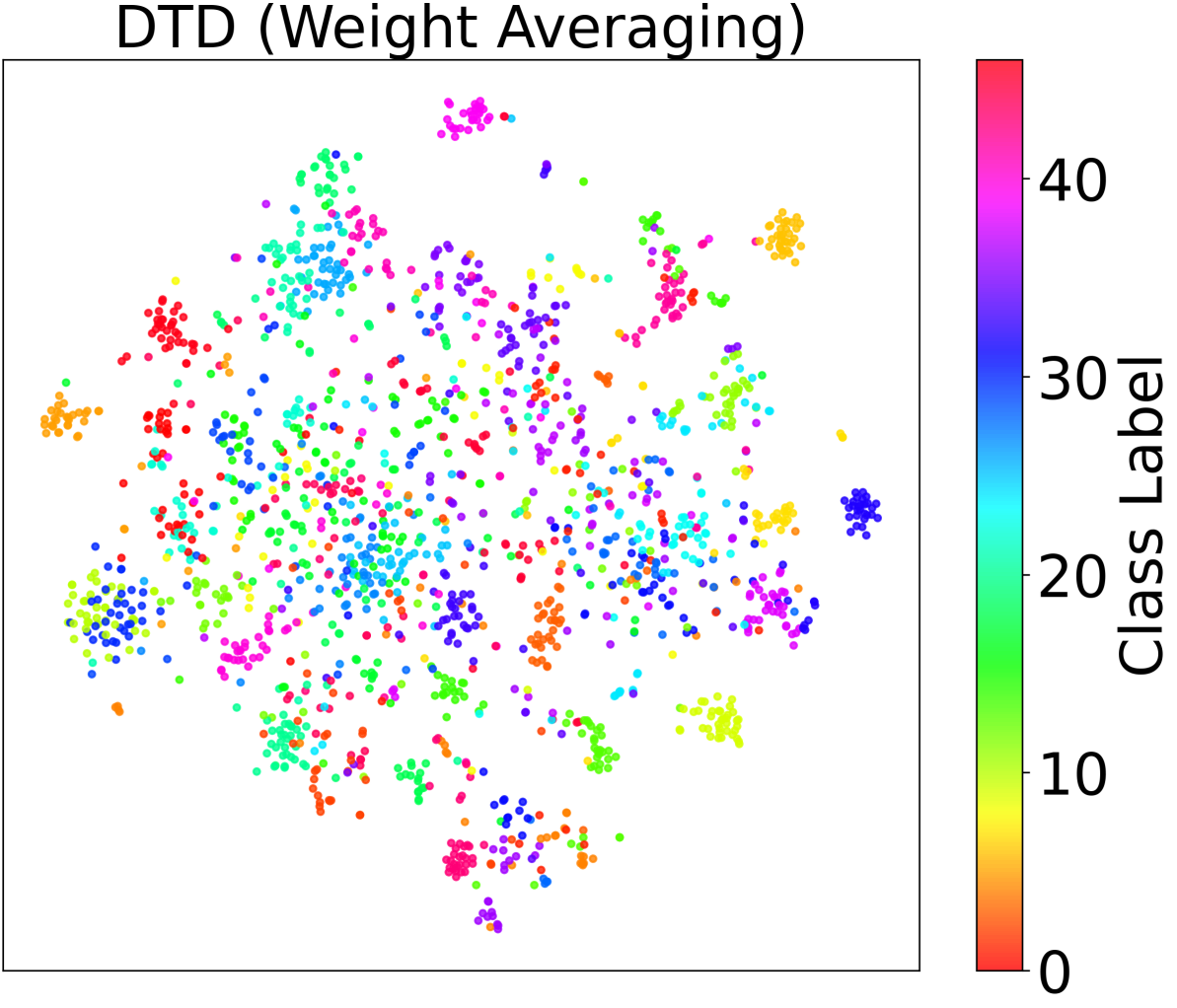} & \includegraphics[width=0.16\linewidth]{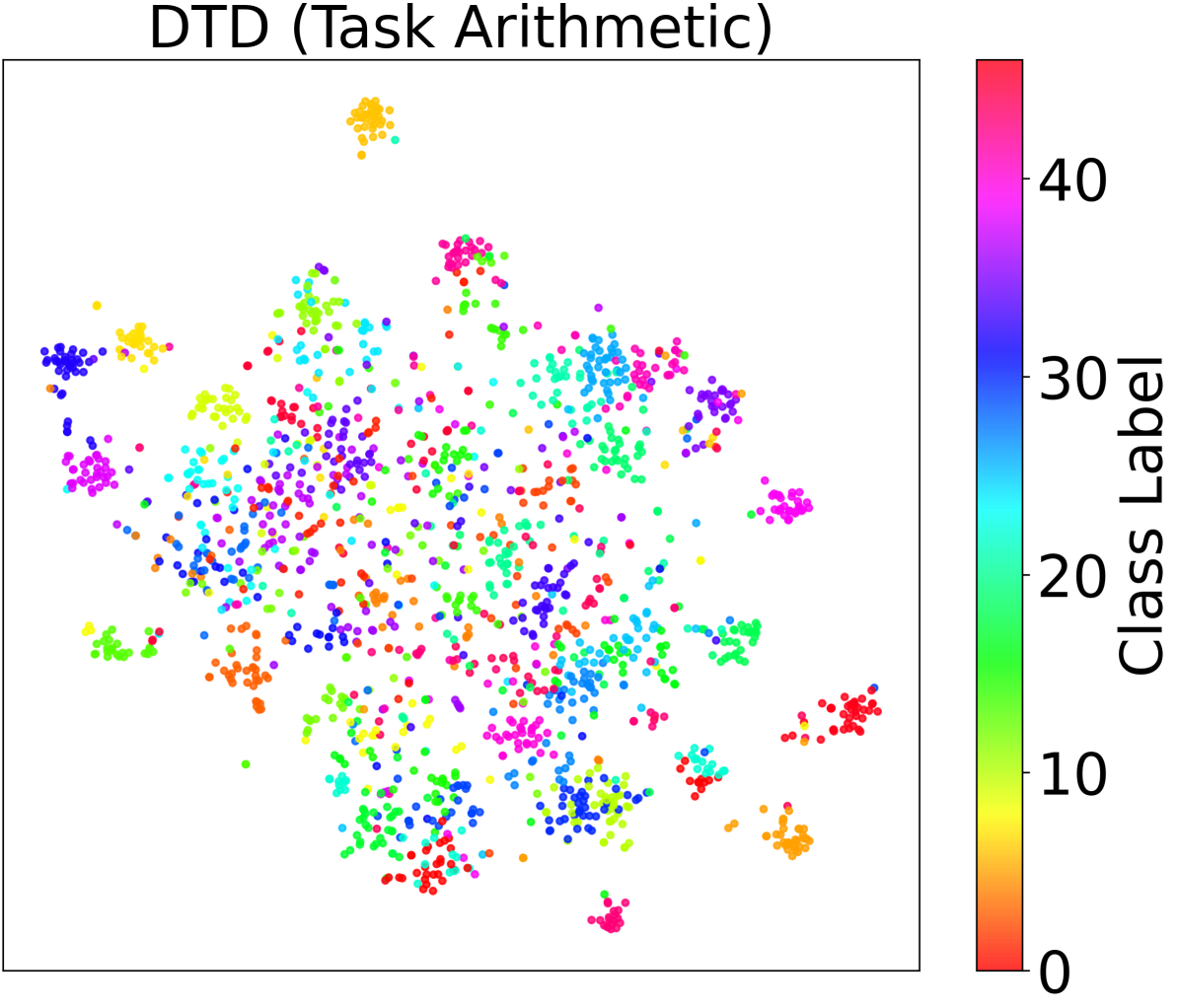} & \includegraphics[width=0.16\linewidth]{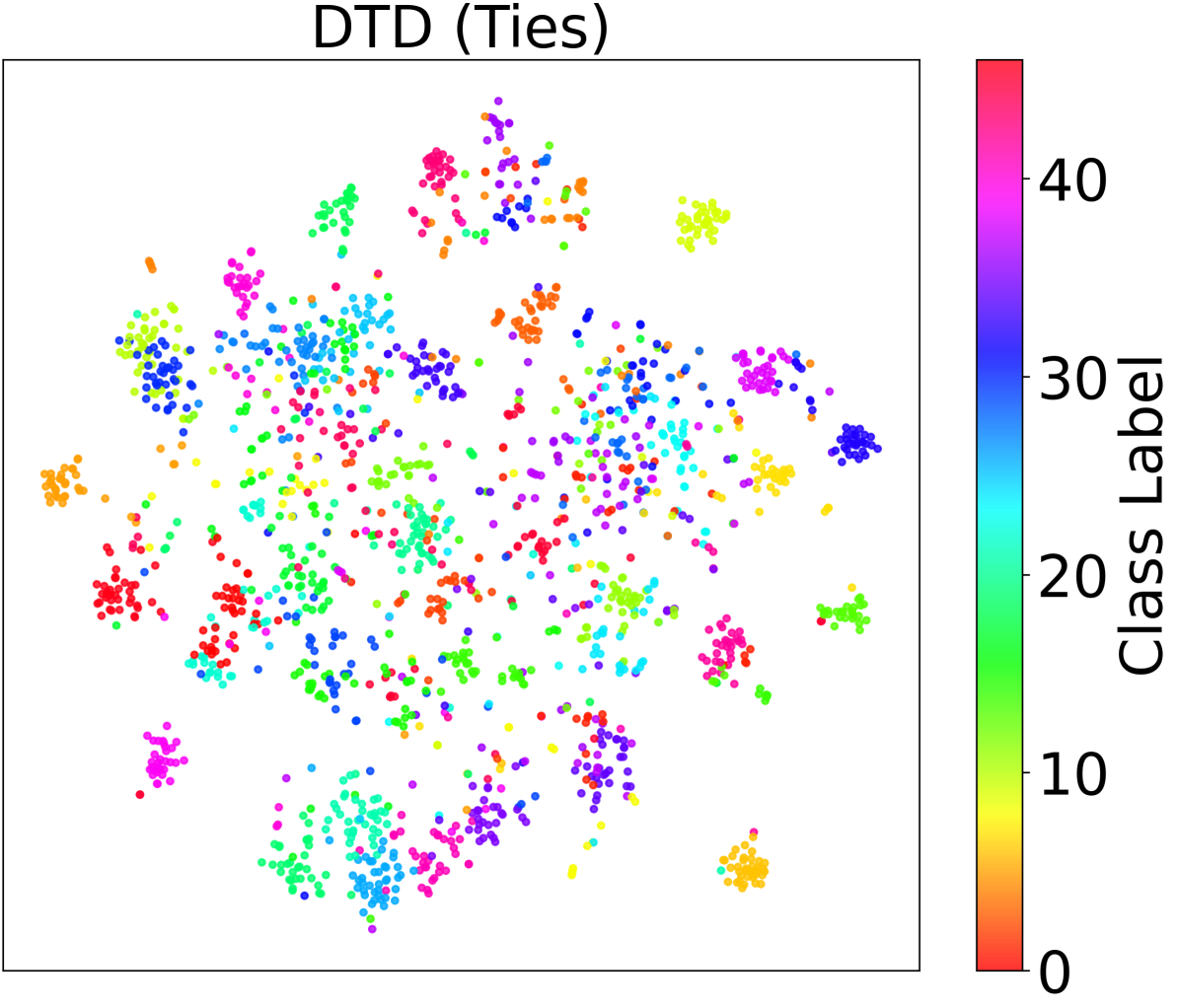}\\
    \end{tabular}
    \caption{T-SNE visualization of merging representations for each merging method on CV tasks. Noted, due to the large number of classes in the CV datasets, we did not display the T-SNE results of the fine-tuned model for better visualization.}
    \label{fig:t-sne cv}
\end{figure*}

T-SNE visualization of merging representation produced by different methods on other NLP and CV tasks is shown in Figure~\ref{fig:visualization-other} and Figure~\ref{fig:t-sne cv}. From these Figures, we observe that on other datasets, the merged representations also exhibit clear clustering behavior, indicating that the merged representations retain the necessary task-relevant information.

\section{Experimental Setup}
\subsection{Datasets}
\label{sec:app:data}
For CV tasks, we select SUN397 (397 classes)~\cite{xiao2016sun}, Cars (196 classes)~\cite{Krause_Stark_Deng_Fei-Fei_2013}, RESISC45 (45 classes)~\cite{Cheng_Han_Lu_2017}, EuroSAT (10 classes)~\cite{Helber_Bischke_Dengel_Borth_2019}, SVHN (10 classes)~\cite{netzer2011reading}, GTSRB (43 classes)~\cite{Stallkamp_Schlipsing_Salmen_Igel_2011}, MNIST (10 classes)~\cite{deng2012mnist}, and DTD (47 classes)~\cite{cimpoi2014describing}.

For NLP tasks, we select AG News (4 classes)~\cite{zhang2015character}, Yelp (5 classes)~\cite{yelp2014}, Amazon (5 classes)~\cite{mcauley2013hidden}, Yahoo (10 classes)~\cite{zhang2015character}, and DBpedia (14 classes)~\cite{dbpedia}.

Our dataset selection is consistent with prior studies~\cite{pmlr-v235-yang24t,wei2025probsurgery}. Detailed descriptions and statistical characteristics of the datasets can be found in these works.

\subsection{Baselines}
\label{sec:app:baseline}
Following~\citet{wei2025probsurgery}, we conduct comparisons with the following representative methods.

\begin{itemize}
    \item \textbf{Pre-trained Model}: The foundational pre-trained backbone evaluated directly in a zero-shot manner without task-specific tuning.

    \item \textbf{Fine-tuned Model}: Models fine-tuned independently on each specific task, serving as the single-task performance upper bound.

    \item \textbf{Traditional MTL}: A model trained jointly on all datasets using a shared backbone, serving as the multi-task upper bound.
    
    \item \textbf{Weight Averaging (WA)}~\cite{wortsman2022model_model_soup}: A baseline approach that performs simple averaging of parameters from independently fine-tuned models to form a unified backbone.
    
    \item \textbf{Task Arithmetic (TA)}~\cite{ilharco2022editing_task_arithmetic}: Characterizes task-specific updates as vectors in the parameter space, enabling the manipulation of model behaviors through arithmetic operations such as addition and negation.
    
    \item \textbf{Ties-Merging}~\cite{yadav2024ties}: Mitigates parameter interference by pruning redundant updates, resolving sign conflicts via a ``dominant direction" criterion, and keeping only the consensus parameters.
    
    \item \textbf{Fisher Merging}~\cite{matena2022fishermerging}: Utilizes the Fisher Information Matrix to assign higher importance weights to critical parameters, thereby preserving task-specific knowledge during the averaging process.
    
    \item \textbf{RegMean}~\cite{jin2022dataless_regmean}: Formulates merging as a least-squares regression problem, minimizing the distance between the activations of the merged model and those of fine-tuned models.
    
    \item \textbf{Concrete TA \& AM}~\cite{tang2023concrete}: Proposes a continuous relaxation of Task Arithmetic, employing a concrete distribution to learn optimal fractional masks and merging coefficients.
    
    \item \textbf{AdaMerging}~\cite{yangadamerging}: An unsupervised approach that adaptively learns layer-wise or parameter-wise merging coefficients by minimizing the prediction entropy on unlabeled test data.
    
    \item \textbf{Surgery}~\cite{pmlr-v235-yang24t}: Addresses representation discrepancy by introducing learnable auxiliary modules. It explicitly aligns the representation of fine-tuned models with the merged backbone to minimize geometric misalignment.
    
    \item \textbf{ProbSurgery}~\cite{wei2025probsurgery}: A state-of-the-art extension of Surgery that incorporates uncertainty modeling. It formulates representation alignment as a probabilistic process to robustly handle noise and parameter conflicts.
\end{itemize}

\subsection{Implementation Details}
\label{sec:app: implementation}
\paragraph{Setting of Hyperparameter} For the scaling hyperparameters of the Task Arithmetic and Ties models, we follow the original paper by using a portion of the labeled data for selection. These parameters are then fixed across all experiments to ensure a fair comparison. We employ the Adam optimizer across all experiments. For both CV and NLP tasks, we set the mask learning rate to $0.1$ and the L2 regularization weight to $5 \times 10^{-9}$, enabling the model to converge to a sparsity level (masking rate) between 10\% and 90\%. For task-specific parameters, we use a learning rate of $1 \times 10^{-3}$ with a batch size of 16 for CV tasks, and $5 \times 10^{-4}$ with a batch size of 24 for NLP tasks.

\paragraph{Formulation of Classifiers in CV Tasks}
In the ViT series of experiments, the image input is processed through the image encoder, and text inputs (label names) are processed through the text encoder, obtaining the image embedding $\mathbf{E}_{\rm image}$ and text embeddings $\mathbf{E}_{\rm text}$ respectively. Classification is performed by matching $\mathbf{E}_{\rm image}$ with $\mathbf{E}_{\rm text}$.
In the settings of most model merging papers, the text encoder is frozen, meaning that $\mathbf{E}_{\rm text}$ remains unchanged. In this case, we consider $\mathbf{E}_{\rm text}$ can be understood as the classifier parameter $\mathbf{W}_t$, that is, initializing the classifier with label name embeddings. This approach is consistent with the way many text classification methods initialize label embeddings. This structural equivalence explains why we explicitly discuss a `classifier head' despite employing a ViT-based architecture.

\section{Main Results on ViT-B/16}
\label{sec:app:vit-16}
Main results of MOMA on ViT-B/16 are shown in Table~\ref{tab:vit16}. This finding is in line with the results observed for ViT-B/32.

\begin{table*}[h]
    \centering
    \resizebox{\linewidth}{!}{
    \begin{tabular}{l|cccccccc|c}
    \toprule
        \textbf{Method}  & \textbf{SUN397} & \textbf{Cars} & \textbf{RESISC45} & \textbf{EuroSAT} & \textbf{SVHN} & \textbf{GTSRB} & \textbf{MNIST} & \textbf{DTD} & \textbf{Avg.} \\
        \midrule
        Pre-trained &63.8 &64.6 &65.7 &54.5 &52.0 &43.3 &51.7 &45.1 &55.0  \\
        Fine-tuned &81.8 &86.8 &96.9 &99.7 &97.8 &99.1 &99.7 &82.0 &92.9  \\
        \midrule
        Fisher Merging~\citep{matena2022fishermerging} &68.5 &69.9 &75.2 &80.4 &73.2 &61.2 &94.5 &50.7 &71.7  \\
        RegMean~\citep{jin2022dataless_regmean} &69.1 &71.6 &77.6 &88.8 &83.7 &70.2 &96.9 &54.6 &76.6  \\
        \midrule
        Weight Averaging  & 67.7 & 70.0 & 75.3 & 79.5 & 74.9 & 60.1 & 94.4 & 43.8 & 70.7 \\
        \quad w/ Surgery~\citep{pmlr-v235-yang24t} &70.3 &72.4 &88.8 &97.6 &82.0 &83.1 &98.1 &68.5 &82.6 \\
        \quad w/ ProbSurgery~\citep{wei2025probsurgery} & 74.0 & 79.4 & 95.7 & 99.7 & 87.8 & 98.8 & 99.4 & 81.2 & \textbf{89.5}\\
        \rowcolor{gray!15} \quad w/ MOMA &70.48 & 63.41 & 89.94 & 97.48 & 93.92 & 90.27 & 98 & 74.95 & 84.81 \\
        \midrule
        Task Arithmetic~\citep{ilharco2022editing_task_arithmetic} & 61.1 & 65.9 & 74.0 & 76.2 & 88.0 & 73.9 & 98.4  & 53.0 & 73.8 \\
        \quad w/ Surgery~\citep{pmlr-v235-yang24t}&68.3 &72.3 &88.7 &97.7 &91.0 &89.5 &98.9 &72.9 &84.9 \\
        \quad w/ ProbSurgery~\citep{wei2025probsurgery}&71.9 &79.3 &95.8 &99.7 &93.5 &99.0 &99.4 &81.5 &90.0 \\
        \rowcolor{gray!15} \quad w/ MOMA &75.4 &78.5 &94.7 &99.4 &97.3 &98.7 &99.4 &79.8 &\textbf{90.4} \\
        \midrule
        Ties~\citep{yadav2024ties} & 69.1 & 72.5 & 80.5 & 84.0 & 85.0 & 71.5 & 98.1 & 54.9 & 77.0 \\
        \quad w/ Surgery~\citep{pmlr-v235-yang24t}&73.0 &76.2 &90.7 &98.1 &89.7 &86.7 &98.7 &75.2 &86.0 \\
        \quad w/ ProbSurgery~\citep{wei2025probsurgery} &75.1 &80.9 &95.8 &99.7 &92.2 &99.0 &99.5 &82.0 &\textbf{90.5} \\
        \rowcolor{gray!15} \quad w/ MOMA &75.3 &78.4 &94.4 &99.3 &97.2 &98.3 &99.4 &79.3 &90.2 \\
    \bottomrule    
    \end{tabular}
    }
    \caption{Comparison of MOMA on ViT-B/16.}
    \label{tab:vit16}
\end{table*}

\section{Aligned-C: Aligning the Classifier from the Perspective of the Evaluation}
\label{app:sec:aligned-c}
\subsection{Current Evaluation of Merging Models}
\begin{figure*}[h]
    \centering
    \includegraphics[width=0.95\linewidth]{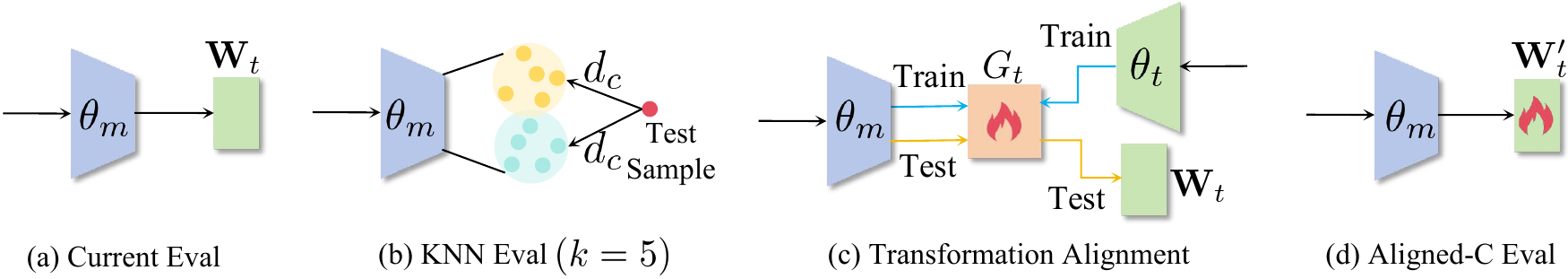}
    \caption{Illustrations of some variants referred to in this paper.}
    \label{fig:illustration}
\end{figure*}

Existing merging methods in classification tasks primarily focus on merging fine-tuned encoders, while not performing the merging operation on fine-tuned classifiers. That is, the merging encoders should be evaluated on the fine-tuned classifier, which can be formulated as:
\begin{equation}
    \hat{Y}_t = g(x; \theta_m)^\top \mathbf{W}_t,
    \label{eq:current_evaluation}
\end{equation}
where $\mathbf{W}_t$ represents the weight of the fine-tuned classifier of task $t$, which maps the merging outputs to the predicted labels $\hat{Y}_t$. In the following descriptions, this currently adopted evaluation protocol is referred to as `Current Eval', shown in Figure~\ref{fig:illustration}.

\subsection{KNN Evaluation of Merging Models}
As described in Section~\ref{sec:knn}, instead of utilizing the fine-tuned classifier head, we select $K$ samples as anchors to perform k-Nearest Neighbor (KNN) classification to evaluate the merging methods. We refer to this protocol as `KNN Eval'.

\begin{figure}[h]
  \centering
    \includegraphics[width=0.5\linewidth]{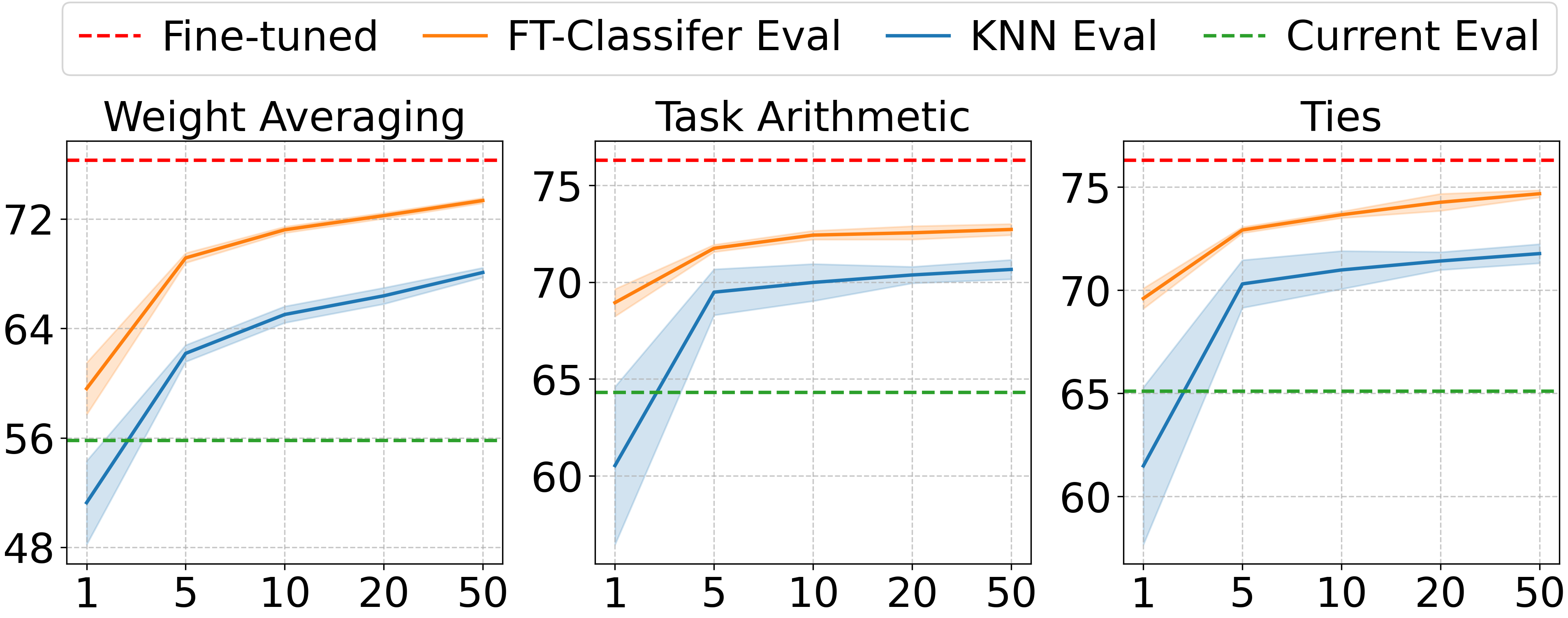}
    \caption{Performance of different evaluation protocols under varying numbers of few-shot examples $K$ using BERT. Corresponding detailed results for $k=5$ are shown in Table~\ref{tab:ft-classifier}.}
\label{fig:aligned-c}
\end{figure}

\begin{figure*}[h]
    \centering
    \begin{tabular}{@{}c@{\hspace{4mm}}c@{}}
        \includegraphics[width=0.48\linewidth]{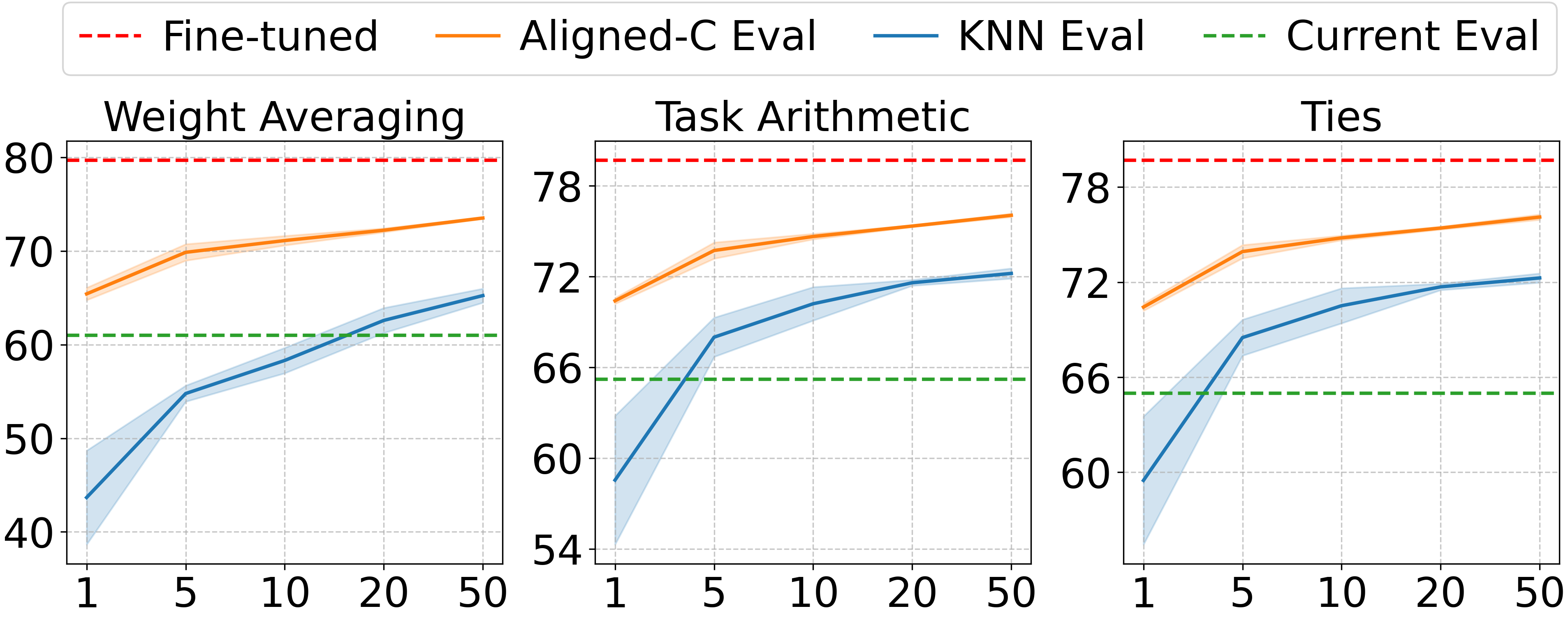} & \includegraphics[width=0.48\linewidth]{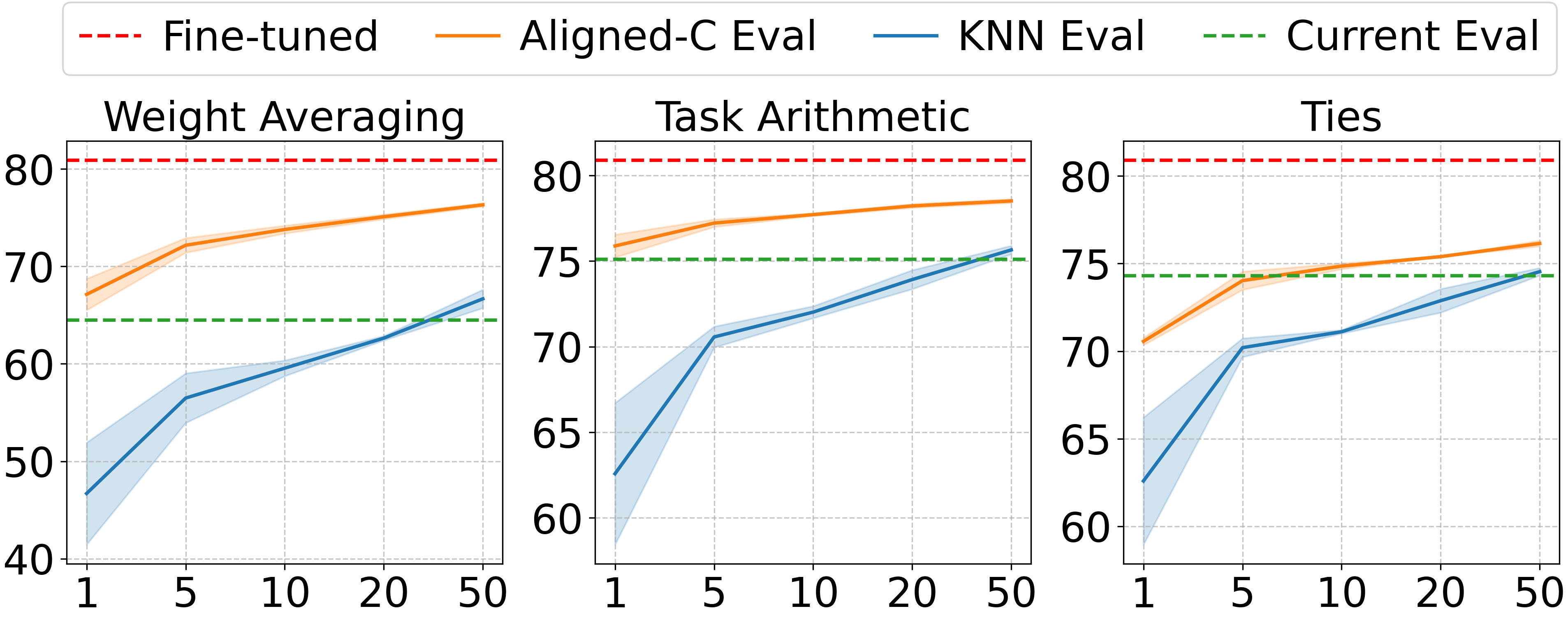} \\
        (a) Performance on T5-base & (b) Performance on T5-large
    \end{tabular}
    \caption{Performance of different evaluation protocols under T5-base and T5-large.
    Detailed results on $k=20$ are reported in Table~\ref{tab:ft-classifier-t5-side}.
    }
    \label{fig:t5}
\end{figure*}

\begin{figure*}[h]
    \centering
    \begin{tabular}{@{}c@{\hspace{4mm}}c@{}}
        \includegraphics[width=0.48\linewidth]{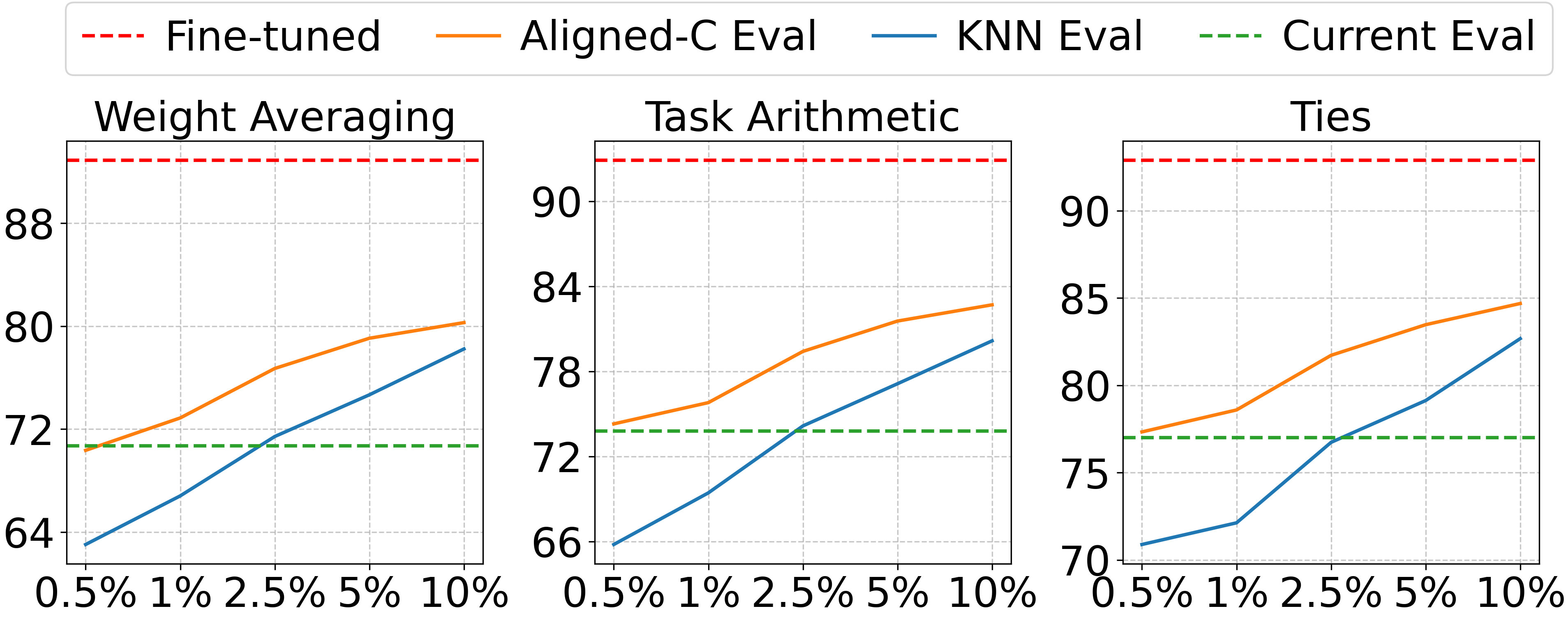} & \includegraphics[width=0.48\linewidth]{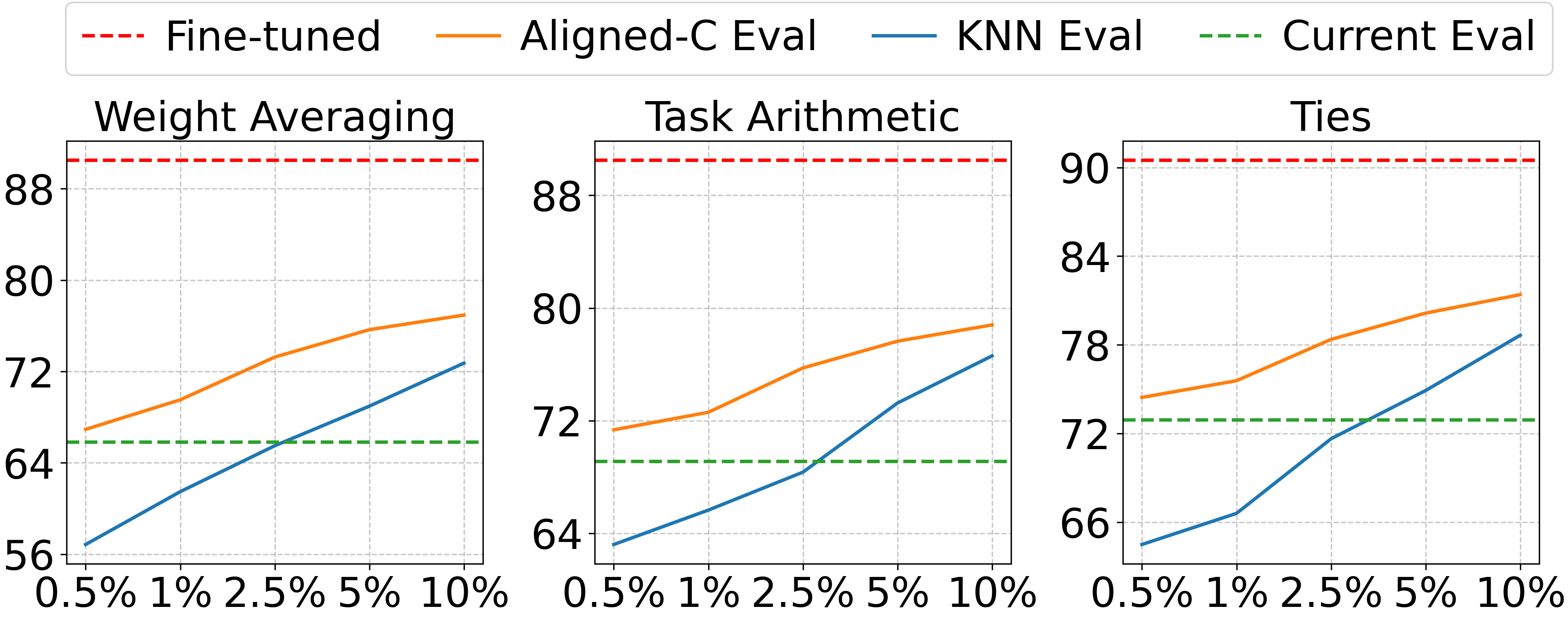}\\
        (a) CV tasks performance on ViT-B/16 &
        (b) CV tasks performance on ViT-B/32
    \end{tabular}
    \caption{The performance of different evaluation protocols on CV tasks under ViT-B/16 and ViT-B/32.}
    \label{fig:cv}
\end{figure*}

\subsection{Aligned-C Evaluation}
\label{sec:Aligned-C}
Model merging combines the parameters of multiple fine-tuned models to produce a merged model. Consequently, as discussed before, the merging representation inevitably shifts in the latent space relative to the fine-tuned representation, and this shift is unavoidable.
In this case, using $\mathbf{W}_t$, which is aligned well with fine-tuned representations, to evaluate the merged model is inappropriate. The evaluation result reflects not the capability of the merged model, but rather its compatibility with the fine-tuned classifier $\mathbf{W}_t$. 
Moreover, through experiments, we find that simply aligning the merging representation and the fine-tuned representation in the latent space can significantly improve the classification performance.
Naturally, this motivates us to propose a new and reasonable evaluation protocol.

Revisiting the objective of model merging, which aims to generate a new model $\theta_m$ using $\theta_1, ..., \theta_T$ that performs well across all $T$ tasks, we note that the upper bound of $\theta_m$'s performance on a specific task $t$ corresponds to the result from the fine-tuned model $\theta_t$. Therefore, the objective of model merging is to minimize this performance discrepancy of $\theta_m$ and $\theta_t$~\cite{ortiz2024task}.
By optimizing this discrepancy, we could directly train an aligned classifier for $\theta_m$. That is, in the new evaluation protocol, we introduce an additional post-processing step while keeping the parameter scale and network structure unchanged, referred to as Aligned-C Eval, which is defined as:
\begin{equation}
\begin{aligned}
    \arg\min_{\mathbf{W}_t^{\prime}} \sum_{t=1}^T KL (g(x; \theta_m) \! ^\top \! \mathbf{W}_t^{\prime}, g(x; \theta_t) \! ^\top \! \mathbf{W}_t )
\end{aligned}
\label{eq:objective_head}
\end{equation}
where $KL$ is the KL divergence~\cite{kullback1951information}, and the new classifier $\mathbf{W}_t^{\prime}$ is initialized with the value of $\mathbf{W}_t$.

In model merging, obtaining a few-shot dataset for each task is a relatively flexible requirement. Many studies use few-shot data to guide the merging methods~\cite{xiao2023lm_cocktail}, and some even rely on large-scale labeled or unlabeled data to determine merging weights~\cite{yangadamerging,jin2022dataless_regmean} or the conflict parameters to be pruned~\cite{ortiz2024task,zhumodel}. As a result, we also incorporate a few-shot sample for aligned classifier training in Aligned-C Eval.

Comparable results of Current Eval, KNN Eval, and Aligned-C Eval are presented in Figure~\ref{fig:aligned-c}, Figure~\ref{fig:t5} and Figure~\ref{fig:cv}, and detailed results, utilizing few-shot samples are shown Table~\ref{tab:ft-classifier}, Table~\ref{tab:ft-classifier-t5-side}, and Table~\ref{tab:ft-classifier-cv}. From the results, we observe that after simple training, even on 1-shot samples on NLP tasks, Aligned-C Eval performs significantly better than the Current Eval. This further demonstrates that the misalignment issue is highly manageable. Such a lightweight and computationally feasible post-processing operation facilitates the evaluation of merged models.

\begin{table*}[h]
    \centering
    \small
    \begin{tabular}{c|ccccc|c}
    \toprule
        Method  & AG News & Yelp & Amazon & Yahoo & DBpedia & Average \\
        \midrule
        Fine-tuned model & 91.6  & 60.4  & 57.9  & 72.7  & 98.8  & 76.3 \\
         Multi-task Learning & 90.6 & 59.1 & 55.6 & 71.3 & 98.5 & 75.0 \\
        \midrule
        Weight Averaging  & 79.2  & 49.8  & 45.0  & 50.3  & 55.1  & 55.8 \\
        + KNN Eval & 81.9 $\pm$ 1.6 & 42.1 $\pm$ 4.3 & 41.2 $\pm$ 1.5 & 54.1 $\pm$ 3.5 & 91.5 $\pm$ 1.8 & 62.2 $\pm$ 0.6 \\
        + Aligned-C Eval & \textbf{86.0 $\pm$ 1.0} & \textbf{51.5 $\pm$ 1.9} & \textbf{46.3 $\pm$ 0.6} & \textbf{66.2 $\pm$ 1.1} & \textbf{95.8 $\pm$ 0.4} & \textbf{69.2 $\pm$ 0.4} \\
        \midrule
        Task Arithmetic & 82.9 & 55.8 & 48.4 & 53.1 & 81.5 & 64.3 \\
        + KNN Eval & 86.4 $\pm$ 1.3 & 54.2 $\pm$ 2.6 & 49.0 $\pm$ 1.6 & 61.6 $\pm$ 1.0 & 96.3 $\pm$ 0.2 & 69.5 $\pm$ 1.2 \\
        + Aligned-C Eval & \textbf{87.5$\pm$ 0.7} & \textbf{55.6$\pm$ 0.5} & \textbf{51.5 $\pm$ 0.6} & \textbf{67.6 $\pm$ 0.7} & \textbf{96.6 $\pm$ 0.3} & \textbf{71.8 $\pm$ 0.2} \\
        \midrule
        Ties & 83.3  & 55.7  & 47.0  & 55.9  & 83.4  & 65.1 \\
        + KNN Eval & 87.0 $\pm$ 1.6 & 54.6 $\pm$ 2.9 & 49.5 $\pm$ 1.5 & 63.4 $\pm$ 0.6 & 97.1 $\pm$ 0.2 & 70.3 $\pm$ 1.2 \\
        + Aligned-C Eval & \textbf{88.1 $\pm$ 0.5} & \textbf{57.1 $\pm$ 0.9} & \textbf{52.6 $\pm$ 0.6} & \textbf{69.1 $\pm$ 0.6} & \textbf{97.7 $\pm$ 0.1} & \textbf{72.9 $\pm$ 0.1} \\
    \bottomrule
    \end{tabular}
    \caption{Comparison of Aligned-C Eval and KNN Eval with the corresponding Current Eval (represented by merging method name). The results are based on a few-shot setting of $k = 5$.}
    \label{tab:ft-classifier}
\end{table*}

\begin{table*}[h]
    \centering
    
    \begin{minipage}{0.48\textwidth}
    \centering
    \small
    \setlength{\tabcolsep}{3pt}
    \begin{tabular}{c|ccccc|c}
    
    \multicolumn{7}{c}{(a) T5-base} \\
    \toprule
        Method  & AG News & Yelp & Amazon & Yahoo & DBpedia & Avg. \\
        \midrule
        Fine-tuned & 94.5  & 66.6  & 62.9 & 75.2 & 99.1  & 79.6 \\
        \midrule
        WA & 79.0 & 51.8 & 51.7 & 42.2 & 79.9 & 60.9 \\
        +KNN & \textbf{85.4} & 48.8 & 47.3 & 41.5 & 94.6 & 63.5 \\
        +Aligned-C & 85.3 & \textbf{60.5} & \textbf{55.3} & \textbf{61.8} & \textbf{97.1} & \textbf{72.0} \\
        \midrule
        TA & 80.6 & 59.2 & 55.2 & 51.5 & 79.5 & 65.2 \\
        +KNN & 87.7 & 62.0 & 57.4 & 53.5 & 97.1 & 71.5 \\
        +Aligned-C & \textbf{88.2} & \textbf{64.0} & \textbf{58.2} & \textbf{67.6} & \textbf{98.4} & \textbf{75.3} \\
        \midrule
        Ties & 80.0  & 59.2  & 55.4  & 50.7  & 79.3  & 64.9 \\
        +KNN & 87.3 & 62.4 & 56.7 & 54.7 & 97.2 & 71.7 \\
        +Aligned-C & \textbf{88.3} & \textbf{63.8} & \textbf{58.1} & \textbf{68.0} & \textbf{98.4} & \textbf{75.4} \\
    \bottomrule
    \end{tabular}
    \end{minipage}
    \hfill
    \begin{minipage}{0.48\textwidth}
    \centering
    \small
    \setlength{\tabcolsep}{2pt}
    \begin{tabular}{c|ccccc|c}
    \multicolumn{7}{c}{(b) T5-large} \\
    \toprule
        Method  & AG News & Yelp & Amazon & Yahoo & DBpedia & Avg. \\
        \midrule
        Fine-tuned & 94.9  & 68.6 & 64.9 & 77.1 & 99.1 & 80.9 \\
        \midrule
        WA & 79.2 & 55.5 & 44.3 & 49.0 & 94.5 & 64.5 \\
        +KNN & 87.9 & 43.6 & 40.1 & 47.2 & 95.6 & 62.9 \\
        +Aligned-C & \textbf{88.0} & \textbf{62.6} & \textbf{59.7} & \textbf{67.9} & \textbf{98.2} & \textbf{75.3} \\
        \midrule
        TA & 90.8 & 58.3 & 60.5 & 72.9 & 92.7 & 75.1 \\
        +KNN & \textbf{91.3} & 59.0 & 54.4 & 62.7 & 97.7 & 73.0 \\
        +Aligned-C & 91.2 & \textbf{65.9} & \textbf{61.9} & \textbf{73.5} & \textbf{98.7} & \textbf{78.3} \\
        \midrule
        Ties & 91.3 & 56.0 & \textbf{60.3} & \textbf{73.2} & 90.4 & 74.3 \\
        +KNN & \textbf{91.8} & 59.1 & 51.5 & 65.3 & 97.5 & 73.0 \\
        +Aligned-C & 88.4 & \textbf{63.9} & 58.2 & 68.0 & \textbf{98.4} & \textbf{75.4} \\
    \bottomrule
    \end{tabular}
    \end{minipage}
    \caption{Comparison on five NLP classification tasks using T5-base and T5-large under few-shot setting ($k = 20$). Avg. represents Average. KNN represents KNN Eval, and Aligned-C refers to our proposed Aligned-C Eval.}
    \label{tab:ft-classifier-t5-side}
\end{table*}

\begin{table*}[h]
    \centering
    \resizebox{0.8\linewidth}{!}{
    \begin{tabular}{c|cccccccc|c}
    \multicolumn{10}{c}{(a) ViT-B/16} \\
    \toprule
        Method  & SUN397 & Cars & RESISC45 & EuroSAT & SVHN & GTSRB & MNIST & DTD & Average \\
        \midrule
        Fine-tuned & 81.8 & 86.8 & 96.9 & 99.7  & 97.8  & 99.1 & 99.7 & 82.0 & 92.9 \\
        \midrule
        Weight Averaging  & 67.7 & 70.0 & 75.3 & 79.5 & 74.9 & 60.1 & 94.4 & 43.8 & 70.7 \\
        +KNN Eval & 65.9 & 69.4 & \textbf{87.6} & 88.8 & 77.3 & 81.7 & 95.5 & \textbf{59.6} & 78.2 \\
        +Aligned-C Eval & \textbf{71.5} & \textbf{73.6} & 86.7 & \textbf{91.9} & \textbf{79.5} & \textbf{83.1} & \textbf{97.1} & 58.7 & \textbf{80.3} \\
        \midrule
        Task Arithmetic & 61.1 & 65.9 & 74.0 & 76.2 & 88.0 & 73.9 & 98.4  & 53.0 & 73.8 \\
        +KNN Eval & 65.0 & 67.9 & 84.9 & 85.9 & 88.9 & 88.5 & \textbf{98.8} & 61.3 & 80.2 \\
        +Aligned-C Eval & \textbf{68.1} & \textbf{72.3} & \textbf{86.3} & \textbf{90.8} & \textbf{90.2} & \textbf{90.3} & \textbf{98.8} & \textbf{64.6} & \textbf{82.7} \\
        \midrule
        Ties & 69.1 & 72.5 & 80.5 & 84.0 & 85.0 & 71.5 & 98.1 & 54.9 & 77.0 \\
        +KNN Eval & 68.9 & 73.8 & 88.4 & 91.2 & 86.7 & 88.2 & 98.4 & 66.1 & 82.7 \\
        +Aligned-C Eval & \textbf{73.1} & \textbf{77.0} & \textbf{89.1} & \textbf{93.9} & \textbf{88.0} & \textbf{89.7} & \textbf{98.7} & \textbf{68.1} & \textbf{84.7} \\
    \bottomrule
    \multicolumn{10}{c}{}\\
    \multicolumn{10}{c}{(b) ViT-B/32} \\
    \toprule
        Method  & SUN397 & Cars & RESISC45 & EuroSAT & SVHN & GTSRB & MNIST & DTD & Average \\
        \midrule
        Fine-tuned & 75.3 & 77.7 & 96.1 & 99.7 & 97.5 & 98.7 & 99.7 & 79.4 & 90.5 \\
        \midrule
        Weight Averaging  & 65.3 & 63.4 & 71.4 & 71.7 & 64.2 & 52.8 & 87.5 & 50.1 & 65.8 \\
        +KNN Eval & 63.5 & 57.7 & 82.8 & 83.5 & 69.0 & 80.3 & 93.4 & 51.8 & 72.7 \\
        +Aligned-C Eval & \textbf{69.1} & \textbf{67.1} & \textbf{82.9} & \textbf{88.6} & \textbf{73.5} & \textbf{83.1} & \textbf{95.4} & \textbf{55.7} & \textbf{76.9} \\
        \midrule
        Task Arithmetic & 55.2 & 54.9 & 66.7 & 78.9 & 80.2 & 69.7 & 97.3 & 50.4 & 69.1 \\
        +KNN Eval & 59.7 & 58.4 & 79.4 & 88.7 & 83.7 & 89.6 & 97.8 & 55.4 & 76.6 \\
        +Aligned-C Eval & \textbf{62.5} & \textbf{61.5} & \textbf{80.4} & \textbf{89.9} & \textbf{86.4} & \textbf{89.9} & \textbf{98.4} & \textbf{61.3} & \textbf{78.8} \\
        \midrule
        Ties & 65.0 & 64.4 & 74.8 & 77.4 & 81.2 & 69.3 & 96.5 & 54.5 & 72.9 \\
        +KNN Eval & 65.2 & 60.8 & 84.0 & 87.7 & 83.2 & 88.5 & 97.4 & 62.5 & 78.6 \\
        +Aligned-C Eval & \textbf{69.7} & \textbf{68.7} & \textbf{84.6} & \textbf{89.8} & \textbf{85.8} & \textbf{89.9} & \textbf{98.0} & \textbf{64.7} & \textbf{81.4} \\
    \bottomrule
    
    \end{tabular}
    }
    \caption{Comparison on CV tasks by ViT. The results in this Table are based on a sampling rate of 10\%.}
    \label{tab:ft-classifier-cv}
\end{table*}

\subsection{$\theta_m$ vs. $\theta_b$}

Previous analyses demonstrate that different $\theta_m$ achieves significant improvements when evaluated with an aligned classifier. To further verify that the performance improvement stems from the inherent capability of the $\theta_m$ itself rather than additional information introduced by aligning, we tested the results of the original pre-trained models $\theta_b$ under different evaluation protocols. The experimental results are presented in Table~\ref{tab:fm_fb}.

From the Table, we observe that regardless of the evaluation protocol, $\theta_m$ consistently outperforms $\theta_b$. This demonstrates that all $\theta_m$ models effectively incorporate knowledge from fine-tuned models. Meanwhile, Aligned-C Eval does not show significant improvements over Current Eval when applied to $\theta_b$, indicating that models with limited parameters (only the classifier) cannot retain sufficient task-specific information. This proves that Aligned-C Eval just activates the fine-tuned information that $\theta_m$ already contains.

\subsection{Aligned-C vs. Surgery}
\label{sec:surgery}
The related work Surgery~\cite{pmlr-v235-yang24t} characterizes the overall difference between the merging representations and the fine-tuned representations as ``representation bias'', and reduces this bias by training additional parameters on a large amount of unlabeled data. 

\begin{figure*}[h]
    \centering
    \begin{minipage}{0.48\textwidth}
        \centering
        \includegraphics[width=\linewidth]{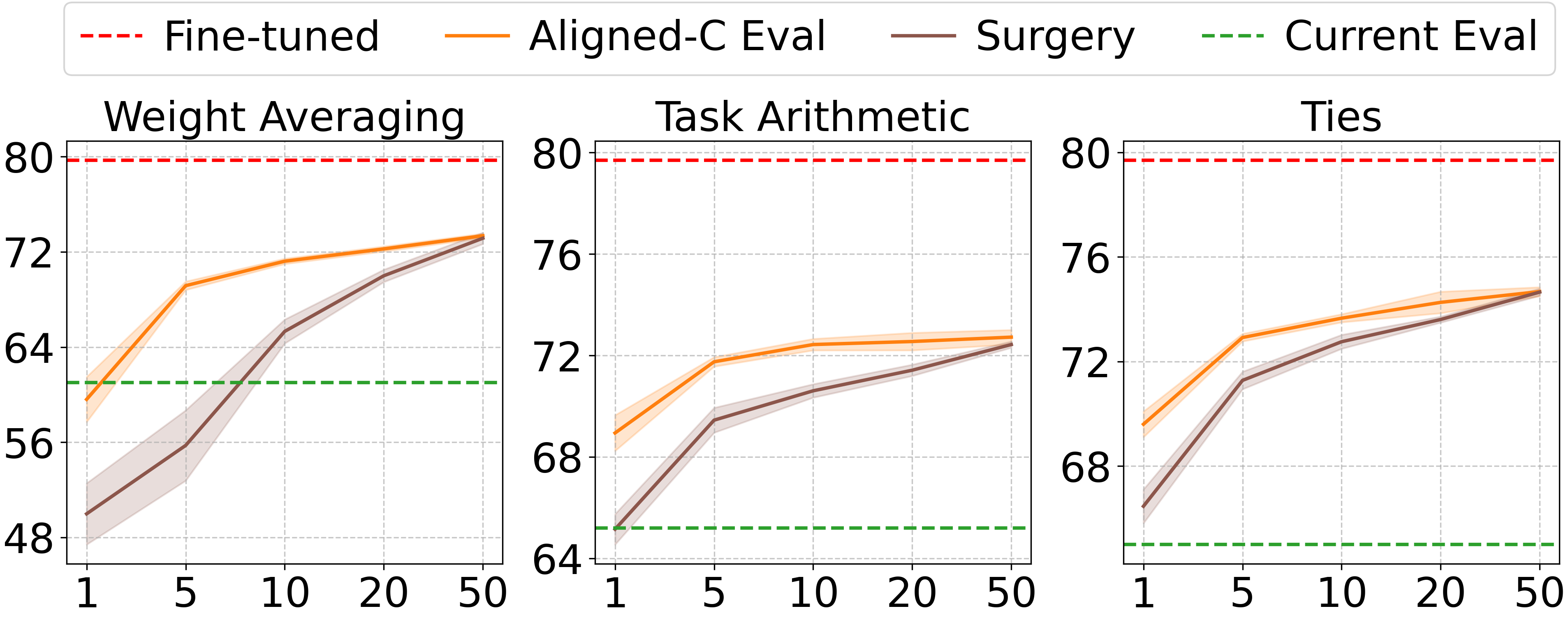} 
        \caption{Comparison of Aligned-C Eval and Surgery. Aligned-C outperforms Surgery particularly when only a small number of samples are available.}
        \label{fig:vs-surgery}
    \end{minipage}
    \hfill
    \begin{minipage}{0.48\textwidth}
        \centering
        \makeatletter\def\@captype{table}\makeatother 
        
            \begin{tabular}{l|cccc}
            \toprule
                Method & $\theta_b$ & WA & TA & Ties \\
                \midrule
                Current Eval & 17.5 & 55.8 & 64.3 & 65.1  \\
                KNN Eval & 20.9 & 62.2 & 69.5 & 70.3  \\
                Aligned-C Eval & 22.0 & 69.2 & 71.8 & 72.9 \\
            \bottomrule
            \end{tabular}
        \caption{Comparison of different $\theta_m$ and $\theta_b$ on different evaluation protocols.}
        \label{tab:fm_fb}
    \end{minipage}
\end{figure*}

We argue that merging models naturally alters the representation distribution (representation bias) but does not necessarily harm classification performance. The main cause of performance decline is the misalignment between the merging representation and the fine-tuned classifier. We also show that this misalignment is simple (converging to an orthogonal transformation) and does not affect the intrinsic properties of merging representations. Consequently, it can be mitigated by training an aligned classifier for each task with very few samples. In addition, the Aligned-C can also keep the parameter scale and network structure unchanged. In the context of model merging, we believe that such a lightweight and computationally efficient approach is more favorable. Results in Figure~\ref{fig:vs-surgery} demonstrate that the Aligned-C outperforms Surgery when only a small number of samples are available.

\end{document}